%% file: main.tex
\documentclass[sglanonrev]{informs4-nored}
\RequirePackage{endnotes}

\OneAndAHalfSpacedXI
\usepackage{algorithm}
\usepackage{algpseudocode}
\usepackage[dvipsnames]{xcolor}
\usepackage{tikz}
\usetikzlibrary{arrows.meta, positioning, fit, backgrounds, calc}

\usepackage{natbib}
 \bibpunct[, ]{(}{)}{,}{a}{}{,}%
 \def\bibfont{\small}%
\usepackage{bibunits}
\defaultbibliographystyle{informs2014}
\defaultbibliography{carshare.bib}

\newcommand{\T}{\top}
\newcommand{\Expt}{\mathbb{E}}

\newcommand{\Prob}{\mathrm{Pr}}

\newcommand{\fv}{\boldsymbol{f}}
\newcommand{\gv}{\boldsymbol{g}}
\newcommand{\hv}{\boldsymbol{h}}

\newcommand{\xv}{\boldsymbol{x}}
\newcommand{\gammav}{\boldsymbol{\gamma}}
\newcommand{\wv}{\boldsymbol{w}}
\newcommand{\lambdav}{\boldsymbol{\lambda}}
\newcommand{\ev}{\boldsymbol{1}}
\newcommand{\Iv}{\boldsymbol{I}}

\newcommand{\yv}{\boldsymbol{y}}
\newcommand{\zv}{\boldsymbol{z}}
\newcommand{\dv}{\boldsymbol{d}}

\newcommand{\Pv}{\boldsymbol{P}}

\newcommand{\muv}{\boldsymbol{\mu}}

\newcommand{\Xv}{\boldsymbol{X}}

\newcommand{\regret}{\mathrm{Regret}}
\newcommand{\pregret}{\mathrm{PseudoRegret}}

\newcommand{\NC}{[n]}

\newcommand{\bigo}{O}

\newcommand{\reposition}{M}
\newcommand{\loss}{\lambda}

\newcommand{\bslevel}{{\boldsymbol{S}}}
\newcommand{\per}{{\boldsymbol{\Gamma}}}
\newcommand{\policy}{\pi}

\usepackage{mathtools}

\newcommand{\K}{\mathbb{K}}

\newcommand{\pcr}[1]{{\normalfont\text{\fontfamily{pcr}\selectfont #1}}}
\usepackage{bbm}
\newtheorem{innercustomgeneric}{\customgenericname}
\providecommand{\customgenericname}{}

\newcommand{\newcustomtheorem}[2]{%
  \newenvironment{#1}[1]
  {%
    \renewcommand\customgenericname{{\normalfont\TheoremHeaderFont#2}}% Use small caps for the name  
   \renewcommand\theinnercustomgeneric{\normalfont ##1}%
   \vspace{-1em} % Adjust this value to control space above
   \innercustomgeneric
  }
  {\endinnercustomgeneric\vspace{-1em}} % Adjust this value to control space below
}

\newcustomtheorem{condition}{Condition}

\makeatletter
\def\paragraph{\@startsection{paragraph}{4}{0pt}{0.1pt plus 0.1pt minus 0.1pt}{-0.5em}% %use 0pt instead of 10pt to remove indent
    {\fs.11.{13.6}.\it\sffamily}}
\makeatother

\usepackage{enumitem}% http://ctan.org/pkg/enumitem

\usepackage[dvipsnames]{xcolor}
\definecolor{darkblue}{RGB}{26, 13, 171}

\usepackage[hidelinks]{hyperref}
\hypersetup{
    colorlinks,
    breaklinks,
    linkcolor=BrickRed,
    urlcolor=black,
    anchorcolor=black,
    citecolor=black
}

\usepackage[title]{appendix}

\usepackage{algorithm}
\usepackage{algpseudocode}

\usepackage{mathrsfs}

\usepackage{booktabs}
\usepackage{bbm}

\usepackage{changepage}

\usepackage{graphicx}
\usepackage[labelformat=simple, font= normalsize]{subcaption}

\EquationsNumberedThrough    % Default: (1), (2), ...
\TheoremsNumberedThrough     % Preferred (Theorem 1, Lemma 1, Theorem 2)
\ECRepeatTheorems  %  

\MANUSCRIPTNO{}

\begin{document}
%%%%%%%%%%%%%%%%

% Outcomment only when entries are known. Otherwise leave as is and
%   default values will be used.
%\setcounter{page}{1}
%\VOLUME{00}%
%\NO{0}%
%\MONTH{Xxxxx}% (month or a similar seasonal id)
%\YEAR{0000}% e.g., 2005
%\FIRSTPAGE{000}%
%\LASTPAGE{000}%
%\SHORTYEAR{00}% shortened year (two-digit)
%\ISSUE{0000} %
%\LONGFIRSTPAGE{0001} %
%\DOI{10.1287/xxxx.0000.0000}%

% Author's names for the running heads
% Sample depending on the number of authors;
% \RUNAUTHOR{Jones}
% \RUNAUTHOR{Jones and Wilson}
% \RUNAUTHOR{Jones, Miller, and Wilson}
% \RUNAUTHOR{Jones et al.} % for four or more authors
% Enter authors following the given pattern:
%\RUNAUTHOR{}
\RUNAUTHOR{Jiang, Sun, and Shen}

% Title or shortened title suitable for running heads. Sample:
% \RUNTITLE{Bundling Information Goods of Decreasing Value}
% Enter the (shortened) title:
\RUNTITLE{Spatial Supply Repositioning with Censored Demand Data}

% Full title. Sample:
% \TITLE{Bundling Information Goods of Decreasing Value}
% Enter the full title:
\TITLE{Spatial Supply Repositioning with Censored Demand Data}

% Block of authors and their affiliations starts here:
% NOTE: Authors with same affiliation, if the order of authors allows,
%   should be entered in ONE field, separated by a comma.
%   \EMAIL field can be repeated if more than one author
%----------
\ARTICLEAUTHORS{%
\AUTHOR{Hansheng Jiang 
}
\AFF{Rotman School of Management, University of Toronto.  
\EMAIL{}
%\EMAIL{hansheng.jiang@rotman.utoronto.ca}
} 
%\URL{}

\AUTHOR{Chunlin Sun 
}
\AFF{Stanford University. 
\EMAIL{}
%\EMAIL{chunlin@alumni.stanford.edu}
}

\AUTHOR{Zuo-Jun Max Shen 
}
\AFF{Faculty of Engineering and Faculty of Business and Economics,
University of Hong Kong. 
\EMAIL{}
%\EMAIL{maxshen@hku.hk}
}
%\URL{}
% Enter all authors
} % end of the block of ARTICLEAUTHORS

\ABSTRACT{We consider a network inventory system motivated by one-way, on-demand vehicle sharing services. Under uncertain and correlated network demand, the service operator periodically repositions vehicles to match a fixed supply with spatial customer demand while minimizing costs. Finding an optimal repositioning policy in such a general inventory network is analytically and computationally challenging. We introduce a base-stock repositioning policy as a multidimensional generalization of the classical inventory rule to $n$ locations, and we establish its asymptotic optimality under two practically relevant regimes. We present exact reformulations that enable efficient computation of the best base-stock policy in an offline setting with historical data.  In the online setting, we illustrate the challenges of learning with censored data in networked systems through a regret lower bound analysis and by demonstrating the suboptimality of alternative algorithmic approaches. We propose a Surrogate Optimization and Adaptive Repositioning algorithm and prove that it attains an optimal regret of $O(n^{2.5} \sqrt{T})$, which matches the regret lower bound in $T$ with polynomial dependence on $n$. Our work highlights the critical role of inventory repositioning in the viability of shared mobility businesses and illuminates the inherent challenges posed by data and network complexity. Our results demonstrate that simple, interpretable policies, such as the state-independent base-stock policies we analyze, can provide significant practical value and achieve near-optimal performance.
 }%

% \FUNDING{H. Jiang's research is supported by NSERC [grant number, funding agency].}

%Supplemental Material:
%Data Ethics & Reproducibility Note:

% Sample
%\KEYWORDS{Stochastic programming, Decision support, Uncertainty, Disaster response, Optimization}

% Fill in data. If unknown, outcomment the field
%\HISTORY{Received: Month DD, YYYY; Accepted: Month DD, YYYY; Published Online: Month DD, YYYY}

\maketitle
%%%%%%%%%%%%%%%%%%%%%%%%%%%%%%%%%%%%%%%%%%%%%%%%%%%%%%%%%%%%%%%%%%%%%%

% Text of your paper here

%***************************************************
\begin{bibunit}

\input{files/intro}

\input{files/formulation}

\input{files/offline_opt}

\input{files/offline}

\input{files/ogd_regret_analysis}
\input{files/extension}

\input{files/numerical_withfigures} % need to comment out 

\input{files/conclusion}

% Appendix here
% Options are (1) APPENDIX (with or without general title) or
%             (2) APPENDICES (if it has more than one unrelated sections)
% Outcomment the appropriate case if necessary
%
% \begin{APPENDIX}{<Title of the Appendix>}
% \end{APPENDIX}
%
%   or
%
% \begin{APPENDICES}
% \section{<Title of Section A>}
% \section{<Title of Section B>}
% etc
% \end{APPENDICES}

% Acknowledgments here
% \ACKNOWLEDGMENT{We would like to express our sincere gratitude to [acknowledge individuals, organizations, or institutions] for their invaluable contributions to this research. We are also grateful to [mention any additional acknowledgements, such as technical assistance, data providers, or colleagues] for their support and assistance throughout the course of this work.}

% \begingroup \parindent 0pt \parskip 0.0ex \def\enotesize{\footnotesize} \theendnotes \endgroup

\putbib

\end{bibunit}

%********************************************
% Appendix here
%********************************************

{

\clearpage

\begin{bibunit}
\begin{appendices}

{\centering \large Supplemental Materials for 
``Spatial Supply Repositioning with Censored Demand Data''}

%---change the equation numbering in the appendix------%
\renewcommand{\thesection}{\Alph{section}}
\numberwithin{equation}{section}
\numberwithin{theorem}{section}
\numberwithin{proposition}{section}
\numberwithin{lemma}{section}
\numberwithin{definition}{section}
\numberwithin{figure}{section}
\numberwithin{table}{section}
\numberwithin{assumption}{section}
\numberwithin{corollary}{section}
% Reset algorithm numbering in each section
\numberwithin{algorithm}{section}
%------------------------------------------------------%

% Commented out when submission
%\tableofcontents

\input{files/appendix_fullinfo}

\input{files/appendix_new_regret}

\input{files/appendix_omitted_proofs}

\input{files/appendix_numerical}

\input{files/appendix_extension}

\renewcommand\refname{References for Supplemental Materials}
\putbib
\end{appendices}
\end{bibunit}

}

%%%%%%%%%%%%%%%%%
\end{document}

%% file: files/intro.tex
\section{Introduction}
\label{section:introduction}
Urban traffic congestion and vehicle emissions are pressing issues in major cities worldwide. Free-floating carsharing has gained prominence over the past decade, with platforms such as Share Now in Europe, GIG in the United States, and Evo in Canada \citep{shaheen2020mobility}. In these services, a provider operates a fleet of vehicles distributed across a service region, which customers can rent on-demand for one-way trips. This flexible model offers greater autonomy and privacy than traditional ride-hailing or public transit \citep{martin2020evaluation}. For example, a customer can rent a private vehicle for the entire duration of a multi-stop grocery trip, offering a more convenient user experience than coordinating with a ride-hailing driver or navigating fixed bus routes. Beyond convenience, empirical studies estimate that each shared car may replace around 8 privately owned cars \citep{jochem2020does}, directly reducing vehicle ownership and associated emissions.

Despite their potential, these businesses face significant operational challenges, most notably the \emph{spatial mismatch} between vehicle supply and customer demand. Vehicle unavailability not only causes immediate revenue loss but can also erode customer trust and loyalty, which undermines long-term revenue, as show by several empirical evidences (see, e.g., \citet{kabra2020bike}). Such unavailability issues are common because customers’ one-way trips \emph{continually imbalance} the fleet distribution: vehicles tend to accumulate in certain locations while other high-demand areas become depleted of vehicles. Without intervention, these imbalances lead to a vicious cycle of lost sales and under-utilization. 

\begin{figure}[!htb]
\caption{Illustration of network imbalances with three locations.}
\label{fig:network_illustration}
\centering\resizebox{0.6\linewidth}{!}{
\begin{tikzpicture}[
    >={Stealth},
    thick,
    every node/.style={font=\small},
    demand node/.style={
        circle, % Shape of the node (can be rectangle, ellipse, etc.)
        draw=black, % Outline color
        shade, % Enables shading
        top color=Orange!10, % Top shading color
        bottom color=Orange!10 % Bottom shading color
    },
    inventory node/.style={
        rectangle, % Shape of the node (can be rectangle, ellipse, etc.)
        draw=black, % Outline color
    }
]

% ---------- colors & styles ----------
\definecolor{SurplusBlue}{RGB}{31,119,180}
\definecolor{DeficitRed}{RGB}{214,39,40}
\tikzset{
  surplus/.style  ={fill=SurplusBlue!22,  draw=SurplusBlue!60!black},
  deficit/.style  ={fill=DeficitRed!22,   draw=DeficitRed!60!black},
  balanced/.style ={fill=Green!22,         draw=Green!60!black},
}

% Inventory and Demand Nodes
% Location i
\node[inventory node, deficit, minimum width=0.6cm] (xi) at (-2, 0) {$x_{t,i}$}; % Inventory node
\node[demand node, minimum width=1.9cm] (di) at (-5, 0) {$d_{t,i}$}; % Demand node
\draw[dashed] (xi) -- node[midway, above] {\textit{Censored}} (di); % Connection between inventory and demand

% Location j
\node[inventory node, surplus, minimum width=2.2cm] (xj) at (4, 2.5) {$x_{t,j}$}; % Inventory node
\node[demand node, minimum width=1cm] (dj) at (6, 2.5) {$d_{t,j}$}; % Demand node
\draw[dashed] (xj) -- (dj); % Connection between inventory and demand

% Location k
\node[inventory node, balanced, minimum width=1.2cm] (xk) at (4, -2.5) {$x_{t,k}$}; % Inventory node
\node[demand node, minimum width=1.3cm] (dk) at (6, -2.5) {$d_{t,k}$}; % Demand node
\draw[dashed] (xk) -- (dk); % Connection between inventory and demand

% Arrows with transition probabilities
\draw[->] (xi) to[bend left=25] node[midway, above left] {OD probability $P_{t,ij}$} (xj); % From i to j
\draw[->] (xj) to[bend left=20] node[midway, below right] {$P_{t,ji}$} (xi); % From j to i
\draw[->] (xi) to[bend left=20] node[midway, above] {$P_{t,ik}$} (xk); % From i to k
\draw[->] (xk) to[bend left=20] node[midway, below] {$P_{t,ki}$} (xi); % From k to i
\draw[->] (xj) to[bend left=20] node[midway, right] {$P_{t,jk}$} (xk); % From j to k
\draw[->] (xk) to[bend left=20] node[midway, left] {$P_{t,kj}$} (xj); % From k to j

% Self-loops
\draw[->] (xi) edge[loop below, looseness=25] node[midway, left] {$P_{t,ii}$} (xi); % Self-loop on xi
\draw[->] (xj) edge[loop above, looseness=25] node[midway, left] {$P_{t,jj}$} (xj); % Self-loop on xj
\draw[->] (xk) edge[loop below, looseness=25] node[midway, left] {$P_{t,kk}$} (xk); % Self-loop on xk
    
% Dashed boundary
\draw[dashed, thick, gray] (-6.6, -4) rectangle (7, 4.2); % Service region boundary

% Add labels for locations (optional)
\node[above] at (-2, 0.4) {Supply};
\node[above] at (-5.1, 1.0) {Demand};
\end{tikzpicture}
}
\end{figure}
To combat spatial imbalances, carsharing operators rely on service staff  to reposition vehicles from surplus locations to deficit locations (e.g., from location $j$ to location $i$ in Figure~\ref{fig:network_illustration}), often during off-peak times such as overnight \citep{yang2022overnight}. Unlike ride-hailing where highly frequent trips and  profit-motivated drivers create constant opportunities for network rebalancing \citep{chen2024real}, carsharing operators lack similar organic levers. Given the labor costs associated with repositioning, a critical \emph{trade-off} between repositioning costs and lost sales costs arises. The difficulty of managing this trade-off is reflected in the financial struggles of on-demand vehicle sharing startups worldwide, which often rely on government subsidies and venture capital. For example, Share Now had to exit the North American market, and GIG Car Share announced it would terminate services by the end of 2024 \citep{YahooFinance2023GigCarShare}, citing difficulties such as ``\emph{decreased demand, rising operational costs, and changes in consumer commuting patterns}.''

\subsection{Main Contributions}

Motivated by these practical concerns, we study a fundamental problem of  \emph{spatial supply repositioning with censored demand data}, for which we present rigorous theoretical analyses and develop efficient learning algorithms with provable performance guarantees. Our problem is rooted in the rich literature on inventory control with demand learning, but it possesses a unique structure that distinguishes it from classical models.
\begin{itemize}
    \item First, we operate in a closed network with a fixed total inventory. Unlike retail inventory systems that can be replenished from an external supplier, vehicles are not consumed; they are rented and subsequently return to the network.
    \item Second, this inventory is mobile and intrinsically coupled across locations. A customer trip creates a correlated state transition, simultaneously decreasing inventory at the origin and increasing it at the destination. This coupling means that local supply levels are interdependent, which precludes purely localized control strategies.
    \item Finally, these network dynamics are compounded by the challenge of learning demand from censored data. The fixed fleet size prevents the use of common exploration strategies, such as overstocking all locations to observe true demand. Furthermore, a stockout at one location could be a consequence of vehicle flows originating from entirely different parts of the network.
\end{itemize}

Therefore, the challenge of decision-making in our setting is intricately coupled with the network's fixed-supply and multi-dimensional nature. To the best of our knowledge, this is the first study to address inventory control and demand learning in a closed network with such bidirectional flows. We summarize our main contributions as follows.

\begin{enumerate}
    \item {\bf Modeling and Structural Results.} 
    We present a parsimonious contunous-state average-cost Markov decision process (MDP) model that captures the key features of vehicle sharing networks: multi-location reusable inventory, random one-way trip flows, lost sales, and periodic repositioning. We rigorously prove the existence of a stationary optimal policy under the average-cost criterion, which is a non-trivial fact because the state and action spaces are continuous and multi-dimensional. We then introduce a class of simple base-stock policies for repositioning. Such policies are easy to implement and widely used as heuristics in practice. We analyze their performance and show that the best base-stock policy is asymptotically optimal in two practically relevant regimes: (i) a large fleet regime, where the number of locations $n$ grows large; and (ii) a high lost-sales regime, where stock-outs are very costly relative to repositioning.

    \item {\bf Offline Optimization of Repositioning Policy.} Computing the best base-stock levels from data is not straightforward, even in an offline setting with \emph{uncensored} demand. The basic formulation leads to a \emph{non-convex} stochastic optimization because of the piecewise-linear lost sales cost and the coupling across locations. We reformulate the offline problem exactly as a mixed-integer linear program (MILP) that can be solved with standard solvers. Furthermore, we identify a mild condition on the relationship between lost-sale cost and reposition cost, under which the offline problem simplifies to a linear program, which yields global optima and is more computationally efficient. We also derive generalization bounds for the offline solution, which statistically characterizes how the policy learned from a finite sample of demand data will perform close to optimal on the true distribution.

    \item {\bf Online Repositioning with Censored Demand. } We tackle the full learning-while-doing problem where (i) the demand distribution is unknown and (ii) only censored demand is observed over time. A naive approach treating each base-stock vector as an ``arm'' in a multi-armed bandit would suffer a regret bound $\widetilde{O}(T^{\frac{n}{n+1}})$ growing sublinearly in time $T$ but \emph{exponentially} in the number of locations $n$. We overcome this dimensionality challenge by designing a new algorithm, called \emph{Surrogate Optimization and Adaptive Repositioning} (\pcr{SOAR}) algorithm, that exploits the structure of the network cost function. The \pcr{SOAR} algorithm uses a carefully constructed surrogate cost function that provides gradient signals even with censored observations. By solving a sequence of small linear programs, \pcr{SOAR} adjusts the repositioning targets on the fly and provably converges to the optimum. We prove that \pcr{SOAR} achieves a regret on the order of $O(n^{2.5}\sqrt{T})$. Notably, this regret grows only sublinearly in $T$ and the dependence on $T$ is $\sqrt{T}$, which is independent of $n$. In fact, up to a polynomial factor in $n$, our regret rate matches the best-known lower bound $\Omega(\sqrt{T})$ for even single-location inventory learning problems. \pcr{SOAR} is also computationally efficient: each period’s update involves solving a linear program of size $O(n)$, making it scalable to large networks. We further show that these performance guarantees hold under both stochastic i.i.d. demand and adversarial demand sequences, which underscores the robustness of \pcr{SOAR}.

    \item {\bf Fundamental Limits and Further Insights.} To understand the fundamental difficulty of learning in multi-location systems, we establish the regret lower bound and consider alternative algorithms. We prove a regret lower bound of $\Omega(n\sqrt{T})$ for any learning algorithm in our setting, which implies that some dependence on the number of locations $n$ is unavoidable. Our \pcr{SOAR} algorithm’s regret scaling $O(n^{2.5}\sqrt{T})$ is only polynomially worse in $n$ and thus near-optimal in its dependence on both $T$ and $n$. Additionally, we examine special cases and simplified settings to build intuition. We construct a class of instances to illustrative how demand censoring can fundamentally prevent learning of true demand. On the flip side, if demand is fully observable, i.e., no censoring, we show that a simple dynamic learning strategy can achieve the optimal $\widetilde{O}(\sqrt{T})$ regret. We also consider a network independence scenario, and show that a one-time learning algorithm that leverages offline solution enjoys a provable regret guarantees of $\widetilde{O}(T^{2/3})$. These analyses deepen our understanding of when efficient learning is or isn’t possible, and they delineate the boundary between tractable and intractable cases for future research.

    \item {\bf Extension and Implication.} We extend our framework beyond the basic one-way rental model to accommodate more complex and realistic operational scenarios. We consider a setting where each period contains multiple rental subperiods with heterogeneous trip durations and start times. We demonstrate that our algorithmic approach, \pcr{SOAR}, can be adapted to this challenging setting while preserving its theoretical regret guarantees and numerical effectiveness. More broadly, we hope the analysis in this work could inform decision-making in other applications that involve periodic inventory allocation across networks with demand uncertainty.

\end{enumerate}

\subsection{Related Literature}
\label{section:literature_review}
\paragraph{Inventory Repositioning in Network.}  Early studies on repositioning in shared mobility examine stylized two‑location settings \citep{li2010determining}, while general $n$‑location formulations have appeared only recently. Representative approaches include distributionally robust optimization \citep{he2020robust}, two‑stage stochastic integer programming \citep{lu2018optimizing}, cutting‑plane approximate dynamic programming for discounted‑cost formulations \citep{benjaafar2018dynamic}, fluid‑model–based linear programming policies \citep{hosseini2021dynamic}, and mean-field approximation-based policies \citep{akturk2022managing}. These studies are either analytical or approximation‑based, assume known demand distribution (and thus do not learn from data), or require extensive histories of uncensored demand. None adopts an \emph{online learning} perspective in which the platform simultaneously collects censored demand and decides how to reposition the fleet.

Inventory repositioning is also studied in the transshipment literature, but a key difference is that inventory exits the system after sale and can be replenished from outside suppliers, instead of managing a fixed, reusable supply. The closest analogue is multi‑period proactive transshipment with lost sales; even there, optimal policies are intractable for general $n$‑location networks, with recent progress largely in two‑location settings \citep{abouee2015optimal}.

\paragraph{Asymptotic Optimality of Base‑Stock Policy.}
Our asymptotic results contribute to the literature on near‑optimality of base‑stock policies (see \citet{goldberg2021survey}). Pioneering work \citep{huh2009asymptotic} establishes asymptotic optimality in single‑location settings where the decision is scalar, whereas asymptotic optimality with multi‑dimensional decisions typically arises in models with perishability, service‑level constraints, or positive lead times \citep{wei2021deterministic,bu2024asymptotic}. By contrast, our base‑stock \emph{repositioning} policy specifies an $n$‑dimensional target vector across locations in a closed network with lost sales and bidirectional flows. We show that the best such policy is asymptotically optimal in practically relevant regimes, thereby extending base‑stock optimality guarantees to high‑dimensional \emph{network} inventory systems. Relatedly, \citet{devalve2022base} provide constant‑factor guarantees for base‑stock policies in newsvendor networks with backlogging, while our results address a different regime with finite, reusable inventory and lost sales.

\paragraph{Decision-Making with Censored Demand.}
Our learning-while-repositioning problem is related to the literature on decision-making with \emph{censored} demand. Inferring lost demand via app analytics or customer tracking is often impractical due to significant privacy hurdles and the introduction of intractable sampling biases \citep{xu2025locational}. The impact of censoring on decision quality is well documented. Even in \emph{offline} settings, censored data complicate estimation and policy optimization (see, e.g., \citet{besbes2013implications, fan2022sample,bu2023offline}). For single-location lost-sales inventory, \citet{huh2009adaptive} achieve $\widetilde{O}(T^{2/3})$ regret, while subsequent work attains the optimal $\widetilde{O}(\sqrt{T})$ rate  \citep{zhang2020closing,agrawal2019learning,ding2024feature}. Beyond the single-location case, online learning has also been explored in multi-echelon networks \citep{bekci2023inventory,lyu2023minibatch}, typically with \emph{external replenishment} and \emph{one-directional} flows that differ fundamentally from our closed, reusable-inventory network with bidirectional movements.

A notable approach in inventory learning is to allocate abundant stock to reduce censoring and thereby evaluate multiple policies offline (see, e.g., \citet{yuan2021marrying,chen2024learning}). This strategy is not operationally viable in our problem because the fixed inventory is typically insufficient to cover the support of demand across all locations.  A line of recent works (see, e.g., \citet{gong2021bandits,jia2024online,tang2024online}) adopt the idea of modeling MDP policies as bandit arms, including a $\widetilde{O}(\sqrt{T})$ regret rate driven by online stochastic convex optimization \citep{jia2024online}. Nevertheless, without convexity in the policy space and given the dimensionality $n$, we note in Section~\ref{section:lb_and_dimension} that a Lipschitz bandit-based approach would result in an regret bound of $\widetilde{O}\left(T^{\frac{n}{n+1}}\right)$ with unfavorable dependence on $n$. To our knowledge, there are no online learning results with regret guarantees for an \emph{average-cost} \emph{network} inventory problem with arbitrary inventory flows and censored observations. Our work addresses this gap while yielding tight online learning regret guarantees under both i.i.d. and adversarial inputs. We defer a more detailed discussion of related literature and technical differences to Section~\ref{sec:alg_des}.

%% file: files/formulation.tex
\section{Model}
\label{section:model}

% In this section, we introduce the formulation of the one-way on-demand vehicle sharing service and its cost structures,  and then we quantitatively model the vehicle repositioning problem as a Markov Decision Process (MDP).

\paragraph{Inventory Network.} The network contains $n \geq 2$ locations, denoted by $\NC = \{1,\dots,n\}$. Customers can pick up vehicles from any location $i\in \NC$ at the beginning of period $t$ and return them to any location $j \in \NC$ at the end of period $t$. Let $d_{t,i}$ denote the uncensored demand at location $i$ in period $t$,  defined as the number of vehicles requested to depart from location $i$ at the start of period $t$, and let $\dv_t = \{d_{t,i}\}_{i\in \NC}$ denote the demand vector. We assume the review and rental periods coincide, and each rental unit is used at most once per review period, consistent with \cite{he2020robust,akturk2022managing}. Section~\ref{subsection:extension} relaxes this to allow heterogeneous rental and review lengths. Depending on inventory sufficiency at each location, some requests may be lost, so realized demand may be lower than $\dv_t$. The origin-to-destination (OD) probability matrix $\Pv_t=(P_{t,ij})_{1\le i,j\le n}$ collects the fractions of rentals that return across locations, where $P_{t,ij}$ is the fraction of vehicles rented at location $i$ in period $t$ that are returned to location $j$ at the end of $t$. We assume all vehicles rented at the start of $t$ are returned to some location by the end of $t$, so each row of $\Pv_t$ is stochastic: $\sum_{j=1}^n P_{t,ij}=1$ for all $i$ and $t$.

\begin{assumption}\label{assumption3}
The joint distribution of $\{(\dv_t, \Pv_t)\}$ is independently and identically distributed (i.i.d.) across different time period $t$, following some distribution $\muv$. 
\end{assumption}
\begin{remark}
Notably, Assumption~\ref{assumption3} allows spatial correlation across the $n$ locations and the  correlation  between $\dv_t$ and $\Pv_t$. As we discuss in Section~\ref{subsec:network_indep}, an additional network independence assumption on the demand would significantly simplify learning.
\end{remark}
\begin{remark}
    The i.i.d. stochastic assumption is widely adopted in the inventory control literature with demand learning (see, e.g., \citet{chen2022elements}). While we introduce Assumption~\ref{assumption3} to facilitate theoretical analysis of MDPs, we note that our \pcr{SOAR} algorithm in Section~\ref{section:ogr} actually does \emph{not} rely on this assumption and achieves optimal regret even under adversarial demand scenarios, as rigorously proved in Theorem~\ref{thm:gdr}. Moreover, the model extension in Section~\ref{subsection:extension} also accommodates cyclic demand patterns, further relaxing the i.i.d. assumption.
\end{remark}

\paragraph{Inventory Update.} At the beginning of period $t$, after observing the pre-repositioning inventory $\xv_t=(x_{t,1},\dots,x_{t,n})$, the service provider selects a target post-repositioning inventory $\yv_t=(y_{t,1},\dots,y_{t,n})$ and repositions to reach $\yv_t$. The fleet size is fixed; we treat inventory as divisible and normalize the total to one. Hence $\xv_t,\yv_t\in\Delta_{n-1}$, where $\Delta_{n-1}(K) := \{ (x_1,\dots, x_n) \mid \sum_{i=1}^n x_i = K, x_i \geq 0 \text{ for all } i\}$ and  $\Delta_{n-1} := \Delta_{n-1}(1)$.

\begin{figure}[!htb]
\centering
% \vspace{-0.5em}
\caption{Sequential events of demand arrival and repositioning operation.}
\label{fig:inventory_update}
\begin{tikzpicture}[>=Latex,
% Define styles for different types of boxes
    x-box/.style={draw, rounded corners, fill=blue!20, opacity=0.8},
    y-box/.style={draw, rounded corners, fill=purple!20, opacity=0.8},
    d-box/.style={draw, rounded corners, fill=teal!20, opacity=0.8},
    note/.style={align=center, font=\footnotesize}
    ]
% Timeline
\draw[->, thick] (0,0) -- (9,0);
% Time points, ticks, and labels
\foreach \i/\x in {1/0, 2/2, 3/4, 4/6} {
    % Vertical tick
    \draw[thick] (\x,0.2) -- (\x,0);
    % Labels below and above the tick
    \node[x-box, below] at (\x,-0.4) {$\xv_{\i}$};
    \node[y-box, above] at (\x,0.4) {$\yv_{\i}$};
}
% Dots for continuation
\draw[thick] (8,0.2) -- (8,0); % Vertical tick at the end
\node[below] at (8,-0.4) {$\cdots$};
\node[above] at (8,0.4) {$\cdots$};
\node at (10,0) {Time};
% Period index annotations
\foreach \i/\x in {1/1, 2/3, 3/5, 4/7} {
    \node at (\x, -0.2) {\footnotesize Period \i};
}
% Trip index annotations
\foreach \i/\x in {1/1, 2/3, 3/5, 4/7} {
    \node[d-box] at (\x, 1.3) {\footnotesize $(\dv^{c}_{\i}, \Pv_{\i})$};
}
% Annotations for reposition and rental
\foreach \x in {-2} {
    \node[orange] at (\x,0.6) {Reposition to};
    \node[teal] at (\x,-0.6) {Rentals occur};
}
% Dashed curves for flow
\foreach \x/\nextx in {0/2, 2/4, 4/6, 6/8} {
    % Curve from \xv_t to \yv_t
    \draw[thick, ->, orange] (\x,-0.4) .. controls (\x-0.4,0.1) .. (\x,0.4);
    % Curve from \yv_t to \xv_{t+1}
    \draw[dashed, thick, ->, teal] (\x,0.4) .. controls (\x+1.1,0.25) .. (\nextx,-0.4);
    % Curve from \dv_t to \xv_{t+1}
    \draw[dashed, thick, ->, teal] (\x+1,1) .. controls (\x+1.1,0.25) .. (\nextx,-0.4);
}
\end{tikzpicture}
% \vspace{-1em}
\end{figure}

After rentals in period $t$, the inventory at location $i$ at the start of period $t{+}1$ is
$
x_{t+1,i}
= (y_{t,i}-d_{t,i})^{+} + \sum_{j=1}^{n} \min(y_{t,j},d_{t,j})\,P_{t,ji},
$
where $(y_{t,i}-d_{t,i})^{+} := \max\{y_{t,i}-d_{t,i},0\}$ is the leftover inventory at $i$, and the sum captures vehicles returned to $i$. In vector form,
\begin{equation}
\label{eq:state_update_vector}
\xv_{t+1} \;=\; (\yv_t-\dv_t)^{+} \;+\; \Pv_t^{\T}\min(\yv_t,\dv_t),
\end{equation}
where $(\cdot)^{+}$ and $\min(\cdot,\cdot)$ are applied elementwise and $\dv_t^{c}:=\min(\yv_t,\dv_t)$ denotes censored demand. Figure~\ref{fig:inventory_update} illustrates this update.

\paragraph{Cost Structure.}
We consider two cost components: lost sales and repositioning. Lost sales reflect not only foregone trip revenue but also broader opportunity costs (e.g., churn, brand dilution, slower growth, and idle-time depreciation). With unit lost-sales costs $l_{ij}$, the period-$t$ loss is
\begin{equation}
\label{eq:lost_Sales_cost}
L(\yv_t,\dv_t,\Pv_t)
= \sum_{i=1}^n \sum_{j=1}^n l_{ij}\, P_{t,ij}\, (d_{t,i}-y_{t,i})^+ .
\end{equation}

Repositioning moves inventory from $\xv_t=(x_{t,1},\dots,x_{t,n})$ to $\yv_t=(y_{t,1},\dots,y_{t,n})$. Given unit repositioning costs $c_{ij}$ and flows $\xi_{t,ij}$, the single-period cost is the optimal value of the minimum-cost flow:
\begin{align}
\reposition(\yv_t-\xv_t)
= \min_{\{\xi_{t,ij}\}} \ & \sum_{i=1}^n \sum_{j=1}^n c_{ij}\, \xi_{t,ij} \label{eq:repo_cost}\\
\text{s.t. } \ & \sum_{i=1}^n \xi_{t,ij} - \sum_{k=1}^n \xi_{t,jk} = y_{t,j}-x_{t,j}, \quad j=1,\dots,n, \label{eq:flow_balance}\\
& \xi_{t,ij} \ge 0, \quad i,j=1,\dots,n, \label{eq:flow_nonneg}
\end{align}

Notationally, we write the optimum as $\reposition(\yv_t-\xv_t)$ because the net change $\yv_t-\xv_t$ fully determines \eqref{eq:flow_balance}. 
The total period-$t$ cost is
\begin{equation}
\label{eq:total_cost}
C_t(\xv_t,\yv_t,\dv_t,\Pv_t)
= \reposition(\yv_t-\xv_t) + L(\yv_t,\dv_t,\Pv_t).
\end{equation}

\paragraph{MDP Formulation.}
We model repositioning as an average-cost MDP with state $\xv_t\in\Delta_{n-1}$, the pre-repositioning inventory at time $t$. Because $\yv_t$ suffices to determine both the state update \eqref{eq:state_update_vector} and the cost \eqref{eq:total_cost}, we take $\yv_t\in\Delta_{n-1}$ as the action, rather than the flow solution $\{\xi_{t,ij}\}$. Let $\mathcal{F}_t$ denote the history up to time $t$, comprising realizations of $\min(\dv_\tau,\yv_\tau)$ and $\Pv_\tau$ for $\tau=1,\dots,t$.

An \emph{admissible} policy $\pi$ maps $(\xv_t,\mathcal{F}_{t-1})$ to $\yv_t$. Given initial state $\xv_1=\xv$, the $T$-period average cost under $\pi$ is
\begin{equation}
\label{eq:average_cost_defn}
v_T^{\pi}(\xv)
= \frac{1}{T}\sum_{t=1}^{T} \Expt\left[C_t^{\pi} \,\middle|\, \xv_1=\xv \right]
= \frac{1}{T}\sum_{t=1}^{T} \Expt\left[C_t^{\pi}(\xv_t,\yv_t,\dv_t,\Pv_t)\,\middle|\, \xv_1=\xv \right],
\end{equation}
where $\{\xv_t\}_{t\ge2}$ and $\{\yv_t\}_{t\ge1}$ evolve under $\pi$. In the infinite-horizon setting, we minimize the long-run average cost $v^{\pi}(\xv)$, formally defined via the standard average-cost criterion.
%(the limit need not exist on continuous state spaces; see Section~\ref{section:fullinfo})
The average-cost objective is natural here because discount factors may be unclear or close to one under many business scenarios. Although discounted problems are analytically more tractable with effectively finite horizon $\approx 1/(1-\rho)$ given discount factor $\rho$, we show in Section~\ref{subsection:fullinfo_opt} that, under general conditions, the long-run average cost is independent of the initial state, which in return facilitates the comparison of different repositioning policies.

%% file: files/offline_opt.tex
\section{Benchmark Policy and Learning Setup}
\label{section:fullinfo}

In this section, we provide a rigorous statement on the existence of stationary optimal policy in our average-cost continuous-state MDP. Given the intractability of the optimality policy, we propose and establish the asymptotic optimality of base-stock polices under two limiting regimes. The best base-stock policy is then used as the benchmark policy for the regret definition.

\subsection{Preliminaries of the MDP}

\label{subsection:fullinfo_opt}

We establish the existence of a \emph{stationary} optimal repositioning policy following the celebrated vanishing discount approach \citep{schal1993average}. For any discount rate $\rho\in(0,1)$ and initial state $\xv$, the optimal long-run discounted cost function $v_\rho^*(\xv)$  is defined as 
\begin{equation}
\label{eq:discounted_value_function}
v^*_\rho(\xv) := \min_{\policy} \sum_{t=1}^\infty \rho^t \Expt^\policy\left [   C_t(\xv_t, \policy(\xv_t), \dv_t, \Pv_t) \mid \xv_1 = \xv \right].
\end{equation}
The optimality condition under the discounted cost setting is 
\begin{equation}
v^*_\rho(\xv) =  \min_{\yv \in \Delta_{n-1}} \left\{  \Expt_{\dv,\Pv}[C(\xv,\yv,\dv,\Pv)] + \rho \int v^*_{\rho} (\xv')d \Prob(\xv'\mid\xv,\yv) \right\}.
\end{equation}

\begin{theorem}[Existence of Stationary Optimal Policy]
\label{theorem:optimality_exist}
For any $\xv \in \Delta_{n-1}$, the limit 
$\DS
\lambda^* = \lim_{\rho\rightarrow 1} (1-\rho)v^*_\rho(\xv)
$
exists and does not depend on $\xv$. Moreover, there exists a stationary optimal policy $\pi^*$ such that $\lambda^* = \lim_{T\rightarrow \infty}\frac{1}{T}\sum_{t=1}^T\Expt^{\pi^*}[C_t\mid\xv_1= \xv] \text{ for all }\xv \in \Delta_{n-1}$.
\end{theorem}

Clearly, $v^*_\rho(\xv)$ defined in \eqref{eq:discounted_value_function} could be unbounded when $\rho$ goes to $1$, and one cannot simply take $\rho\rightarrow 1$ in $v^*_\rho(\xv)$ to obtain the optimal value function under the average cost setting. Instead, when proving Theorem~\ref{theorem:optimality_exist}, we consider the relative discount function $ r_{\rho}(\xv) := v^*_{\rho}(\xv) -m_\rho$ as $\rho \rightarrow 1$  where $m_{\rho}: = \inf_{\xv \in \Delta_{n-1}} v^*_{\rho}(\xv)$. Then the optimality condition in the discounted cost case can be rewritten as
\begin{equation}
(1-\rho)m_\rho + r_\rho(\xv) =  \min_{\yv \in \Delta_{n-1}} \left\{   \Expt_{\dv,\Pv}[C(\xv,\yv,\dv,\Pv)] + \rho \int r_{\rho} (\xv')d \Prob(\xv'\mid\xv,\yv) \right\}.
\label{eq:discounted_optimality_condition}
\end{equation}

Under appropriate conditions, $r_{\rho}(\xv)$ is finite for all $\rho\in (0,1)$ and the limit of $(1-\rho)m_{\rho}$ is well-defined as $\rho\rightarrow 1$. The main technical challenge lies in identifying the right set of conditions and validating that the conditions hold in our problem context. We focus on the following set of conditions from \citet[Theorem 1]{feinberg2012average}. %\citet[Proposition 1.3]{schal1993average} and 

\begin{definition}[Condition W*]
%The following conditions hold.
\begin{enumerate}[label={(\roman*)}] 
%\addtolength{\itemindent}{2em}
\item The transition probability $\Prob(\cdot\mid\xv,\yv)$ is weakly continuous.  \label{condition_w1}
\item The cost function $c(\xv,\yv): = \Expt_{\dv,\Pv}[C(\xv,\yv,\dv,\Pv)]$ is inf-compact. \label{condition_w2}
\end{enumerate}
\label{condition_w}
\end{definition}
\begin{definition}[Condition B]
%The following conditions hold. 
\begin{enumerate}[label={(\roman*)}] 
%\addtolength{\itemindent}{2em}
    \item $ \inf_{\xv \in \Delta_{n-1}} \inf_\pi \limsup_{T \rightarrow \infty}   \frac{1}{T} \sum_{t=1}^T \Expt^\pi[C_t] < + \infty $. \label{condition_b1}
    \item The relative discount function $\DS r_{\rho}(\xv) := v^*_{\rho}(\xv) -m_{\rho} $ satisfies that $ \sup_{\rho_0\leq \rho <1}r_{\rho}(\xv)<\infty $ for all $\xv \in \Delta_{n-1}$. \label{condition_b2}
\end{enumerate}
\label{condition_b}
\end{definition}

It is particularly non-trivial to verify Condition B\ref{condition_b2}, which we summarize into Proposition \ref{prop:verify_condition_b}. To prove Proposition~\ref{prop:verify_condition_b}, we use a constructive approach to establish the communicating properties of the states in $\Delta_{n-1}$. We can then control the differences in discounted value functions, and thus bound the relative discount function $r_\rho$.

%[Boundedness of Relative Discount Function]
\begin{proposition}
Condition B\ref{condition_b2} holds for our vehicle sharing model, i.e., $ \sup_{\rho_0\leq \rho <1}r_{\rho}(\xv)<\infty $ for all $\xv \in \Delta_{n-1}$.
\label{prop:verify_condition_b}
\end{proposition}

After establishing the existence of the optimal policy, we note that the favorable ``no-repositioning'' property of the optimal policy observed in the discounted-cost setting---which enables efficient computation of discounted cost value function in \cite{benjaafar2018dynamic}---does not extend to the average-cost setting. These computational challenges motivate our development of simple and interpretable policies that maintain practical effectiveness, which we introduce in the subsequent subsection.

\subsection{Base-Stock Repositioning Policy: Asymptotic Optimality Under Two Regimes}
\label{subsec:base_stock_policy}

Due to the intractability of the state-dependent optimal policy, we study a class of base-stock repositioning policies, and the naming is in analogy to the classic base-stock policy in inventory control. A base-stock repositioning policy $\policy^\bslevel$ with a  base-stock level $\bslevel\in \Delta_{n-1}$ repositions the inventory $\xv_t$ to the level $\bslevel$ at each period $t$. Different from typical single-product inventory control where the base-stock level is a single value, the base-stock level in our vehicle sharing model is an $n$-dimensional vector $(s_1,\dots,s_n)$ lying in the set $\Delta_{n-1}$.

\begin{theorem}[Asymptotic Optimality I]
\label{thm:asymptotic_optimal}
Assume that there exists $\alpha_0>0$ such that 
\begin{equation} \label{eq:assump_positive_lost_sales}
\Expt\left[L(\yv, \dv, \Pv)  \right] \geq \alpha_0 \sum_{i,j} l_{ij}\text{ for all } \yv \in \Delta_{n-1}.
\end{equation}
Let $\Gamma:= \sum_{i,j}l_{ij}/\sum_{i,j}c_{ij} $ denotes the ratio between the sum of all lost sales costs and the sum of all repositioning costs. The best base-stock repositioning policy with level $  \bslevel^* $ 

satisfies that
\begin{equation}
1\leq \limsup_{T\rightarrow\infty}  \frac{ \sum_{t=1}^T \Expt^{\policy_{\bslevel^*}}[C_t|\xv_1] }{T\lambda^*   } \leq \frac{1}{1-\alpha_0^{-1} \Gamma^{-1}}.
\end{equation}
Consequently, the base-stock policy $\policy_{\bslevel^*}$ is asymptotically optimal in the following sense, 
\begin{equation*}
\limsup_{T\rightarrow \infty} \frac{ \sum_{t=1}^T \Expt^{\policy_{\bslevel^*}}[C_t\mid\xv_1] }{ T\lambda^*   } =  1+ \Theta( \Gamma^{-1}) \text{ and } \limsup_{\Gamma \rightarrow \infty}  \limsup_{T\rightarrow \infty} \frac{ \sum_{t=1}^T \Expt^{\policy_{\bslevel^*}}[C_t\mid\xv_1] }{ T\lambda^*   } = 1.
\end{equation*}

\end{theorem}
The limiting regime in Theorem~\ref{thm:asymptotic_optimal} corresponds to when the ratio of lost sales cost to repositioning cost is large. Importantly, this ratio is defined at the aggregate level, and we do not require that the ratio $l_{ij}/c_{ij}$ is large for every pair of $i,j$. This limiting regime is particularly relevant when service providers prioritize minimizing user dissatisfaction, and assigning higher costs to lost sales aligns with such objectives, which can also be especially motivated by the need for market growth in competitive environments. A similar limiting regime is established for the classical base-stock policy by the seminal work \citep[Theorem 3]{huh2009asymptotic} where demand is unbounded and the ratio of the lost sales cost and the holding cost goes to infinity.

Theorem~\ref{thm:asymptotic_optimal} is also relevant in a non-asymptotic sense. Assumption \eqref{eq:assump_positive_lost_sales} in Theorem~\ref{thm:asymptotic_optimal} requires that lost sales cost is not negligible for any deterministic base-stock level $\yv$.  Intuitively, $\alpha_0$ in \eqref{eq:assump_positive_lost_sales} represents a minimum probability of demand loss throughout the network.  Considering that the total number of vehicles is fixed and cannot be moved up to an arbitrary inventory level, Assumption \eqref{eq:assump_positive_lost_sales} is a relatively mild assumption in our vehicle sharing model.  Provided that $\alpha_0\Gamma > 1$, the bound in Theorem~\ref{thm:asymptotic_optimal}  gives a valid performance bound on the base-stock policy. 

\begin{theorem}[Asymptotic Optimality II] \label{theorem:asymptotic_optimal_largen}Assume that the demands $\{d_{t,i}\}_{i=1}^n$ are independent and identically distributed across $n$ locations, and there exists a constant $p_0>0$ such that $\DS \Prob\left(d_{t,i} - \Expt[d_{t,i}]> \mathrm{Var}(\theta)\right) \geq p_0 $. Let  $D_t= \sum_{i}d_{t,i}$ denote the total demand across the network, and $\Expt[D_t] = 1$, $\mathrm{Var}(D_t) = \sigma^2$ for some scalar $\sigma>0$, and let $c_{\mathrm{M}} := \max_{i,j} c_{ij}$ and $l_0:=\min_{i,j} l_{ij}>0$. The best base-stock repositioning policy with level $  \bslevel^* $  satisfies that

\begin{equation}
\label{eq:non_asym_largn}
    1 \leq \limsup_{T\rightarrow \infty} \frac{\sum_{t=1}^T \Expt^{\pi^*}[C_t|\xv_1]}{T\lambda^*} \leq \left(1 -  \frac{2c_{\mathrm{M}}}{{\sqrt{n}\sigma l_0 p_0}} \right)^{-1}.
\end{equation}
Consequently, the policy $\pi_{\bslevel^*}$ is asymptotically optimal in the following sense,
\begin{equation*}
\limsup_{T\rightarrow \infty}  \frac{ \sum_{t=1}^T \Expt^{\policy_{\bslevel^*}}[C_t\mid\xv_1] }{T\lambda^* } = 1+ \Theta(n^{-\frac{1}{2}}) \text{ and } \limsup_{n \rightarrow \infty} \limsup_{T\rightarrow \infty}  \frac{ \sum_{t=1}^T \Expt^{\policy_{\bslevel^*}}[C_t\mid\xv_1] }{T\lambda^* } = 1.
\end{equation*}
\end{theorem}
In Theorem~\ref{theorem:asymptotic_optimal_largen}, we show another asymptotic optimality of the base-stock repositioning policy  when the number of locations in the networks goes to infinity, and we have also provided a non-asymptotic bound in \eqref{eq:non_asym_largn}. A similar limiting regime of large network size is considered in \cite{akturk2022managing}, but their analysis is based on a mean-field approximation. The main intuition of proving Theorem~\ref{theorem:asymptotic_optimal_largen} is that managing inventory across $n$ locations is the opposite of ``risk pooling''. Because the system suffers from lost sales cost at each location individually, the aggregate lost sales scales up with the number of locations $n$ even if the variance $\sigma^2$ of the total demand $D_t$ is constant. Theorem~\ref{theorem:asymptotic_optimal_largen} is valuable from the operational perspective because the network with a large number of locations is considerably harder to analyze, yet the simple base-stock repositioning policy can be guaranteed to achieve asymptotic optimality in this limiting regime.

\subsection{Performance Metric of Repositioning Policies}
\label{sec:reg_meas}

\paragraph{Benchmark Policy.} The asymptotic optimality results in Theorems~\ref{thm:asymptotic_optimal} and \ref{theorem:asymptotic_optimal_largen} (Section~\ref{subsec:base_stock_policy}) imply that, although the optimal repositioning policy is intractable, the best base-stock policy is a reliable proxy when lost sales costs dominate or when the number of locations is large. This benchmark aligns with inventory-control results where simple base-stock policies exhibit (asymptotic) optimality (see, e.g., \citet{yuan2021marrying,gong2021bandits,jia2024online}). It also coincides with the standard best fixed policy benchmark in adversarial online learning, as discussed later in Theorem \ref{thm:gdr} of Section \ref{sec:regret_analysis}.

To facilitate discussion, similar to \citet{agrawal2019learning,yuan2021marrying}, we introduce the {modified cost} defined as
$
\widetilde{C}_t(\xv_t,\yv_t,\dv_t,\Pv_t)
= C_t(\xv_t,\yv_t,\dv_t,\Pv_t)-\sum_{i=1}^n \sum_{j=1}^n l_{ij} P_{t,ij} d_{t,i} = M(\yv_t-\xv_t) - \sum_{i=1}^n \sum_{j=1}^n l_{ij} P_{t,ij}\min\{d_{t,i},y_{t,i}\}.
$
Because $\Expt\left[C_t-\widetilde{C}_t\right]=\Expt\left[\sum_{i=1}^n \sum_{j=1}^n l_{ij} P_{t,ij} d_{t,i}\right]$ does not depend on the repositioning policy, replacing $C_t$ by $\widetilde{C}_t$ preserves differences in expected average costs across policies. We therefore conduct the regret analysis using $\widetilde{C}_t$.

\paragraph{Regret Definition.}
Over a horizon $T$, an online algorithm \pcr{ALG} sequentially selects $\yv_t$ based on the current state $\xv_t$ and history $\mathcal{F}_{t-1}$ (censored demand and transition matrices from the previous $t-1$ periods), incurring $\widetilde{C}^{\pcr{ALG}}_t$ at time $t$. Given $\xv_1$, the regret against a fixed base-stock policy $\pi_{\bslevel}$ is
\begin{equation}
\regret(T,\bslevel)
:= \Expt\left[\sum_{t=1}^{T} \widetilde{C}^{\pcr{ALG}}_t \biggm| \xv_1 \right]
- \Expt\left[\sum_{t=1}^{T} \widetilde{C}^{\bslevel}_t \biggm| \xv_1 \right],
\label{eq:regret_T_S_defn}
\end{equation}
where $\widetilde{C}^{\bslevel}_t := \widetilde{C}^{\pi_{\bslevel}}_t$. The worst-case regret relative to the best base-stock level is
\begin{equation}
\regret(T)
:= \Expt\left[\sum_{t=1}^{T} \widetilde{C}^{\pcr{ALG}}_t \biggm| \xv_1 \right]
- \min_{\bslevel \in \Delta_{n-1}}
\Expt\left[\sum_{t=1}^{T} \widetilde{C}^{\bslevel}_t \biggm| \xv_1 \right].
\label{eq:regret_T_defn}
\end{equation}

\begin{remark}
The base-stock vector that minimizes the $T$-horizon objective in \eqref{eq:regret_T_defn} need not equal $\bslevel^*$ from Section~\ref{subsec:base_stock_policy}, which minimizes the infinite-horizon criterion $\loss^{\bslevel}(\xv)$.
%$\loss^{\bslevel}(\xv):=\Expt\left[\lim_{T\to\infty}\frac{1}{T}\sum_{t=1}^{T}\widetilde{C}^{\bslevel}_t \biggm| \xv_{1}=\xv\right]$.
An alternative metric compares \pcr{ALG} with $\bslevel^*$ \citep{jia2024online}, yielding the \emph{pseudoregret}
\begin{equation}
\pregret(T)
:= \Expt\left[\sum_{t=1}^{T} \widetilde{C}^{\pcr{ALG}}_t \biggm| \xv_1 \right]
- T\lambda^{\bslevel^*}.
\label{eq:pseudo_regret_defn}
\end{equation}
As a corollary of our Proposition~\ref{prop:offline_concentration} (Section~\ref{subsec:generalization_bound}), $|\regret(T)-\pregret(T)| \leq \widetilde{O}(\sqrt{T})$. Hence, the two notions are equivalent for learning-rate purposes and we thus focus on regret definition in \eqref{eq:regret_T_defn}.
\end{remark}

%% file: files/offline.tex
\section{Offline Computation of Best Base-Stock Policy}
\label{section:learning}

Before moving to the online learning problem, we discuss a (simpler) problem of offline computing the best base-stock policy. This offline problem turns out to be non-convex, even in the presence of \emph{uncensored} demand data. 

Given an initial inventory level $\xv\in\Delta_{n-1}$ and historical observations $\{(\dv_s,\Pv_s)\}_{s=1}^t$, it is formulated as the following problem over $\bslevel \in \mathbb{R}^n$:
\begin{align}
    \min_{\bslevel\in \Delta_{n-1}} \ & \sum_{s=1}^{t} \reposition(\bslevel-\xv_s) - \sum_{s=1}^{t} \sum_{i=1}^n \sum_{j=1}^n l_{ij} \cdot P_{s,ij} \min\{d_{s,i},S_{i}\}    \label{opt:offline}\\
    \text{subject to} \ & \xv_1=\xv, \ \xv_{s} = (\bslevel - \dv_s)^+  +\Pv_{s-1}^{\top}\min(\bslevel,\dv_{s-1})\text{, for all $s=2,\dots,t$},\label{eq:offline_constr}
\end{align} 
where the repositioning cost $M(\cdot)$ is given by the minimum network cost flow \eqref{eq:repo_cost}. At first glance, \eqref{opt:offline} appears to be a piecewise-linear program: the lost-sales term
$l_{ij} P_{s,ij}\min\{d_{s,i},S_i\}$ is concave in $S_i$, and $M(\cdot)$ is derived from a linear program. However, \eqref{opt:offline} is \emph{non-convex} in $\bslevel$. Eliminating $\xv_s$ via \eqref{eq:offline_constr} rewrites the repositioning input as
$\bslevel-\xv_s=\min(\bslevel,\dv_{s})-\Pv_{s-1}^{\top}\min(\bslevel,\dv_{s-1})$, and the nested $\min(\cdot)$ terms drive the non-convexity. Consequently, solving \eqref{opt:offline} is nontrivial even with uncensored data.

\subsection{Exact Reformulation and Efficient Computation}
\label{subsection:reformulation_milp}

To address the non-convexity, we provide a mixed-integer linear programming (MILP) reformulation (Proposition~\ref{prop:offline_MIP}). The construction may be useful beyond our setting for operations problems with demand censoring. Off-the-shelf solvers handle the formulation effectively, and our small-scale experiments return exact solutions.

We introduce censored-demand variables $\{m_{s,i}\}$ and network-flow variables $\{\xi_{s,ij}\}$, and enforce $m_{s,i}=\min\{d_{s,i},S_i\}$ for all $s,i$ using $nt$ binary variables $\{z_{s,i}\}$. The key step sorts the demand sequence for each $i$ and encodes equality via linear inequalities with these binaries; permutation matrices $\{\boldsymbol{\Gamma}_i\}_{i\in[n]}$ extend the construction to the unsorted case. The approach builds on recent techniques for non-convex piecewise-linear optimization \citep{huchette2022nonconvex} with more details in Appendix~\ref{subsec:proof_reformulation}.

\begin{proposition}[MILP Reformulation]
\label{prop:offline_MIP}
The offline problem \eqref{opt:offline} can be reformulated as a mixed integer linear programming (MILP) problem as follows.
\begin{align}
\label{eq:offline_milp_opt}
    \min_{S_i, m_{s,i}, \xi_{s,ij}, z_{s,i} }\  & \sum\limits_{s=1}^{t}\sum_{i=1}^n \sum_{j=1}^n  c_{ij} \xi_{s,ij}  
    -\sum\limits_{s=1}^{t} \sum\limits_{i=1}^{n}\sum\limits_{j=1}^{n}l_{ij} P_{s,ij}m_{s,i}
    %+\sum\limits_{s=1}^{t} \sum\limits_{i=1}^{n}\sum\limits_{j=1}^{n}l_{ij} P_{s,ij}d_{s,i} 
    \\
    \text{subject to}\ &\sum_{i=1}^n \xi_{s,ij} - \sum_{k=1}^n \xi_{s,jk} = m_{s,j}-\sum\limits_{i=1}^{n} P_{s,ij}m_{s,i}, \text{ for all } j = 1,\dots, n \text{ and $s=1,\dots,t$,}  \nonumber\\
    & \xi_{s,ij} \geq 0, \forall i = 1,\dots, n, \text{ for all } j = 1,\dots, n\text{ and $s=1,\dots,t$,} \nonumber\\
    &\sum\limits_{i=1}^{n} S_i=1, \bslevel=\{S_i\}_{i=1}^{n}\in[0,1]^{n} , \nonumber\\   
    & (m_{1,i}, m_{2,i},\dots,m_{t,i})^\T  = \per_i^\T(\tilde{m}_{1,i}, \tilde{m}_{2,i},\dots,\tilde{m}_{t,i})^\T
    \text{ for all $i=1,..,n$,}   \nonumber\\
    & \per_i (d_{1,i}, d_{2,i},\dots,d_{t,i})^\T  = (\tilde{d}_{1,i}, \tilde{d}_{2,i},\dots,\tilde{d}_{t,i})^\T \text{ for all $i=1,..,n$,}   \nonumber\\
    &\sum \limits_{s=1}^{t} z_{s+1,i}\cdot \tilde{d}_{s,i}\leq  S_i
    \leq \sum\limits_{s=1}^{t} z_{s,i}\cdot \tilde{d}_{s,i}+z_{t+1,i}, \text{ for all $i=1,\dots,n$,} \nonumber\\
    &-2(1-z_{s',i})\leq \tilde{m}_{s,i}-S_i\leq 2(1-z_{s',i}), \text{ for all $1\leq s'\leq s\leq t$ and $i=1,..,n$} \nonumber\\
    &-2(1-z_{s',i})\leq \tilde{m}_{s,i}-\tilde{d}_{s,i}\leq 2(1-z_{s',i}), \text{ for all $1\leq s < s'\leq t+1$ and $i=1,..,n$} \nonumber\\
    &\sum\limits_{s=1}^{t+1} z_{s,i} = 1, \text{ for all $i=1,\dots,n$,} \nonumber\\
    &\zv_{s}=\{z_{s,i}\}_{i=1}^{n}\in\{0,1\}^n, \text{ for all $s=1,\dots,t+1$}, \nonumber
\end{align}
For each $i$, the  permutation matrix $\per_i$ of size $t \times t$ is defined such that the elements in $\per_i d_{:,i}$ are in non-decreasing order, where $d_{:,i} = (d_{1,i}, d_{2,i},\dots,d_{t,i})^\T$ is demand at location $i$ for all times. 
\end{proposition}

The MILP \eqref{eq:offline_milp_opt} has $O(n^2 t+ n t^2)$ constraints and $O(n^2 t)$ variables. While the size scales polynomially in $n$ and $t$, MILPs can be slow for large instances. This trade-off is natural: by recasting a non-convex problem as a MILP, we gain access to mature solvers at the expense of potential computational burden. To identify settings where \eqref{opt:offline} is efficiently solvable, we introduce a mild cost condition, Assumption \ref{assump:lsc>rc}. Several works have adopted equivalent assumptions in the vehicle sharing literature, including \cite{benjaafar2018dynamic} and \cite{he2020robust}. Notably, \citet[Condition 1]{devalve2022base} employs an analogous assumption to prove approximation guarantees in an inventory fulfillment network problem with backlogged demand. Assumption \ref{assump:lsc>rc} corresponds precisely to the limiting regime where the base-stock repositioning policy is optimal in Theorem~\ref{thm:asymptotic_optimal}.

\begin{assumption}[Cost Condition] \label{assump:lsc>rc}
\begin{equation}
 \sum_{i=1}^{n} l_{ji} P_{t,ji} \geq \sum\limits_{i=1}^{n}P_{t,ji}c_{ij}, \text{ for all $j=1,\dots,n$}.
 \label{eq:lsc>rc}
\end{equation}
\end{assumption}

Considering Assumption~\ref{assump:lsc>rc} from a practical perspective, lost sales costs extend beyond trip prices, encompassing opportunity costs from vehicle depreciation during idle periods, customer churn, reduced market presence, and weakened brand loyalty. In contrast, repositioning costs, while including tangible expenses like labor and fuel, can be minimized through operational efficiencies such as task batching and advanced routing algorithms. This aligns with empirical evidences in vehicle sharing systems, such as the real data calibration in \citet[Appendix I.3]{akturk2022managing}. Under Assumption \ref{assump:lsc>rc}, the offline problem \eqref{opt:offline} can be reformulated as a linear program, with details provided in Appendix~\ref{subsec:lp_reformulate}. 

However, it is important to note that Assumption \ref{assump:lsc>rc} still does \emph{not} enable convexity of the cost functions with respect to policy $\bslevel$ in online repositioning. To address this non-convexity challenge in online learning, we introduce \emph{surrogate costs} in Section~\ref{section:ogr} to disentangle intertemporal dependencies in our \pcr{SOAR} algorithm. Without such a cost condition, analysis becomes significantly more challenging, typically requiring approximation methods such as mean-field approximation \citep{akturk2022managing} and fluid approximation \citep{hosseini2021dynamic}. For general cost structures, a Lipschitz bandit-based algorithm in Section~\ref{section:lb_and_dimension} that provides regret guarantees without requiring Assumption \ref{assump:lsc>rc}, albeit with critical dependence on the network size $n$. In Section~\ref{subsec:network_indep}, we introduce a one-time learning algorithm that leverages our MILP reformulation and achieves tight regret guarantees when network demands are independent.

\subsection{Generalization Bound and Lipschitz Property}
\label{subsec:generalization_bound}
The offline solution obtained from \eqref{opt:offline} relies on $t$ observations, and we examine its out-of-sample performance through the lens of generalization error. Proposition \ref{prop:offline_concentration} establishes that for any large $T>t$, with high probability at least $1-3T^{-2}$, the deviation between the $t$-period average cumulative realized cost and the single-period expected cost is uniformly bounded by $O(\sqrt{\log T}/\sqrt{t})$ across all base-stock repositioning policies $\bslevel\in\Delta_{n-1}$. This bound indicates that the generalization error converges uniformly to zero across the policy space $\Delta_{n-1}$ at a squared root rate as the sample size grows.

\begin{proposition}
    \label{prop:offline_concentration}
    Under Assumption~\ref{assumption3}, for any $t\leq T$, \begin{equation*}
        \sup_{\bslevel\in\Delta_{n-1}} \left\vert
            \frac{1}{t}\sum\limits_{s=1}^{t}\widetilde{C}_s(\xv_{s+1}^{\bslevel},\bslevel,\dv_s,\Pv_s)
            -
            \mathbb{E}[\widetilde{C}_1(\xv_1^{\bslevel},\bslevel,\dv_1,\Pv_1)]
        \right\vert
        \leq
        10n^3 \left(\max_{i,j}c_{ij}+\max_{i,j}l_{ij}\right) \cdot \frac{\sqrt{\log T}}{\sqrt{t}}
    \end{equation*}
    holds with probability no less than $1-3T^{-2}$, where $\xv_{s}^{\bslevel}=(\bslevel-\dv_s)^{+}+\Pv_s^{\top}\min\{\bslevel,\dv_s\}$ for all $s\geq 1$.
\end{proposition}
To Proposition \ref{prop:offline_concentration}, we also establish a Lipschitz property of the cost function with respect to $\bslevel$ in Lemma~\ref{lem:h_lipschitz}. To facilitate the exposition, we introduce simplified notation that is used repeatedly throughout our concentration analysis. Let $\fv_{\bslevel}:\mathbb{R}^{n}\times\mathbb{R}^{n\times n} \rightarrow \mathbb{R}^{n}\times\mathbb{R}^{n\times n}$ be a vector-valued function 
$\fv_{\bslevel}(\dv,\Pv):=(\min(\dv,\bslevel),\Pv) $ defined on $\left\{(\dv,\Pv):\dv\in\Delta_{n-1},\Pv\in\mathbb{R}^{n\times n}\right\}$ for any $\bslevel\in\Delta_{n-1}$, and let $h:\mathbb{R}^{n}\times\mathbb{R}^{n\times n}\rightarrow\mathbb{R}$ be the cost function
\begin{equation}
    \label{def:h}
    h(\yv,\Pv) =
    M\left(\yv-\Pv^{\top}\yv\right) - \sum\limits_{i=1}^{n}\sum\limits_{j=1}^{n}l_{ij}\cdot P_{ij}y_i
\end{equation}
defined on $\left\{(\yv,\Pv):\yv\in[0,1]^n, \Pv\in[0,1]^{n\times n}, \Pv\boldsymbol{1}=\boldsymbol{1} \right\}$.  The introduced mappings $\fv$ and $h$ enable us to leverage the following vector-contraction inequality  in Lemma~\ref{lem:contraction} \citep[Corollary 1]{maurer2016vector} to bound the Rademacher complexity.

\begin{lemma}
%[Vector-Contraction Inequality]
    \label{lem:contraction}
    Let $\mathcal{X}$ be any set, $(x_1, \dots, x_t) \in \mathcal{X}^t$, let $\mathcal{F}$ be a class of functions $\fv: \mathcal{X} \rightarrow\mathbb{R}^n$ and let $h_i: \mathbb{R}^n\rightarrow \mathbb{R}$ have Lipschitz norm $L$. Then
    \begin{align*}
        \mathbb{E}\left[
            \sup_{\fv\in\mathcal{F}}
            \sum\limits_{s=1}^{t} \sigma_s h_s(\fv(x_s))
        \right]
        \leq
        \sqrt{2}L\mathbb{E}\left[
            \sup_{\fv\in\mathcal{F}}
            \sum\limits_{s=1}^{t} \sum\limits_{k=1}^{n} \sigma_{s,k} f_k(x_s)
        \right],
    \end{align*}
    where $\sigma_s$ and $\{\sigma_{s,k}\}_{k=1}^{n}$ are independent uniform distributions on $\{-1,1\}$ for all $s=1,...,t$, and $f_k(\cdot)$ is $k-th$ component of $\fv(\cdot)$.
\end{lemma}

The contraction inequality in Lemma~\ref{lem:contraction} is a generalization of the well-known Talagrand's lemma, which can be viewed as a scalar version of this contraction lemma. 

%We refer to Corollary 1 in \cite{maurer2016vector} for the proof.

\begin{lemma}
%[Lipschitz Property] 
For any $\yv, \yv'\in[0,1]^{n}$, and probability transition matrices $\Pv,\Pv'\in[0,1]^{n\times n}$, it holds that
\begin{equation*}
        \left\vert
            h(\yv, \Pv)-h(\yv', \Pv')
        \right\vert
        \leq
        n^2\cdot (\max_{i,j}c_{ij}+\max_{i,j}l_{ij}) \cdot(\|\yv-\yv'\|_2+\|\Pv-\Pv'\|_{F}).  \label{lem:h_lipschitz}
    \end{equation*}
\end{lemma}
In proving Proposition~\ref{prop:offline_concentration}, we build on the Lipschitz reduction (Lemma~\ref{lem:h_lipschitz}) and the vector-contraction inequality (Lemma~\ref{lem:contraction}), and then applies symmetrization via a Rademacher-complexity bound together with a generalized Massart finite-class estimate (Lemmas~\ref{lem:rademacher_complexity} and \ref{lem:massart}) to bound the error uniformly over $\bslevel\in\Delta_{n-1}$.

%% file: files/ogd_regret_analysis.tex
\section{Online Repositioning with Tight Regret Guarantee}
\label{section:ogr}
In this section we introduce our Surrogate Optimization and Adaptive Repositioning (\pcr{SOAR}) algorithm (Algorithm~\ref{alg_nloc:ogr}). A core idea is to replace the true period costs with a sequence of \emph{surrogate cost} that decouple the intertemporal dependencies induced by inventory flows. At each iteration, we solve a tailored linear program whose dual variables, together with censor indicators, are used to construct a subgradient of the surrogate cost, enabling principled first-order updates of the repositioning targets. The procedure requires minimal data, is computationally light, and comes with strong performance guarantees. In particular, the regret bound for \pcr{SOAR} holds under adversarial demand sequences and does not rely on Assumption~\ref{assumption3}.

\subsection{Learning Challenges and Algorithm Design}
\label{sec:alg_des}
The goal of learning while repositioning is to sequentially choose repositioning levels when only censored network demand is observed. Three features make this problem particularly challenging: (i) the fleet operates in a closed network with a fixed total inventory, so exploration via overstocking is infeasible; (ii) mobility intrinsically couples locations, precluding regional control or location-wise decomposition; and (iii) censoring biases naive estimators and complicates policy evaluation. By contrast, \pcr{SOAR} leverages surrogate costs and LP-based subgradients to update repositioning levels adaptively for the whole network. 

The \emph{non-convexity} in our problem stems from the multi-dimensional decision variables intertwined with demand censoring, which distinguishes it from the non-convexity caused by lead time or fixed costs in the existing literature. Consequently, the approaches to addressing non-convexity in previous works \citep{yuan2021marrying,chen2023learning} are not directly applicable here. Furthermore, due to the correlation across different dimensions, the idea of convex reformulation via variable transformation \citep{chen2022efficient} is also not applicable. Instead, we introduce a novel ``disentangling" idea to achieve convexity in newly defined surrogate costs  in Section~\ref{section:disentangling}, which approximates the original cost objectives well under certain algorithm designs.

While gradient-based approaches have proven effective for adjusting base-stock levels in inventory control (see, e.g., \citet{huh2009nonparametric,yuan2021marrying,lyu2023minibatch}), the network structure and $n$-dimensional gradient in our problem present unique calibration challenges. The subgradient in \pcr{SOAR} is defined through the dual solution of a linear program that encodes the minimum cost flow problem governing inventory repositioning across the network. The validity of such a dual solution gradient is enabled by the surrogate costs that not only approximate the original modified costs well but also exhibit favorable analytical properties. 

Specifically, we demonstrate that the gap between surrogate costs and the original can be bounded in an instance-based fashion by the cumulative changes of policy updates. This gap remains well bounded when the policy updates follow a ``slow-moving'' recommendation, as proved in Lemma~\ref{lem:disentangle}, which also aligns with the step size choice in gradient descent approach. Another challenge stems from constructing linear program and dual solution solely based on censored demand $\min(\dv_t, \yv_t)$, and we exploit the censored structure to recover the true subgradient with respect to $\yv_t$, as proved in Lemma~\ref{lem:gd}.

\begin{algorithm}[!htb]
\caption{\pcr{SOAR}: Surrogate Optimization and Adaptive Repositioning Algorithm}
\label{alg_nloc:ogr}
\begin{algorithmic}[1]
\State \textbf{Input:} Number of iterations $T$, initial repositioning policy $\yv_1$;
\For{$t =1,..., T-1$}
    \Statex \hspace{1em} \emph{\color{Gray} \% Collect censored data}
    \State Set the target inventory be $\yv_t$ and observe realized censored demand $\dv_t^{c}=\min(\yv_t,\dv_t)$; \label{alg_step:set}
    \Statex \hspace{1em} \emph{\color{Gray} \% Solve linear programming involving surrogate costs}
    \State Denote $\boldsymbol{\lambda}_t=(\lambda_{t,1},\dots,\lambda_{t,n})^{\top}$ be the optimal dual solution corresponding to constraints  \eqref{ieq:S_gd} 
    \begin{align}
        \widetilde{C}_t(\xv_{t+1},\yv_t,\dv_t,\Pv_t)=\min\  & \sum_{i=1}^n \sum_{j=1}^n  c_{ij} \xi_{t,ij}  
            - \sum\limits_{i=1}^{n}\sum\limits_{j=1}^{n}l_{ij} P_{t,ij}w_{t,i}\label{opt:one_step}\\
            \mathrm{subject\  to}\ &\sum_{i=1}^n \xi_{t,ij} - \sum_{k=1}^n \xi_{t,jk} = w_{t,j}-\sum\limits_{i=1}^{n} P_{t,ij}w_{t,i}, \mathrm{\  for\  all \ } j = 1,\dots, n,\nonumber\\
            & w_{t,i}\geq 0,\ \xi_{t,ij} \geq 0,\  \mathrm{ for\  all \ } i,j = 1,\dots, n,\nonumber \\
            &w_{t,i}\leq (\dv^c_{t})_i,\ \mathrm{\  for\  all\ } i=1,\dots,n, \label{ieq:S_gd}
    \end{align}
     \Statex \hspace{1em} where $\boldsymbol{\xi_t}=\{\xi_{t,ij}\}_{i,j=1}^{n}$ represent network flows and $\boldsymbol{w}_t=\{w_{t,i}\}_{i=1}^{n}$ are auxiliary variable; \label{alg_step:solve_dual}
    \Statex \hspace{1em} \emph{\color{Gray} \% Construct subgradient from dual solution}
    \State Let
    \[g_{t,i} = \lambda_{t,i} \cdot \mathbbm{1}{\left\{(\dv^c_{t})_i=y_{t,i}\right\}}\text{, \ for all $i\in[n]$},\]
    and define the subgradient as $\gv_t=(g_{t,1},...,g_{t,n})^{\top}$;  \label{alg_step:compute_gradient}
      \Statex \hspace{1em} \emph{\color{Gray} \% Adaptively update inventory level using subgradient}
    \State Update the repositioning policy $\yv_{t+1}=\Pi_{\Delta_{n-1}}\left(\yv_{t}-\frac{1}{\sqrt{t}}\gv_t\right)$; \label{alg_step:update_policy}
\EndFor
\State \textbf{Output:} $\left\{\yv_t\right\}_{t=1}^{T}$.
\end{algorithmic}
\end{algorithm}

\paragraph{Core Ideas of Algorithm \ref{alg_nloc:ogr}.} Within each iteration of Algorithm \ref{alg_nloc:ogr}, Steps \ref{alg_step:set}--\ref{alg_step:compute_gradient} calculate a subgradient of the modified cost $\tilde{C}_t(\xv_{t+1},\yv_t,\dv_t,\Pv_t)$ with respect to $\yv_t$ for each time $t$. The most intricate part of designing Algorithm \ref{alg_nloc:ogr} is identifying the gradient of the surrogate cost function introduced in Section~\ref{section:disentangling}, which we define as the dual of a linear program, and will discuss in more detail in Section \ref{sec:compute_g}. We note that the gradient is non-positive due to the constraint \eqref{ieq:S_gd} in the minimization problem. For any non-zero element $g_{t,i}$ of the gradient, it holds that $(\dv^c_{t})_i=y_{t,i}$, which means that demand might not be completely fulfilled at location $i$. In this case, Step \ref{alg_step:update_policy} will increase the supply correspondingly. The smaller the element $g_{t,i}$ is, the more cost reduction can potentially be brought from increasing inventory at location $i$. Therefore, the gradient descent step has a very nice intuition of ranking the ``priority'' of all the locations in the repositioning operation. Step \ref{alg_step:update_policy} updates the repositioning policy by moving along the direction of the gradient with a small step size ${1}/{\sqrt{t}}$ for all $t=1,\dots,T$ followed by projection onto the feasible space of simplex $\Delta_{n-1}$. The small step size not only helps with algorithm convergence but also guarantees a small approximation error with the surrogate costs, which we will discuss in more detail in Section~\ref{section:disentangling}. It is noteworthy that Algorithm \ref{alg_nloc:ogr} possesses three significant advantages. \begin{itemize}
    \item[(i)] {\bf Minimal Data Requirement.} This online gradient algorithm is applicable by \emph{only} accessing censored data. Particularly, as shown in Steps \ref{alg_step:solve_dual} and \ref{alg_step:compute_gradient}, all the local gradient $\gv_t$ can be obtained with censored demand $\dv_t^{c}$ for all $t$. This weak requirement on data accessibility enables this algorithm to be applied flexibly in environments with limited data availability, and practically speaking,  the service provider would not need to aggressively increase the supply in order to learn the uncensored demand. 
    \item[(ii)] {\bf Computational Efficiency.} Algorithm \ref{alg_nloc:ogr} is computationally efficient at each step throughout all time periods. At each period, Algorithm \ref{alg_nloc:ogr} only computes one linear program with $O(n^2)$ constraints and variables in Step \ref{alg_step:solve_dual} and updates the gradient in Steps \ref{alg_step:compute_gradient} and \ref{alg_step:update_policy}. The corresponding computational complexity is polynomial in the number of locations in the network, yet it remains independent of the time horizon, denoted as $T$.  Such computational efficiency enables rapid adaptation to changes in realized demands across the network. 
    \item[(iii)] {\bf Reliability.} In Section \ref{sec:regret_analysis}, we will see that this algorithm achieves an ${O}(n^{2.5}\sqrt{T})$ regret guarantee with either i.i.d. or adversarial demands and transition probabilities. This theoretical guarantee illustrates the robustness and reliability of this algorithm against any distribution shifts of the demand levels and transition probabilities. 
\end{itemize}

\subsection{Disentangling Dependency via Surrogate Costs.}
\label{section:disentangling}
\paragraph{Twisted Dependency and Non-Convexity.} A key obstacle in optimizing the cumulative modified costs comes from the \emph{twisted dependency} of repositioning policies on the modified costs.  Specifically, the minimization objective of the cumulative modified cost is given by
\begin{equation}
    \label{eq:modi_total_pseudo_cost}
    \sum\limits_{t=1}^{T}\tilde{C}_t(\xv_{t}(\yv_{t-1}),\yv_t,\dv_t,\Pv_t),
\end{equation}
where $\xv_{t+1}=(\yv_{t}-\dv_{t})^{+}+\Pv_{t}\min\{\yv_{t},\dv_{t}\}$ for all $t=1,\dots,T$. In this subsection, with a slight abuse of notation, we will use $\xv_{t+1}(\yv_t)$ and $\xv_{t+1}$ interchangeably to emphasize the dependency between $\xv_{t+1}$ and $\yv_t$ when needed. We note that  $\tilde{C}_t(\xv_t(\yv_{t-1}),\yv_t,\dv_t,\Pv_t)$ depends on the repositioning policies, demands, and origin-to-destination probability at both time $t-1$ through $\xv_t$ and those at time $t$, for all $t=1,\dots,T$. Furthermore, due to the dependence of $\xv_t$ on $\yv_{t-1}$, the cost $\tilde{C}_t(\xv_{t}(\yv_{t-1}),\yv_t,\dv_t,\Pv_t)$ is non-convex in $\bslevel$ even when Assumption~\ref{assump:lsc>rc} holds and $\yv_s = \bslevel$ for all $s=1,\dots,t$  (see the discussion on non-convexity in Section~\ref{section:learning}). This twisted dependency prevents one from solving \eqref{eq:modi_total_pseudo_cost} by applying online gradient-based methods \citep{hazan2022introduction}. 
%, such as mirror descent or standard proximal point method 

\paragraph{Surrogate Costs.}  To remove this obstacle, we propose to disentangle the twisted dependency by considering ``relabeled'' cumulative modified costs. In Lemma \ref{lem:disentangle}, we show that the relabeled cumulative modified cost
\begin{equation}
    \label{eq:relabel_modi_total_pseudo_cost}
    \sum\limits_{t=1}^{T}\tilde{C}_t(\xv_{t+1}(\yv_t),\yv_t,\dv_t,\Pv_t)
\end{equation}
is a disentangled surrogate to \eqref{eq:modi_total_pseudo_cost} with an approximation error $O\left(\sum\limits_{t=1}^{T}\|\yv_{t}-\yv_{t-1}\|_1\right)$, where terms in \eqref{eq:relabel_modi_total_pseudo_cost} depend on separate input variables compared to the original modified cost \eqref{eq:modi_total_pseudo_cost}.
\begin{lemma}
%[Disentangling]
    \label{lem:disentangle}
    Let $\{\yv_t\}_{t=1}^{T}\subseteq\Delta_{n-1}$ be any sequence of repositioning policies. Then, the relabeled modified cost $\tilde{C}_t(\xv_{t+1}(\yv_t),\yv_t,\dv_t,\Pv_t)$ depends only on the repositioning policy and realized demands and transition matrix at time $t$, for all $t=1,\dots,T$. Here, $\xv_{t+1} = (\yv_{t}-\dv_{t})^{+}+\Pv_{t}\min\{\yv_{t},\dv_{t}\}$ for all $t=1,\dots,T$.
    
    Furthermore, the gap between the cumulative modified cost and the cumulative relabeled modified cost can be bounded by the following inequality where $\yv_0:=\xv_1$,
    \begin{align}
        \label{eq:disentangle}
        \left|\sum\limits_{t=1}^{T}\tilde{C}_t(\xv_t,\yv_t,\dv_t,\Pv_t)
        -
        \sum\limits_{t=1}^{T}\tilde{C}_t(\xv_{t+1},\yv_t,\dv_t,\Pv_t)\right|
        \leq
        2\cdot\left(\max_{i,j=1,\dots,n} c_{ij}\right)\cdot\sum\limits_{t=2}^{T}\|\yv_{t}-\yv_{t-1}\|_1.
    \end{align}
\end{lemma}
% \begin{remark} 
% Lemma \ref{lem:disentangle} still holds even if Assumption \ref{assump:lsc>rc} is not true, so this ``disentangling'' technique is potentially useful for solving more general repositioning problems. 
% \end{remark}
\begin{remark}
Lemma \ref{lem:disentangle} indicates that the approximation error of this surrogate cost is controllable, provided that the repositioning policies are \emph{updated slowly}. In particular, the total approximation is bounded by $O(\sqrt{T})$ if one always slightly changes the repositioning policies, e.g.,  $\|\yv_{t+1}-\yv_t\|_1=O(1/\sqrt{t})$ for all $t$, or only updates the policies infrequently, for example, when $\yv_{t+1}\not=\yv_{t}$ holds for at most $O(\sqrt{T})$ times. This insight also coincides with the choice of the step size $O(1/\sqrt{t})$ in Algorithm~\ref{alg_nloc:ogr}.
\end{remark}
\begin{remark} 
Beyond resolving the twisted dependency, it is remarkable that this surrogate cost also helps to circumvent the \emph{non-convexity} challenge. Specifically, it is shown in Lemma~\ref{lem:gd} that $\tilde{C}_t(\xv_{t+1},\yv_t,\dv_t,\Pv_t)$ is a convex function with respect to the corresponding repositioning policy $\yv_t$ for all $t$. 
\end{remark}

\subsection{Construction of the Subgradient Vector} 
\label{sec:compute_g} 
The correctness of Algorithm \ref{alg_nloc:ogr} hinges on the validity of $\gv_t$ as a subgradient, which we formally establish in Lemma \ref{lem:gd} below.
\begin{lemma}[Validity of Subgradient]
    \label{lem:gd}
         Under Assumption \ref{assump:lsc>rc}, given any demand vector $\dv_t$ and origin-to-destination probability $\Pv_t$, surrogate costs $\widetilde{C}_{t}(\xv_{t+1}(\yv_t),\yv_t,\dv_t,\Pv_t)$ introduced in \eqref{eq:relabel_modi_total_pseudo_cost} is a convex function with respect to $\yv_t$ for all $t=1,\dots,T$. 
        
        Furthermore,  $\gv_t$ in Step \ref{alg_step:compute_gradient} of Algorithm \ref{alg_nloc:ogr} is a subgradient of $\widetilde{C}_{t}(\xv_{t+1}(\yv_t),\yv_t,\dv_t,\Pv_t)$ for all $t=1,\dots,T$. 
\end{lemma}

To prove Lemma~\ref{lem:gd} (in Appendix \ref{appendix:pf_lem_gd}),  we consider the following LP \eqref{opt:eqilLP}.
\begin{align}
    \text{LP(}\yv_t) = \min_{\xi_{t,ij},w_{t,i}}  & \sum_{i=1}^n \sum_{j=1}^n  c_{ij} \xi_{t,ij}  
            - \sum\limits_{i=1}^{n}\sum\limits_{j=1}^{n}l_{ij} P_{t,ij}w_{t,i} \label{opt:eqilLP}\\
            \mathrm{subject\  to}\ &\sum_{i=1}^n \xi_{t,ij} - \sum_{k=1}^n \xi_{t,jk} = w_{t,j}-\sum\limits_{i=1}^{n} P_{t,ij}w_{t,i}, \mathrm{\  for\  all \ } j = 1,\dots, n, \label{const:flow} \\
            &w_{t,i}\leq y_{t,i}, \ \mathrm{\  for\  all\ } i=1,\dots,n,\label{const:yt}\\
            & w_{t,i}\leq d_{t,i}, \ \mathrm{\  for\  all\ } i=1,\dots,n,\label{const:dt}\\
             & w_{t,i}\geq 0,\ \xi_{t,ij} \geq 0,\  \mathrm{ for\  all \ } i,j = 1,\dots, n.\nonumber 
\end{align}

LP \eqref{opt:eqilLP} shares the same optimal objective value as the original problem  because the non-linear censoring constraint $w_{t,i}= \min(y_{t,i}, d_{t,i})$ is superseded by the combination of \eqref{const:yt} and \eqref{const:dt} under Assumption \ref{assump:lsc>rc}. Therefore, it suffices to show that $\gv_t$ defined in Algorithm \ref{alg_nloc:ogr} is the gradient of LP \eqref{opt:eqilLP} with respect to $\yv_t$ for all $t$. Let $\boldsymbol{\mu}_t$ and $\boldsymbol{\eta}_t$ denote the dual variables, or Lagrangian multipliers, corresponding to constraints \eqref{const:yt} and \eqref{const:dt}, respectively, and let $\boldsymbol{\pi}_{t}$ denote the dual variable corresponding to constraint \eqref{const:flow}. By optimality of $(\boldsymbol{\mu}_t, \boldsymbol{\eta}_t)$ and strong duality, we have
\begin{equation}
\begin{split}
    \text{D-LP}(\yv_t') - \text{D-LP}(\yv_t)
    &\geq
    \boldsymbol{\mu}_t^{\top}\yv_t'+\boldsymbol{\eta}_t^{\top}\dv_t - \text{D-LP}(\yv_t)\\
    % &=
    % \boldsymbol{\mu}_t^{\top}\yv_t'+\boldsymbol{\eta}_t^{\top}\dv_t - (\boldsymbol{\mu}_t^{\top}\yv_t+\boldsymbol{\eta}_t^{\top}\dv_t)
    % \\
    &=
    \boldsymbol{\mu}_t^{\top}(\yv_t'-\yv_t).
\end{split}
\label{ieq:subg}
\end{equation}
% where the first inequality comes from the feasibility of $\boldsymbol{\mu}_t$ and $\boldsymbol{\eta}_t$ to D-LP$(\yv_t')$ and the maximality of the objective value of this dual problem, the second equality comes from the strong duality of LP$(\yv_t)$, and the last equality is by direct calculation. 

It follows from \eqref{ieq:subg} that any dual optimal solution $\boldsymbol{\mu}_t$ is a subgradient of \eqref{opt:eqilLP} with respect to $\yv_t$. This subgradient, derived from a principled dual argument, also provides clear operational intuition. For any $i$, if $\mu_{t,i} = g_{t,i} = \lambda_{t,i} \cdot \mathbbm{1}{\left\{(\dv^c_{t})_i=y_{t,i}\right\}} < 0$, i.e., the constraint $w_{t,i}\leq y_{t,i}$ in \eqref{const:yt} is binding, it means that location $i$ is in a relative deficit of inventory. Consequently, the subgradient update step , $y_{t,i} - \frac{1}{\sqrt{t}}g_{t,i}$, of Algorithm \ref{alg_nloc:ogr}correctly increases the target inventory level at that deficit location $i$ to better meet future demand.

\subsection{Tight Regret Guarantee Beyond i.i.d. Assumption}
\label{sec:regret_analysis}

We present the theoretical guarantee of Algorithm~\ref{alg_nloc:ogr} in Theorem~\ref{thm:gdr}. 
\begin{theorem}
    \label{thm:gdr}
    Under only Assumption \ref{assump:lsc>rc}, the output of Algorithm \ref{alg_nloc:ogr} satisfies
    \begin{equation}
        \label{ieq:regret_noniid}
        \sum\limits_{t=1}^{T}\widetilde{C}_{t}(\xv_t(\bslevel_{t-1}),\bslevel_{t},\dv_t,\Pv_t)
       - \min_{\bslevel\in\Delta_{n-1}} \sum\limits_{t=1}^{T}\widetilde{C}_{t}(\xv_t(\bslevel),\bslevel,\dv_t,\Pv_t)\leq O(n^{2.5}\cdot\sqrt{T})
    \end{equation}
    for any initial inventory level $\bslevel_0:=\bslevel_1\in\Delta_{n-1}$ and any sequence of demand and origin-to-destination probability pairs $\left\{(\dv_t,\Pv_t)\right\}_{t=1}^{T}$. 
\end{theorem}
The bound in Theorem~\ref{thm:gdr} is optimal in $T$ and holds under only Assumption~\ref{assump:lsc>rc}, without requiring i.i.d. or network independence assumptions. The phrase ``any sequence'' indicates that each demand and origin-to-destination probability pair $(\dv_t, \Pv_t)$ can be chosen adversarially at period $t$ to work against the algorithm. Moreover, $\left\{(\dv_t,\Pv_t)\right\}_{t=1}^{T}$ need not be i.i.d. or exogenous, and may be correlated with both historical and current repositioning policies $\{\bslevel\}_{s=1}^{t}$. We present a natural corollary under i.i.d. assumption in Corollary~\ref{cor:gdr}.
\begin{corollary}    \label{cor:gdr}
    Under the same condition of Theorem~\ref{thm:gdr}, if Assumption \ref{assumption3} also holds, we have
    \begin{equation}
        \label{ieq:regret_iid}
        \mathbb{E}\left[ \sum\limits_{t=1}^{T}\widetilde{C}_{t}(\xv_t(\bslevel_{t-1}),\bslevel_{t},\dv_t,\Pv_t)
        \right]
        - \min_{\bslevel\in\Delta_{n-1}} T\mathbb{E}\left[ \widetilde{C}_{1}(\xv_1(\bslevel),\bslevel,\dv_1,\Pv_1)\right] \leq O(n^{2.5}\cdot\sqrt{T}).
    \end{equation}
\end{corollary}
\begin{remark} 
Regarding the network size $n$, our analysis shows that Algorithm~\ref{alg_nloc:ogr}'s regret bound has a polynomial dependence on $n$. This represents a substantial improvement over the Lipschitz-bandit approach, which has a regret guarantee of $\widetilde{O}(T^{\frac{n}{n+1}})$. The lower bound of $\Omega(n \sqrt{T})$ established in Theorem~\ref{theorem:sqrt_t_lb} proves that some polynomial dependence on $n$ is inevitable. A direction for future research is to determine whether the current polynomial dependence on $n$ can be further refined.
\end{remark}

We provide a sketch of regret analysis below and leave detailed proof to Appendix~\ref{sec:appd_gd}. 
% \paragraph{Sketch of Regret Analysis {\normalfont (Detailed proof in Appendix~\ref{sec:appd_gd})}.} 
A key proof intuition is that, Algorithm \ref{alg_nloc:ogr} introduces noise in updating the repositioning policies through noised subgradients and a slow-decaying stepsize at Steps \ref{alg_step:compute_gradient} and \ref{alg_step:update_policy}. The introduced noise enables the algorithm to explore the decision space efficiently, to cancel out decision errors over time, and thus, to mitigate cumulative costs for adversarial inputs.  
Based on Lemma \ref{lem:disentangle}, we could invoke the convergence rate of the projected online gradient descent algorithm (Lemma~\ref{lem:ogd}) to obtain a regret bound on the cumulative \emph{surrogate costs}.
\begin{equation}
R_1 = 6n^2\left(\max_{i,j}c_{ij}+\max_{i,j}l_{ij}\right)\cdot \sqrt{T}. 
\label{eq:r1}
\end{equation}
Due to the bound in \eqref{eq:disentangle} of Lemma \ref{lem:disentangle}, we could control the approximation error of using surrogate costs by
\[
R_2 = \left(\max_{ij} c_{ij}\right)\|\bslevel_{t-1}-\bslevel_{t}\|_1.
\]
Since the step size is $1/\sqrt{t}$, we can use bound the $\ell_1$ difference $\|\bslevel_{t-1}-\bslevel_{t}\|_1$ by $2\sqrt{n}/\sqrt{t} \|fv_t\|_2$. On the other hand, by the Lipschitz property in Lemma~\ref{lem:h_lipschitz}, the subgradient norms can be bounded by $\Vert \gv\Vert_2 \leq n^2(\max_{i,j} c_{ij}+\max_{i,j} l_{ij})$. It follows that 
\begin{equation}
R_2 \leq 2n^{5/2} \left(\max_{i,j}c_{ij}+\max_{i,j}l_{ij}\right) \sum_{t=1}^T 1/\sqrt{t} \leq 4n^{5/2}\sqrt{T} \left(\max_{i,j}c_{ij}+\max_{i,j}l_{ij}\right).
\label{eq:r2}
\end{equation}
Putting \eqref{eq:r1} and \eqref{eq:r2} together, the cumulative regret is bounded by 
\[
R_1 + R_2 \leq (6n^2+4n^{5/2})\sqrt{T} \left(\max_{i,j}c_{ij}+\max_{i,j}l_{ij}\right)^2 \in O(n^{2.5} \sqrt{T}).
\]

%% file: files/extension.tex
\section{Discussion and Extension}
\label{section:further_and_extension}

Throughout the online learning analysis, we have emphasized how our \pcr{SOAR} algorithm addresses the dual challenges of demand censoring and spatial correlation. After discussing the regret lower bound and dimensionality challenge in Section~\ref{section:lb_and_dimension}, we propose two simple algorithms and bound their regret when either of the two challenges is relaxed in Section~\ref{subsec:counterexample} and \ref{subsec:network_indep}, respectively. Furthermore, in Section~\ref{subsection:extension}, we extend our model to accommodate more complex relationships between review periods and rental periods and demonstrate how our \pcr{SOAR} approach naturally generalizes to this scenario.

\subsection{Lower Bound and Challenge of Dimensionality}
\label{section:lb_and_dimension}

The regret lower bound in Theorem~\ref{theorem:sqrt_t_lb} matches with the regret upper bound of \pcr{SOAR} in Theorem~\ref{thm:gdr}, and thus proves the optimality of the \pcr{SOAR} algorithm. We derive the lower bound based on results on stochastic linear optimization under bandit feedback \citep{dani2008stochastic}.  

%The proof of Theorem~\ref{theorem:sqrt_t_lb} is presented in Appendix~\ref{subsection:proof_lb}. 

\begin{theorem}[Regret Lower Bound]
\label{theorem:sqrt_t_lb}
Given time horizon $T$, for any online learning algorithm \pcr{ALG} for the vehicle repositioning problem with cost structure satisfying Assumption~\ref{assump:lsc>rc}, the worst-case expected regret is at least $\Omega(n\sqrt{T})$.
\end{theorem}

%\todo{get a better lower bound to capture its dependence on $n$}

Interestingly, we can conclude that by assuming that the cost structure following Assumption~\ref{assump:lsc>rc}, we can effectively avoid the curse of dimensionality and obtain a regret bound that does not depend on $n$ in the power of $T$ through \pcr{SOAR}. One may wonder, what if Assumption~\ref{assump:lsc>rc} does not hold? We continue the discussion by noting that a Lipschitz Bandit-based Repositioning algorithm can achieve a regret bound of $\widetilde{O}( T^{\frac{n}{n+1}})$ by adapting the analysis of \cite{agrawal2019learning}. The key difference is that, in the absence of convexity, one cannot invoke online convex optimization as in \cite{agrawal2019learning}; the argument instead relies on leveraging the Lipschitz property (Lemma~\ref{lem:h_lipschitz}), and combining the covering number ($\sim \delta^{-(n-1)}$ for granularity level $\delta$) of the policy simplex $\Delta_{n-1}$ with the regret analysis of Lipschitz bandits \citep{kleinberg2008multi}.

% H =  2(6\max_{i,j} c_{ij} + 2nU\max_{i,j} l_{ij})
\begin{theorem} %[Curse of Dimensionality]
\label{ref:thm_reg}
The Lipschitz Bandit-based Repositioning algorithm's regret is upper-bounded by ${\bigo}\left(n \log T \cdot T^{\frac{n}{n+1}}\right)$.
\end{theorem}

Naturally, Theorem \ref{theorem:sqrt_t_lb} is also a lower bound for the general network, but we are not aware if a stronger lower bound $\Omega(T^{\frac{n}{n+1}})$ can be proved for the vehicle sharing problem where Assumption~\ref{assump:lsc>rc} does not hold. It is worth mentioning that the lower bound $\Omega(T^{\frac{D+1}{D+2}})$ exists for general Lipschitz bandits over a space with covering dimension $D$. In our problem, when we set repositioning cost to be $0$, then the cost function at time $t$ is a specific Lipschitz function $L(\dv,\bslevel_t)$ where $\bslevel_t$  is the base-stock repositioning policy selected at time $t$. Moreover, the covering dimension of $\Delta_{n-1}$ under $\ell_1$ norm is $n-1$. Therefore, by plugging $D = n-1$ into $\Omega(T^{\frac{D+1}{D+2}})$, we obtain a lower bound $\Omega(T^{\frac{n}{n+1}})$, which matches the proved regret upper bound $\widetilde{O}( T^{\frac{n}{n+1}})$ up to multiplicative logarithmic factors. However, this intuition does not directly translate into a rigorous proof because the instances used to achieve to worst case regret lower bound of Lipschitz bandits are a class of ``bump'' functions \citep{kleinberg2008multi}, which do not belong to the class of functions in the form of $L(\dv,\bslevel_t)$. We leave this as an interesting {open problem} for future exploration. If true, this will serve as a direct measure of the inherent complexity of the vehicle repositioning problem without additional structure.

\subsection{Challenge of Censored Data in Network}
\label{subsec:counterexample}

In the following Proposition~\ref{lem:infeas}, we formalize the inherent challenge incurred by demand censoring into a concrete example. We show that it is impossible to identify the true ground distribution of demand by merely observing the censored demand data, even when the dimension is only $2$. 
\begin{proposition}[A Pessimistic Example]
        There exists a set of two-dimensional joint distribution $\mathcal{P}$  such that for any  $(x_0,y_0)\in\{(x_0,y_0):x_0+y_0=1,x_0,y_0\geq0\}$, the censored distribution of $\left(\min(X,x_0),\min(Y,y_0)\right)$ is the same for all  $(X,Y) \in \mathcal{P}$.   \label{lem:infeas}
\end{proposition}
Proposition \ref{lem:infeas} is proved in Appendix~\ref{subsec:proof_lem_infeas} by constructing a set of probability distributions $\mathcal{P}_c$ for $c\in (0.5,1)$,
\[
\begin{split}
   \mathcal{P}_c = \{ & (X,Y) \mid \mathbb{P}(X=1,Y=1) = \mathbb{P}(X=c,Y=c) = p,  \\ 
   &\ \mathbb{P}(X=1,Y=c) = \mathbb{P}(X=c,Y=1) = 0.5 - p\text{, for some } p \in (0,0.5)\}.
\end{split}
\]
The two-dimensional example given in  Proposition \ref{lem:infeas} can be seamlessly extended to arbitrary $n$ dimensions since we can trivially set the demand as constant at all but two locations in an $n$-location network for $n\geq 2$. Through this impossibility result, we underscore the inherent impossibility of learning the joint demand distribution solely from \emph{censored demand} and \emph{limited supply}.

We further elucidate the challenge of censored demand by showing that the learning problem is considerably easier if uncensored demand data is available. It turns out that a simple dynamic learning algorithm ( Algorithm~\ref{alg_nloc:dl_uncensored_doubling}) with a doubling scheme can achieve optimal regret under this scenario without any cost structure assumptions as shown in Theorem \ref{thm:regret_1/2}. 
\begin{theorem}[Optimal Regret with Uncensored Data]
    \label{thm:regret_1/2}
    Given the oracle of uncensored demand data, under only Assumption~\ref{assumption3}, the dynamic learning algorithm, Algorithm~\ref{alg_nloc:dl_uncensored_doubling}, achieves $O\left(n^{3}\cdot \sqrt{T\log T}\right)$ regret.
\end{theorem}

The proof of the Theorem \ref{thm:regret_1/2} follows straightforwardly from the generalization bound that we have proved in Proposition~\ref{prop:offline_concentration}. In terms of computation, the offline problem \eqref{opt:offline} can be tackled by the MILP formulation \eqref{eq:offline_milp_opt} under any cost structure. Since the dynamic learning algorithm requires solving the offline problem in each period, we recommend using the LP reformulation \eqref{opt:offline_LP} instead for more efficient computation whenever the cost structure in Assumption~\ref{assump:lsc>rc}  holds. 

\subsection{Challenge of Network Correlation}
\label{subsec:network_indep}
The impossibility result in Section~\ref{subsec:counterexample} necessitates additional assumptions to facilitate online repositioning. In addition to cost structure, another direction to alleviate the curse of dimensionality is through the network independence assumption, as defined in Assumption \ref{assmp:indepedence}. Similar independence assumptions have been made in inventory control and learning of multi-echelon supply chain networks (see, e.g., \citet{bekci2023inventory,miao2022asymptotically}). We note that even with the demand independence stated in Assumption \ref{assmp:indepedence}, the inventory levels at different locations are still correlated due to the activities of customer trips and repositioning operations, and therefore the resulting problem is still significantly more complicated than the single-location case.

%[Network Independence]
\begin{assumption}
For  $t=1,...,T$, the demands from different locations are independent at each time, i.e.,  for $t=1,...,T$,  $\{d_{t,i}\}_{i\in\mathcal{N}}$ are independent. The demand $\dv_t$ and the probability transition matrix $\Pv_t$ are also independent.
\label{assmp:indepedence}
\end{assumption}
We propose a simple one-time learning algorithm (as described in Algorithm~\ref{alg_nloc:otl}), and show in Theorem \ref{thm:regret_2/3} that it has a regret guarantee of $\widetilde{O}(T^{2/3})$ that does not depend exponentially on $n$. In  Algorithm~\ref{alg_nloc:otl}, the first $n T_0$ time periods are dedicated to collecting uncensored demand data location by location, and then by the independence assumption, $T_0$ effective data samples can be constructed. We stress that the need for the network independence assumption solely comes from the data collection stage (Steps  \ref{alg:data_collect_st}--\ref{alg:data_collect_end} of Algorithm \ref{alg_nloc:otl}). In running Algorithm~\ref{alg_nloc:otl}, the number of exploration periods $T_0$ should be at the scale of $\eta T^{2/3}$ to achieve $\widetilde{O}(T^{2/3})$ regret. The parameter $\eta$, independent of $T$, is used to balance the trade-off of exploration and exploitation. Although the regret in Theorem \ref{thm:regret_2/3} is minimized at $\eta = (n/2)^{2/3}$, we have found that in numerical experiments, a smaller $\eta$ can be sufficient for learning and thus lead to smaller cumulative regret.

\begin{theorem}[Regret Under Network Independence]
    \label{thm:regret_2/3}
    Under Assumption~\ref{assumption3} and Assumption~\ref{assmp:indepedence}, the one-time learning algorithm, Algorithm~\ref{alg_nloc:otl}, achieves $O\left((\eta+ n\eta^{-1/2}) n^{2} T^{2/3}\sqrt{\log T}\right)$ regret when $T_0 = \eta T^{2/3}$ and $\eta$ is an algorithm hyperparameter.
\end{theorem}

To prove the regret bound in Theorem \ref{thm:regret_2/3}, we adopt the generalization bound established in Proposition~\ref{prop:offline_concentration}. We attain the $\widetilde{O}(T^{2/3})$ regret of the one-time learning algorithm (Algorithm \ref{alg_nloc:otl}) in contrast to the ${O}(T^{1/2})$ regret of the dynamic learning algorithm (Algorithm \ref{alg_nloc:dl_uncensored_doubling}) due to the periods needed for collecting uncensored data. While this rate is not optimal, it is still notable as the rate $\widetilde{O}(T^{2/3})$ refrains from the curse of dimensionality and do not depend on $n$ in the power of $T$. Moreover, since the offline problem is only solved once in Algorithm~\ref{alg_nloc:otl}, we can effectively use MILP formulation to solve the offline problem in Algorithm~\ref{alg_nloc:otl}, and therefore both the theoretical guarantee and computational efficiency of Algorithm~\ref{alg_nloc:otl} does not rely on the cost structure. 

In this sense, our proposed Algorithm~\ref{alg_nloc:otl} nicely fills the gap in addressing scenarios where the cost structure in Assumption \ref{assump:lsc>rc} fails to hold, with a $\widetilde{O}(n^2T^{2/3})$ regret guarantee that does not exponentially depend on $n$. 

\subsection{Extension to Heterogeneous Rental Durations and Heterogeneous Start Times}
\label{subsection:extension}

In this subsection, we generalize our analysis to allow \emph{heterogeneous} rental durations and \emph{heterogeneous} start times, and notably we show that our \pcr{SOAR} algorithm, with an appropriate generalization, continues to work with provable theoretical guarantees and numerical effectiveness.

Consistent with recent literature \citep{he2020robust, akturk2022managing}, our main model assumes that rental and review periods are synchronous, with each unit being used at most once per period. In contrast, the model by \cite{benjaafar2018dynamic}, which focuses on real-time dynamic repositioning, considers scenarios where the rental period is an integral multiple of review periods. Our work is thus complementary; in addition to addressing distinct online learning challenges, we model a different operational context characterized by long review periods and less frequent repositioning. To that end, motivated by practices like overnight repositioning \citep{yang2022overnight}, we generalize our model by decomposing each review period into multiple subperiods. This extended framework can capture rentals with durations that are fractions of a review period and accommodate multiple trips within a single period.

\begin{figure}[!htb]
\centering
% \vspace{-0.5em}
\caption{Illustration of inventory with heterogeneous rental durations and heterogeneous start times.}
\label{fig:generalized_inventory_update}
\begin{tikzpicture}[>=Latex,
% Define styles for different types of boxes
    x-box/.style={draw, rounded corners, fill=blue!20, opacity=0.8},
    y-box/.style={draw, rounded corners, fill=purple!20, opacity=0.8},
    note/.style={align=center, font=\footnotesize}]
% Timeline
\draw[->, thick] (0,0) -- (14,0);
% Time points, ticks, and labels
\foreach \i/\x in {1/0, 2/4, 3/8, 4/12} {
    % Vertical tick
    \draw[thick] (\x,0.3) -- (\x,0);
    % Labels below and above the tick
    \node[x-box, below] at (\x,-0.6) {$\xv_{\i}$};
    \node[y-box, above] at (\x,0.8) {$\yv_{\i}$};
}

% Subperiods
\foreach \i/\x in {1/0, 2/4, 3/8} {
    \foreach \j/\subx in {2/0.8} {
        % Subperiod tick
        \draw[thick] (\x+\subx,0.2) -- (\x+\subx,0);
        % Subperiod label
        \node[above] at (\x+\subx,0.8+0.2) {\footnotesize $\xv_{\i\j}$};
        \node[above, red] at (\x+\subx,0.4+0.2) {\footnotesize $\gammav_{\i\j}$};
    }
    \foreach \j/\subx in {5/3.2} {
        % Subperiod tick
        \draw[thick] (\x+\subx,0.1) -- (\x+\subx,0);
        % Subperiod label
        \node[above] at (\x+\subx,0.8+0.2) {\footnotesize $\xv_{\i {H}}$};
        \node[above, red] at (\x+\subx,0.4+0.2) {\footnotesize $\gammav_{\i {H}}$};
    }
    \foreach \j/\subx in { 3/1.6, 4/2.4} {
        % Subperiod tick
        \draw[thick] (\x+\subx,0.1) -- (\x+\subx,0);
        % Subperiod label
        \node[above] at (\x+\subx,0.8+0.2) {\footnotesize $\cdots$};
        \node[above] at (\x+\subx,0.4+0.2) {\footnotesize $\cdots$};
    }
}

% Dots for continuation
\draw[thick] (12,0.3) -- (12,0); % Vertical tick at the end
\node[below] at (13,-0.4) {$\cdots$};
\node[above] at (13,0.4) {$\cdots$};
\node[below] at (14,0) {\footnotesize Time};
\node[orange] at (-1.3,1.1) {\footnotesize Reposition to};
\node[teal] at (-1.3,-0.5) {\footnotesize Rentals occur};
\node[above, red] at (13.8,0.6) {\footnotesize Outstanding inventory};
% Period index annotations
\foreach \i/\x in {1/2, 2/6, 3/10} {
    \node at (\x, -1) {\footnotesize Period \i};
}

% Trip and demand index annotations
\foreach \i/\x in {1/2, 2/6, 3/10} {
    \node[below, teal] at (\x, -0.2) {\footnotesize $\{(\dv^{c}_{\i k}, \Pv_{\i k})\}_{k=1}^H$};
}

% Dashed curves for flow
\foreach \x/\nextx in {0/0.8, 4/4.8, 8/8.8, 12/12.8} {
    % Curve from \xv_t to \yv_t
    \draw[thick, ->, orange] (\x,-0.6) .. controls (\x-0.4,0.1) .. (\x,0.8);
}

\foreach \x/\nextx in {0/0.8, 4/4.8, 8/8.8} {
     % Curve from \xv_tH to \xv_{t(H+1)}
    \draw[thick, ->, dashed, teal] (\x+3.2,0.6) .. controls (\x+3.3,-0.1) .. (\x+4,-0.6);
}

\foreach \x/\nextx in {0/0.8, 4/4.8, 8/8.8}  {
 % Curve from \yv_t to \xv_{t+1}
    \draw[dashed, thick, ->, teal] (\x,0.4+0.1) .. controls (\x+0.4,-0.5) .. (\nextx,0.6);
}
% Curve from x_{12} to x_{13} ...
\foreach \x/\nextx in {0.8/1.6, 1.6/2.4, 2.4/3.2, 4.8/5.6, 5.6/6.4, 6.4/7.2, 8.8/9.6, 9.6/10.4, 10.4/11.2} {
    \draw[dashed, thick, ->, teal] (\x,0.6) .. controls (\x+0.2,-0.5) .. (\nextx,0.6);
}
\end{tikzpicture}
%\vspace{2em}
\end{figure}

To accommodate heterogeneous rental durations and start times, we partition each review period $t$ into $H$ subperiods, indexed by $k=1,\dots,H$; quantities in subperiod $k$ carry the subscript $tk$. Demand may arrive in any subperiod, and rentals may be returned after any number of subperiods, captured by the sequence of demand vectors and origin-to-destination matrices $\{(\dv_{tk}, \Pv_{tk})\}_{k=1}^H$. For any $k=1,\dots,H-1$ and $i=1,\dots,n$, the row sum $\sum_{j} P_{tk,ij}$ may be strictly less than $1$, indicating that a fraction $1-\sum_{j} P_{tk,ij}$ of inventory departing from location $i$ remains unreturned at the end of subperiod $k$. Let $\gammav_{tk}$ denote the outstanding inventory vector (originating from the $n$ locations) at the beginning of the $k$-th subperiod of review period $t$. For notational convenience, set $\yv_t=\xv_{t1}$ and $\xv_{t+1}=\xv_{t(H+1)}$. The inventory dynamics are illustrated in Figure~\ref{fig:generalized_inventory_update} and given by
\begin{align}
    \xv_{t(k+1)} &= (\xv_{tk} - \dv_{tk})^+ + \Pv_{tk}^\T \left[\min(\xv_{tk}, \dv_{tk}) + \gammav_{tk} \right], \quad k = 1,\dots, H, \label{eq:inv_gen_x}\\
    \gammav_{t(k+1)} & = \left[ \min(\xv_{tk}, \dv_{tk}) + \gammav_{tk} \right] \circ \left[ (\Iv-\Pv_{tk}) \ev\right], \quad k = 1,\dots, H, \label{eq:inv_gen_gamma}
\end{align}
where $\circ$ denotes the Hadamard product and $\ev$ is the $n$-dimensional all-ones vector. This subperiod formulation reflects the practice of infrequent repostioning in vehicle-sharing systems\citep{yang2022overnight}.

For any period $t$, $\dv_{t1}, \dots, \dv_{tH}$ do not have to be i.i.d., and $\Pv_{t1}, \dots, \Pv_{tH}$ do not have to be i.i.d. either. This allows for non-stationarity across different subperiods within the same review period. All unreturned units, regardless of their rental start times, are returned before each repositioning operation since these operations occur during low-utility periods when rental activity is minimal. Alternatively, if unreturned units at the end of each review period maintain constant percentage $\rho>0$,  the base-stock repositioning policy would still be well-defined, lying in $\Delta_{n-1}(1-\rho)$.

Because of the possibility of multiple rental trips in one review period, the lost sales cost within period $t$ needs to account for cots summarized over $H$ subperiods, and the {modified} lost sales costs is defined in \eqref{eq:generalized_mod_lsc} by subtracting $\sum_{k=1}^H \sum_{i=1}^n \sum_{j=1}^n l_{ij} \cdot P_{th, ij} d_{tk,i}$ and noting that $x_{tk,i}$ is obtained recursively through \eqref{eq:inv_gen_x}.
\begin{equation}
    \widetilde{L}(\yv_t, \{(\dv_{tk}, \Pv_{tk})\}_{k=1}^H) = -\sum_{k=1}^H \sum_{i=1}^n \sum_{j=1}^n l_{ij} \cdot P_{th, ij} \min(x_{tk,i}, d_{tk, i}). \label{eq:generalized_mod_lsc}
\end{equation}
The repositioning cost at the end of each review period is given by $M(\yv_t - \xv_t)$, where $M(\cdot)$ is from the minimum cost flow problem defined as in \eqref{eq:repo_cost}. The modified total cost of review period $t$ is $ \widetilde{C}(\xv_t, \yv_t, \{(\dv_{tk}, \Pv_{tk})\}_{k=1}^H) = M(\yv_t, \xv_t) + \widetilde{L}(\yv_t, \{(\dv_{tk}, \Pv_{tk})\}_{k=1}^H).$ Consistent with previous analysis, we focus on base-stock type policies and study the online repositioning problem under the challenges of the spatial network structure and access to only realized origin-to-destination matrices $\{\Pv_{tk}\}_{k=1}^H$ and censored demands $\{\min(\xv_{tk}, \dv_{tk})\}_{k=1}^H$.  In Lemma~\ref{lem:disentangle_extended}, we bound the cumulative difference of $\widetilde{C}(\xv_t, \yv_t, \{(\dv_{tk}, \Pv_{tk})\}_{k=1}^H)$ and \emph{surrogate costs} $\widetilde{C}(\xv_{t+1}, \yv_t, \{(\dv_{tk}, \Pv_{tk})\}_{k=1}^H)$, defined through relabelling $\xv$, by the cumulative changes of repositioning policies.

\begin{figure}[!htb]
\caption{Illustration and numerical result of \pcr{SOAR-Extended}.}
    \begin{subfigure}[b]{0.49\textwidth}\resizebox{\linewidth}{!}{%
\begin{tikzpicture}[
    % Node styles with specific color encodings
    ytbox/.style={draw, rounded corners, align=center, minimum width=2.5cm, minimum height=1cm, fill=orange!20},
    s1box/.style={draw, rounded corners, align=center, minimum width=1cm, minimum height=1.4cm, fill=green!20, opacity=0.8},
    subgradbox/.style={draw, rounded corners, align=center, minimum width=2.5cm, minimum height=1cm, fill=blue!20, opacity=0.8},
    redbox/.style={draw, rounded corners, align=center, minimum width=2cm, minimum height=1cm, fill=red!20, opacity=0.8},
    note/.style={align=center, font=\small, fill=none},
    >=Stealth
]
% Update Y_t node at the top
\node (Yt) [ytbox] {Update $\yv_t$};
% Overlaid boxes for H1 to H with notes
\begin{scope}
    \node (H1) [below= 2cm of Yt, xshift=-0.8cm * 8, s1box] {$\dv^{c}_{t1}$ \\ $\Pv_{t1}$};
    \node (H2) [below=2cm of Yt, xshift=-0.6cm* 8, s1box] {$\dv^{c}_{t2}$ \\ $\Pv_{t2}$};
    \node (H3) [below=2cm of Yt, xshift=-0.4cm* 8, s1box] {$\cdots$};
    \node (H4) [below=2cm of Yt, xshift=-0.2cm* 8, s1box] {$\cdots$};
    \node (HH) [below=2cm of Yt, xshift=-0cm* 8, s1box] {$\dv^{c}_{tH}$ \\ $\Pv_{tH}$};

    % Notes above the boxes
    \node[note] at (H1.north) [yshift=0.3cm] {$1$};
    \node[note] at (H2.north) [yshift=0.3cm] {$2$};
    \node[note] at (H3.north) [yshift=0.3cm] {$\cdots$};
    \node[note] at (H4.north) [yshift=0.3cm] {$\cdots$};
    \node[note] at (HH.north) [yshift=0.3cm] {$H$};
    %\node[note] at (H1.west) [xshift=-1.2cm] {Censored data};
\end{scope}

\begin{scope}[on background layer]
% Big translucent box covering H1..HH:
% "fit" automatically covers the listed nodes.
\node (BigBox) [
  draw,
  thick,
  inner sep=0.5cm,
  fill=gray!30,
  rounded corners,
  opacity=0.3,
  fit=(H1)(H2)(H3)(H4)(HH),
  label={[align=center]above: {Censored data}}
] {};
\end{scope}

% Subgradient node
\node (subgrad) [below=2cm of HH, subgradbox, xshift = -0.3cm] {Dual $\{\lambdav_{tk}\}_{k=1}^H$};

%Arrows
\draw[->, thick, dashed] (H1.east) to (H2.west);
\draw[->, thick, dashed] (H2.east) to (H3.west);
\draw[->, thick, dashed] (H3.east) to (H4.west);
\draw[->, thick, dashed] (H4.east) to (HH.west);

\draw[->, thick] (BigBox.south) to[out=270, in=180] (subgrad.west);

% Arrows connecting nodes
\draw[->, thick, dashed] (Yt.south) to[out=210, in=50]  node[midway, above left] {$t\rightarrow t+1$} (H1.north) ;
% \draw[->, thick] (H1.south) to[out=270, in=150] (subgrad.north);
% \draw[->, thick] (H2.south) to[out=270, in=110] (subgrad.north);
% \draw[->, thick] (H3.south) to[out=270, in=110] (subgrad.north);
% \draw[->, thick] (H4.south) to[out=270, in=90] (subgrad.north);
% \draw[->, thick] (HH.south) to[out=270, in=70] (subgrad.north);
% \draw[->, thick, bend left=50] (subgrad.east) to[out=270, in=270] (Yt.east);

% New node on the right
\node (convert) [redbox, right=2.5cm of BigBox]
  {Subgradient $\{\muv_{tk}\}_{k=1}^H$};
  
% Single smooth merged arrow
\draw[->, thick] (BigBox.east) to[out=0, in=180] node[midway, below] {recover} (convert.west);
\draw[->, thick] (subgrad.east) to[out=30, in=240] node[midway, right] {}(convert.west);
\draw[->, thick] (convert.north) to[out=90, in=0] node[midway, above] {$\Pi_{\Delta_{n-1}}$} (Yt.east);

\end{tikzpicture}
}
\vspace{0.5em}
\caption{Schematic representation of Algorithm~\ref{alg_nloc:ogr_extended}.}
\label{fig:illustrate_ogr_extended}
\end{subfigure}
\begin{subfigure}[b]{0.5\textwidth}
\includegraphics[width = \linewidth]{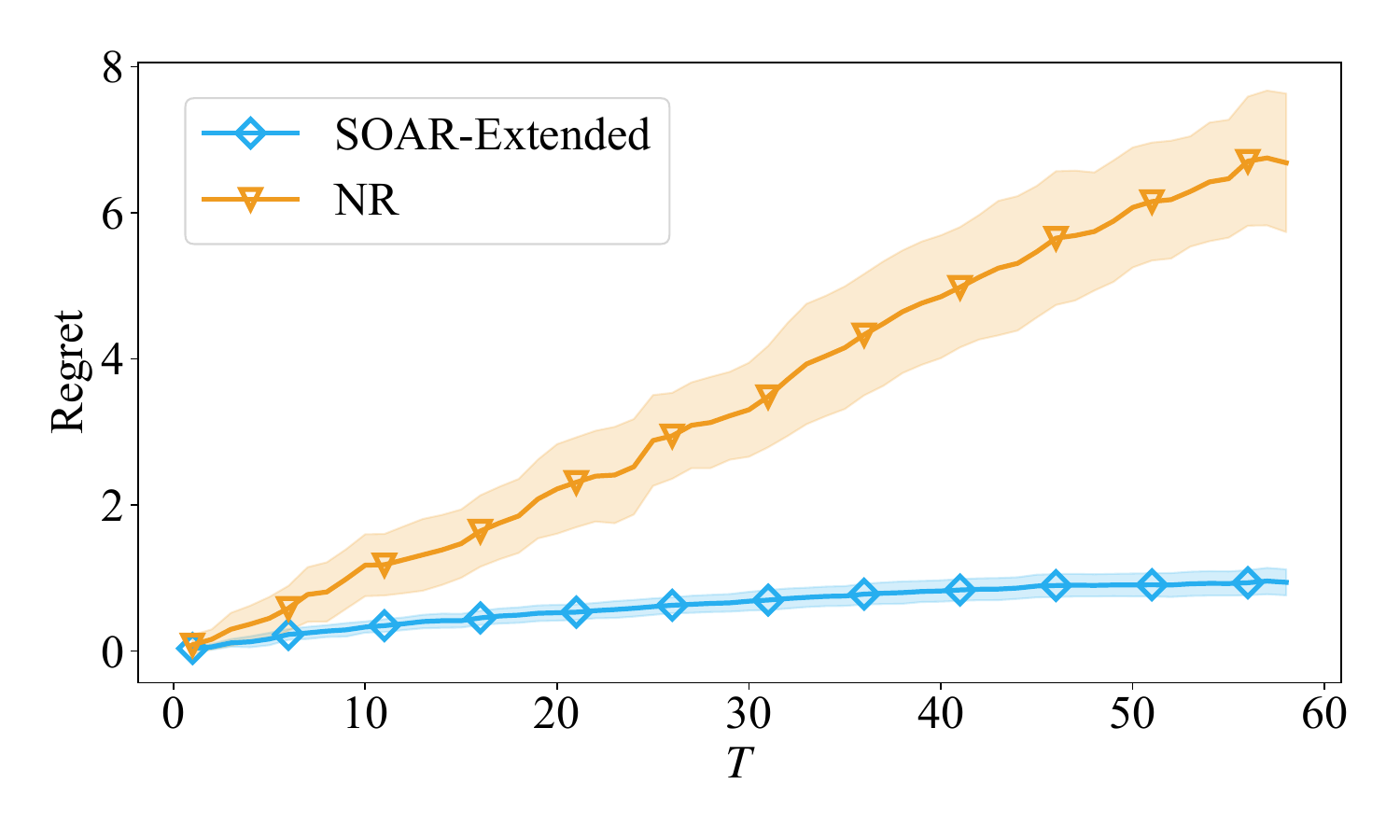}
\caption{Regret performance with $n = 10, H=8$.}
\label{fig:ogr_extended_numerical}
\end{subfigure}
{\footnotesize 
    \emph{Notes.} More numerical results and implementation details are provided in Appendix~\ref{appendix:ogr_extended_numerical}.}
%\vspace{-1em}
\end{figure}
We explain how to apply the principle of \pcr{SOAR} algorithm to the extended model with  an illustration in Figure~\ref{fig:illustrate_ogr_extended}, and present detailed description of \pcr{SOAR-Extended} in Algorithm~\ref{alg_nloc:ogr_extended}. At period $t$, the algorithm is initialized by setting the target inventory as $\xv_{t1} = \yv_t$ and observe realized {censored demands} $\dv_{th}^{c}=\min(\xv_{th},\dv_{th})$ for $h \in [H], t\in[T]$. A key step is to figure out how to find the gradient direction in order to modify the repositioning policy $\yv_t$. We construct the following linear programming problem to minimize the {surrogate costs} $\widetilde{C}(\xv_{t+1}, \yv_t, \{(\dv_{tk}, \Pv_{tk})\}_{k=1}^H)$. 
\begin{align}
    \min_{\xi_{t,ij}, \gamma_{tk, i}, w_{tk, i}} &\sum_{i=1}^n \sum_{j=1}^n c_{ij} \xi_{t, ij} - \sum_{h=1}^H \sum_{i=1}^n \sum_{j=1}^n l_{ij} P_{th, ij} w_{th,i} \nonumber\\
   \text{ subject to} & \sum_{i=1}^n \xi_{t, ij} - \sum_{i' = 1}^n \xi_{t, ji'} = \sum_{k=1}^H \left[ w_{tk,j} -\sum_{i=1}^n P_{tk, ij} ( w_{tk,i} + \gamma_{tk,i})  \right], \forall j \in [n], \nonumber\\
   & \gamma_{t(k+1),i} = (w_{tk,i} + \gamma_{tk,i})\left(1- \sum_{j=1}^n P_{tk,ij} \right), \forall k \in [H], i \in [n], \nonumber \\
   & \gamma_{t1,i} = 0, \forall i \in [n], \nonumber \\
   &  w_{tk, i} \geq 0,\, \xi_{t, ij} \geq 0, \quad \nonumber \\
   & w_{t1, i} \leq (\dv_{t1}^c)_i, \quad
    w_{t2, i} \leq (\dv_{t2}^c)_i, \quad
    \dots \quad
    w_{tH, i} \leq (\dv_{tH}^c)_i, \forall i\in [n].  \label{eq:ogd_olp_extended}
\end{align}
We take $\lambdav_{tk} \in \mathbb{R}^n$ to be the dual optimal solution to the constraints $w_{tk, i} \leq (\dv_{tk}^c)_i$, $\forall i \in [n]$  in \eqref{eq:ogd_olp_extended} for $k\in [H]$, and define $\gv_{tk} = \lambdav_{tk} \circ  \mathbbm{1}\{\dv_{tk}^c = \xv_{tk}\}$. Unlike the original \pcr{SOAR} algorithm, $\gv_{tk}$ no longer represents a subgradient with respect to $\yv_t$ in the surrogate cost function. Instead, we recursively recover  components of the subgradient $\muv_{tk}$ from $\gv_{tk}$ through \eqref{eq:recursion}, with detailed theoretical analysis provided in Appendix~\ref{appendix:ogr_extended_theory}.
\begin{equation}
    \gv_{tk} = \muv_{tk} + (\Iv - \Pv_{tk}) \sum_{l=k+1}^H \muv_{tl} - \sum_{l= k+2}^H \left\{ \sum_{s=k+1}^{l-1}\Pv_{ts} \muv_{tl} \circ \prod_{u=k}^{s-1} \left[(\Iv - \Pv_{tk})\ev  \right] \right\}.
    \label{eq:recursion}
\end{equation}
The repositioning level for the next time period is updated as $\yv_{t+1} = \Pi_{\Delta_{n-1}} \left( \yv_t -  1/(H\sqrt{t}) \sum_{k=1}^H \muv_{tk} \right)$, where $1/(H\sqrt{t})$ is the step size at period $t$.

\begin{theorem} \label{thm:gdr_extended}
Under only Assumption \ref{assump:lsc>rc_extended}, the output of Algorithm \ref{alg_nloc:ogr_extended} over a horizon of $T$ review periods satisfies
    \begin{equation*}
        \sum\limits_{t=1}^{T}\widetilde{C}_{t}(\xv_t(\bslevel_{t-1}),\bslevel_{t},\{(\dv_{tk}, \Pv_{tk})\}_{k=1}^H)
        - \min_{\bslevel\in\Delta_{n-1}} \sum\limits_{t=1}^{T}\widetilde{C}_{t}(\xv_t(\bslevel),\bslevel,\{(\dv_{tk}, \Pv_{tk})\}_{k=1}^H) \leq O(n^{2.5}H\sqrt{T}). 
    \end{equation*}
    for any initial inventory level $\bslevel_0:=\bslevel_1\in\Delta_{n-1}$ and any sequence of demand and origin-to-destination probability matrix $\left\{\left\{(\dv_{tk},\Pv_{tk})\right\}_{k=1}^{H}\right\}_{t=1}^{T}$. 
\end{theorem}

The regret rate of $O(H\sqrt{T})$ holds for any sequence of demand vectors and origin-to-destination matrices, including adversarial cases. The stochastic version of regret for Theorem~\ref{thm:gdr_extended} follows analogously to Corollary~\ref{cor:gdr}. We note that the time horizon contains $\widetilde{T}=TH$ subperiods and therefore the rate is equivalently $O(\sqrt{H\widetilde{T}})$. The price of $\sqrt{H}$ is paid because the decision is only made every $H$ subperiods and mainly comes from the Lipschitz constant bound on the cumulative costs, similar to previously shown in Lemma~\ref{lem:h_lipschitz}. We note that the scenario is different from the batched bandit literature in machine learning, as here not the observations but the decisions are only feasible every $H$ subperiods. 

Theoretically speaking, the Lipschitz bound could be conservative since randomness in returns and the influence of demand parameters means that differences in $\yv_t = \xv_{t1}$ may not necessarily propagate to large differences in $\xv_{tk}$ for subsequent $k$'s through equations \eqref{eq:inv_gen_x} and \eqref{eq:inv_gen_gamma}. Nevertheless, when $H$ is independent of $T$ or grows moderately such as $H = O(\log T)$, our theoretical bound maintains near-optimal regret rate in $T$. Numerically, we have found the algorithm performs well even when $H$ is large. Notably, the sublinear regret rate is evident over very short time horizons such as $T=60$, which contrasts with the linear regret of a no-repositioning policy as shown in Figure~\ref{fig:ogr_extended_numerical}.

%% file: files/numerical_withfigures.tex
\section{Numerical Illustration}
\label{section:numerical}

We compare the numerical performances of \pcr{SOAR} against the clairvoyant best base-stock policy \pcr{OPT},  the no-repositioning policy \pcr{NR}, and the one-time learning \pcr{OTL} approach (Algorithm~\ref{alg_nloc:otl}). When the LP formulation is feasible (i.e., Assumption~\ref{assump:lsc>rc} holds, we use \pcr{OTL-LP} instead of \pcr{OTL-MILP} for higher computational efficiency. The dynamic learning approach (Algorithm~\ref{alg_nloc:dl_uncensored_doubling}) relies on the oracle of uncensored demand data and is thus not included in comparison. For richer comparison, we consider two metrics: 
\[
\text{(i) Regret($T$) in $\log$ scale; (ii) Relative Regret($T$) = 100\% $\times \frac{\text{Regret($T$)}}{\text{$T$-period cumulative cost of \pcr{OPT}}}$.}
\]
We generate the synthetic data under different network scenarios (Appendix~\ref{appendix:synthetic_exp}) and report average performances across multiple repeated runs. 

\begin{figure}[!htb]
\centering
\caption{Comparison of \pcr{SOAR}, No-Repositioning \pcr{NR}, and one-time learning \pcr{OTL-LP} with $n=10$.}
\label{fig:n10-networkindepTrue}
     \begin{subfigure}[t]{0.49\textwidth}
    \centering
    \includegraphics[width=\linewidth]{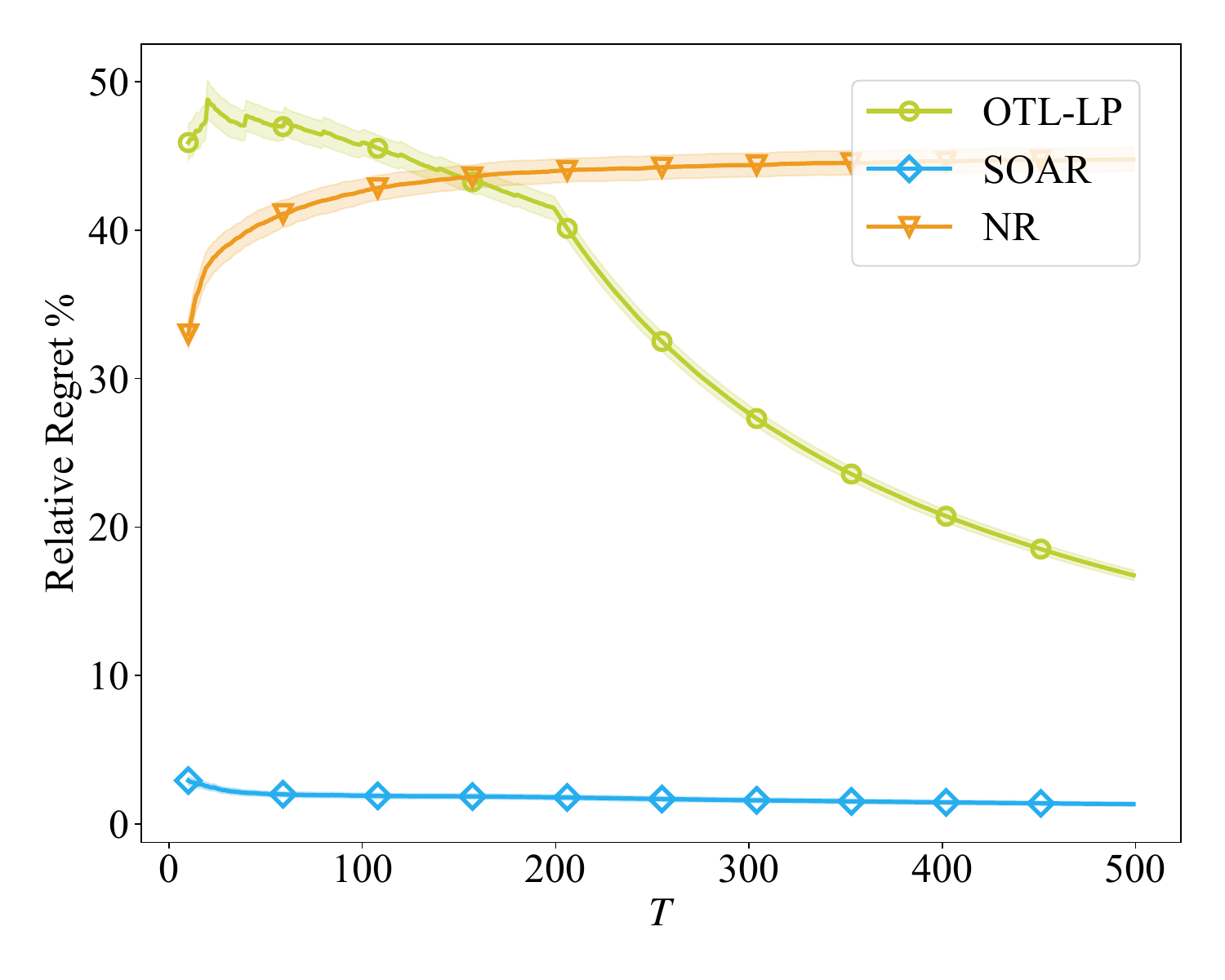}
    \end{subfigure} 
    \begin{subfigure}[t]{0.49\textwidth}
        \centering
    \includegraphics[width=\linewidth]{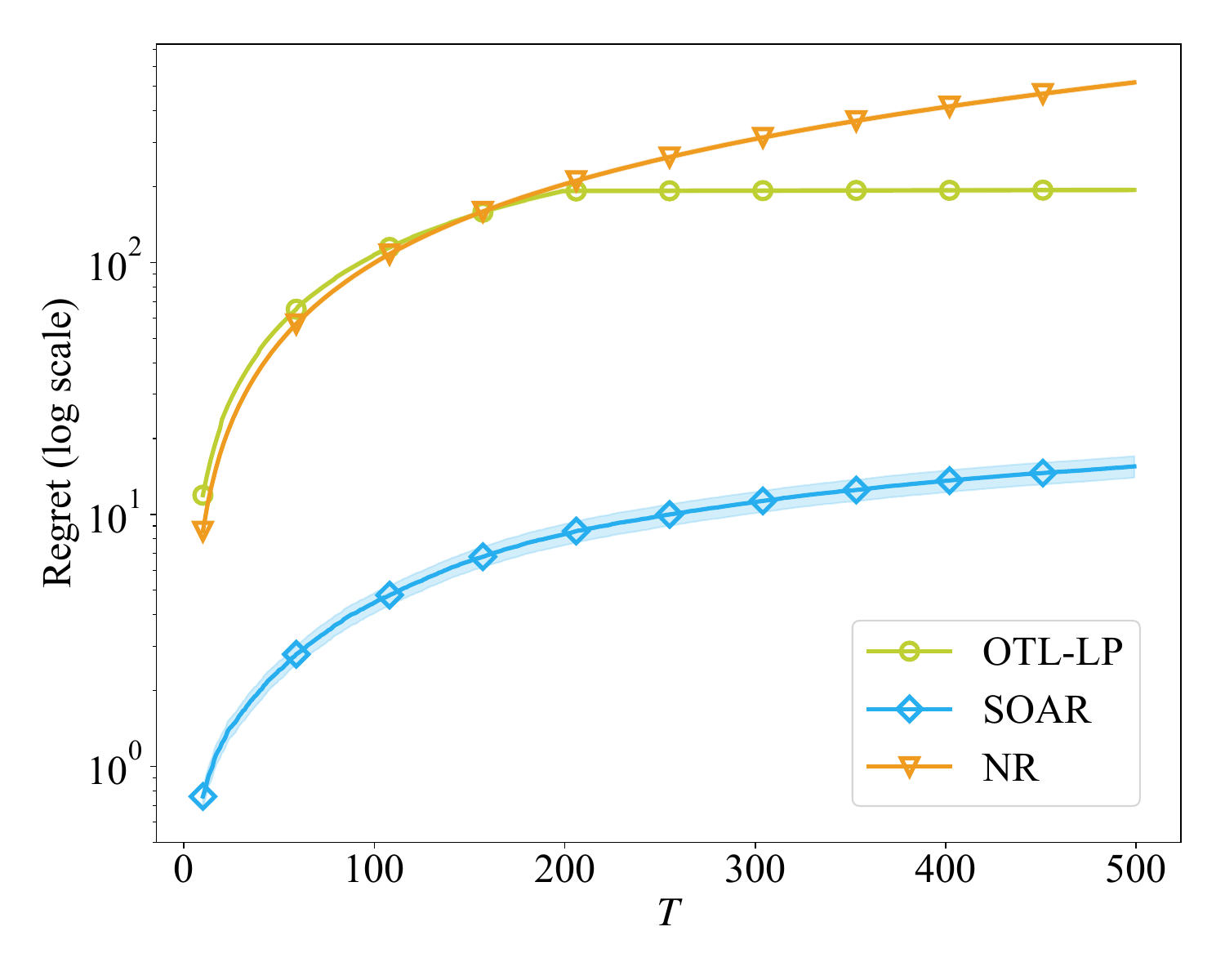}
    \end{subfigure}
\end{figure}

\paragraph{Strong Numerical Performance of \pcr{SOAR}.} Under both metrics, \pcr{SOAR} significantly outperform \pcr{OTL-LP} and \pcr{NR} since the beginning of the time horizon. As shown in Figure~\ref{fig:n10-networkindepTrue}, the relative regret percentage of \pcr{SOAR} is consistently lower than $5\%$. Remarkably, \pcr{SOAR} establishes regret dominance within a short time horizon. This stands in contrast to standard online learning approaches, where demonstrating such dominance in numerical experiments often necessitates a much longer horizon. Indeed, the $500$-period horizon was chosen primarily to allow \pcr{OTL-LP} enough time to complete its exploration phase.

\begin{figure}[!htb]
\centering
\caption{Comparison of \pcr{SOAR}, No-Repositioning \pcr{NR}, and one-time learning \pcr{OTL-LP} with correlated newtwork demand and $n=10$.}
\label{fig:n10-networkindepFalse}
     \begin{subfigure}[t]{0.49\textwidth}
    \centering
    \includegraphics[width=\linewidth]{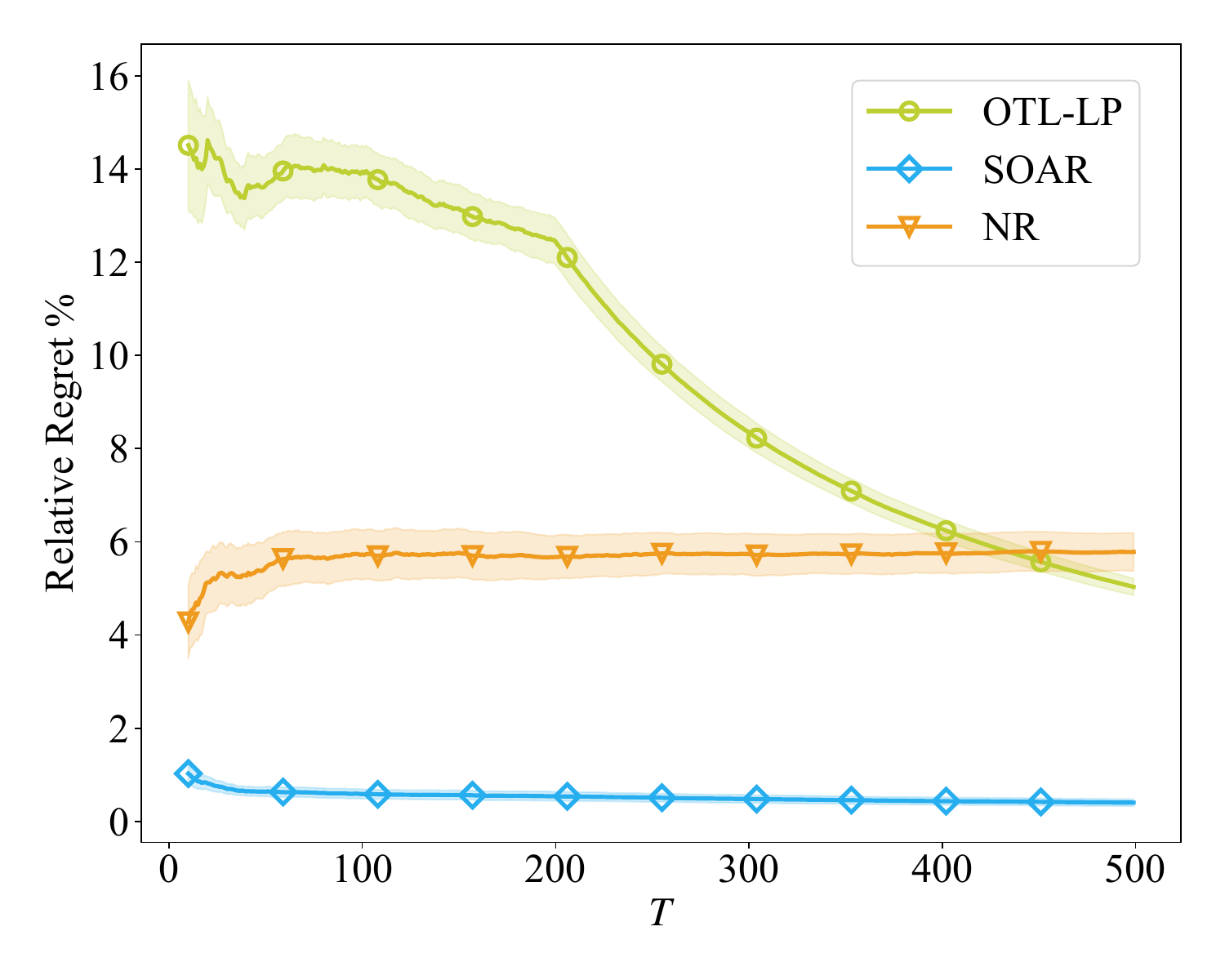}
    \end{subfigure} 
    \begin{subfigure}[t]{0.49\textwidth}
        \centering
    \includegraphics[width=\linewidth]{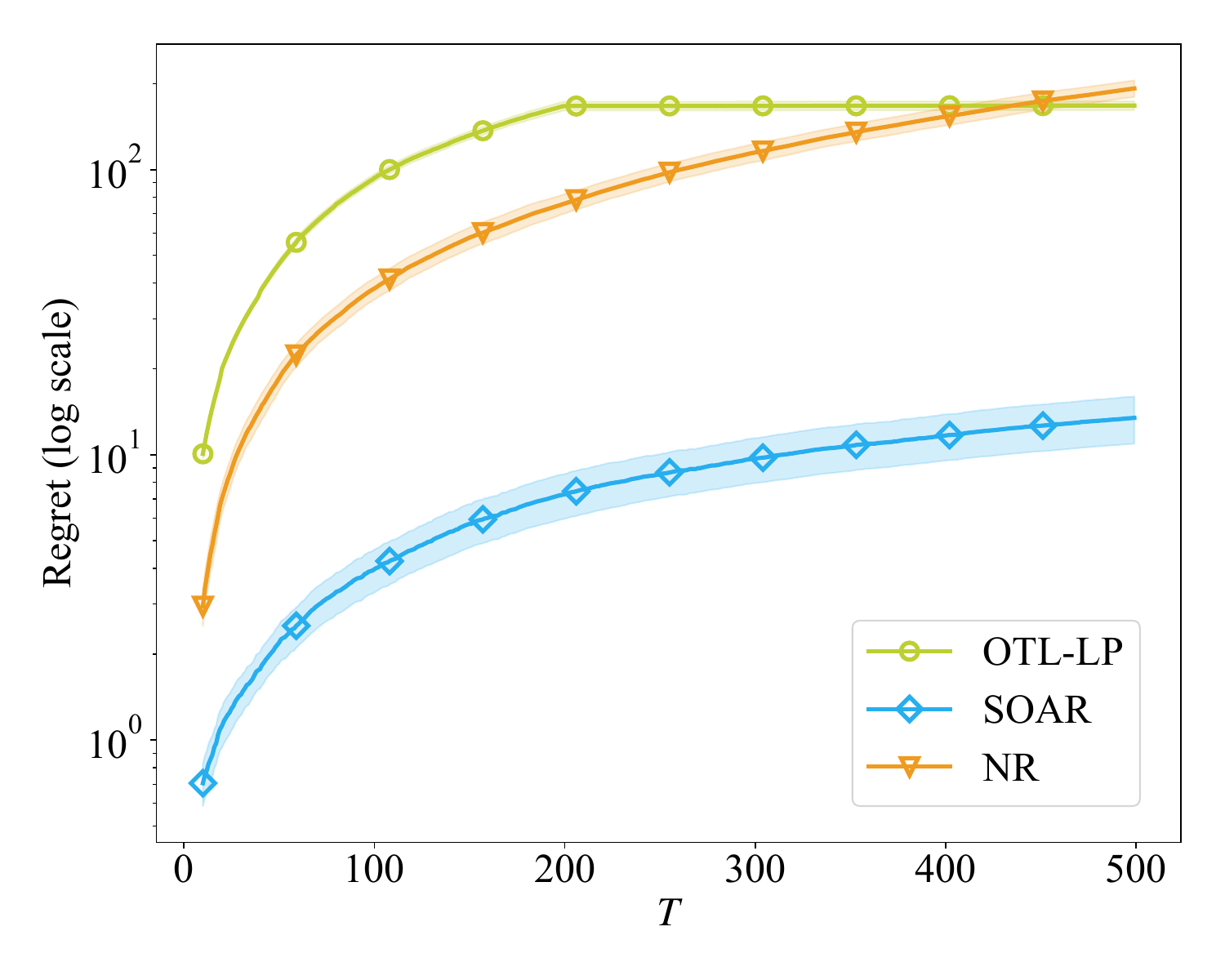}
    \end{subfigure}
\end{figure}

\paragraph{Impact of Network Correlation.} When network demand is correlated (i.e., sampled from a truncated multivariate Gaussian distribution as detailed in Appendix~\ref{appendix:synthetic_exp}), \pcr{OTL-LP} cannot collect true i.i.d. samples during its exploration phase, which theoretically impacts the learned policy. This is reflected in the diminished performance advantage of \pcr{OTL-LP} over \pcr{NR} in Figure~\ref{fig:n10-networkindepFalse}. Nevertheless, \pcr{OTL-LP} eventually outperforms \pcr{NR}, demonstrating that a policy learned from imperfect data is still superior to taking no action. We also note that \pcr{SOAR} still significantly outperforms both \pcr{OTL-LP} and \pcr{NR} from the outset. Moreover, \pcr{SOAR} can achieve relative regret that is zero or even negative. This is possible because the \pcr{OPT} benchmark is computed in an expectation sense, whereas the instance-wise performance of \pcr{SOAR} can be better.

\paragraph{Value of Repositioning.} Another observation we can make from Figure~\ref{fig:n10-networkindepTrue} and \ref{fig:n10-networkindepFalse} is that, the cost of not repositioning at all (\pcr{NR}) can be rather high, with relative regret percentage of over $40\%$ and the regret is noticeably increasing under the $\log$ scale.  This confirms the initial intuition that without active intervention, the system does not self-correct and can instead enter a vicious cycle of lost sales, leading to significant system-wide losses. In contrast, while the exploration phase for \pcr{OTL-LP} is initially costly, which incurs higher regret than \pcr{NR} for the first 100 periods, it rapidly improves once exploitation begins, significantly reducing its overall regret.

\begin{table}[!ht]
\centering
\caption{Regret comparison (post-exploration phase) when the cost condition does not hold.}
\label{table:milp_comparison}
\begin{tabular}{cccccccccc}
\toprule
Regret at Period & 50 & 60 & 70 & 80 & 90 & 100 & 110 & 120\\ \midrule
\pcr{OTL-MILP} & 36.25 & 55.31 & 60.56 & 60.61 & 60.79 & 60.97 & 61.03 & 61.12 \\ \midrule
\pcr{OTL-LP} & 36.25 & 55.31 & 61.95 & 64.99 & 67.78 & 71.07 & 74.31 & 77.16 \\ 
\bottomrule
\end{tabular}
\vspace{-0.5em}
\end{table}
\paragraph{Effectiveness of the Exact MILP Formulation.}
As noted in Section~\ref{subsec:network_indep}, the MILP computing time is not a bottleneck for \pcr{OTL} since it only needs to be solved once, provided it is solvable with given computing resources. To illustrate, we adjust the problem parameters to create a high-repositoning-cost setting where the cost condition \eqref{eq:lsc>rc} is violated. We run the \pcr{OTL} algorithm with the MILP and LP formulations, respectively. Table~\ref{table:milp_comparison} shows that the \pcr{OTL-LP} approach can perform poorly when the cost structure assumption does not hold, whereas the \pcr{OTL-MILP} approach successfully learns the optimal policy and achieves a near-constant regret during the exploitation period. This example highlights the merit of our MILP reformulation for problem instances under general cost structures, a contribution we believe is of broader independent interest.

%Integer programming usually takes, if not impossible, a much longer time than LP even with well-developed solvers in Gurobi, and our MILP problem is no exception. However, as we have mentioned in Section~\ref{subsec:network_indep}, the one-time learning algorithm equipped with MILP offline problem allows us to continue to achieve a sublinear regret when the cost structure assumption does not hold.

%Potential TODO: \paragraph{Performance on Adversarial Data.}

%% file: files/conclusion.tex
\section{Conclusion}
\label{section:conclusion}

Efficient vehicle repositioning is central to the viability of vehicle‑sharing systems and, more broadly, to sustainable urban mobility. We study this problem through the lens of network inventory management, focusing on spatial mismatch and demand censoring. Our analysis establishes fundamental properties of the underlying MDP, demonstrates the asymptotic‑optimality of base-stock policies, and develops data‑driven methods, both offline and online, with provable performance guarantees under censored observations. Methodologically, the paper advances learning with censored data in multi‑location, fixed‑inventory networks; the insights extend beyond vehicle sharing to other inventory systems. For practitioners, our results quantify the trade‑off between repositioning and lost‑sales costs under fleet constraints and show how structure‑exploiting analytics can significantly reduce operating costs.

Building on our analysis, several directions merit further study. On the modeling side, incorporating richer operational features, including batched or asynchronous actions and  contextual information such as weather and traffic can better capture practice while preserving tractability. On the decision side, integrating pricing or other demand‑shaping incentives with repositioning, accounting for infrastructure or maintenance constraints, and designing privacy‑aware estimators for censored demand are all very promising avenues. We view these extensions as natural next steps toward a comprehensive, data‑driven framework for managing future shared‑mobility networks.

%% file: files/appendix_fullinfo.tex
%--------------

\section{Proofs of Optimal Policy}
\label{appendix:fullinfo}

\subsection{Proof of Theorem~\ref{theorem:optimality_exist}}

For the existence of optimal stationary policies in a general Markov decision process with infinite state space, a few sufficient conditions have been proposed in the literature \citep{feinberg2012average}.
For notational simplicity, we define the one-step expected cost function as $c(\xv,\yv) := \Expt_{\dv,\Pv}[C(\xv,\yv,\dv,\Pv)] $.

\proof{Proof of Theorem~\ref{theorem:optimality_exist}.}
The proof is based on the results in \citet[Proposition 1.3]{schal1993average} and \citet[Theorem 1]{feinberg2012average} that state that conditions~W\ref{condition_w1}\ref{condition_w2} and B\ref{condition_b1}\ref{condition_b2} are sufficient. We will provide the verification of the condition~W*\ref{condition_w1} \ref{condition_w2}  and condition~B\ref{condition_b1} below since the  Condition~B\ref{condition_b2} is verified in Proposition~\ref{prop:verify_condition_b}.

{ 

Conditions W*\ref{condition_w1} and W*\ref{condition_w2} are straightforward to verify. Condition W*\ref{condition_w1} holds because the state transition function \eqref{eq:state_update_vector} is continuous \citep[Lemma 3.1]{feinberg2016optimality}. Condition W*\ref{condition_w2} is a slightly stronger version of the $\K$-inf-compact condition in \citet[Assumption W*(ii)]{feinberg2016optimality}. A function is called inf-compact if all of its level sets are compact, namely $ \{(\xv,\yv)\mid c(\xv,\yv) \leq a\} \text{ is compact for all }a \in \mathbb{R}.$ Next, we argue that this stronger $\K$-inf compact property in Condition W*\ref{condition_w2} clearly  holds in our vehicle repositioning problem. Because the cost function $c(\xv,\yv)$ is continuous with respect to $(\xv,\yv)$, the level set, which is the preimage of a closed set $(-\infty, a]$, is also closed for any $a\in \mathbb{R}$. Since the closed level set also belongs to the bounded set $\Delta_{n-1}\times \Delta_{n-1}$, the level set is both bounded and closed, and thus compact.

We summarize the validity of Condition B\ref{condition_b2} into the following Proposition~\ref{prop:verify_condition_b}. 
}
\halmos

\subsection{Proof of Proposition~\ref{prop:verify_condition_b}}

\proof{Proof of Proposition~\ref{prop:verify_condition_b}.}
For any $\rho\in(0,1)$, and let $\xv_\rho$ be a state such that $v^*_\rho(\xv_\rho) = m_\rho := \inf_{\xv \in \Delta_{n-1}} v^*_\rho(\xv)$. Such $\xv_\rho$ always exists because the state space $\Delta_{n-1}$ is compact, and the value function $v^*_\rho(\xv)$ is continuous. Let $\pi_{\rho}$ be a stationary optimal policy under the $\rho$-discounted setting, then by definition $v^*_\rho(\xv_\rho) = v^{\pi_\rho}_\rho(\xv_\rho) =  m_\rho$.

Suppose the initial state is $\xv$. We define a new policy $\sigma$ as follows. For the first time period, $\sigma$ repositions to the level that policy $\pi_{\rho}$ would reposition to at state $\xv_\rho$. After the first period, policy $\sigma$ behaves exactly like $\pi_{\rho}$.  Comparing $v^\sigma_\rho(\xv)$ and $v_\rho^*(\xv_\rho)$, we can see that they only differ in the costs of the first time period. Therefore,
\begin{equation}
     v^\sigma_\rho(\xv) \leq \max_{i,j}c_{ij}n + n U\max_{i,j}l_{i,j} + v^*_\rho(\xv_\rho) = \max_{i,j}c_{ij}n + n U\max_{i,j}l_{i,j} +m_\rho ,
\label{eq:sigma_rho}
\end{equation}
where the first inequality is because the amount of inventory moved from each location is at most the total inventory $1$ and thus the repositioning cost is bounded by $\max_{i,j}c_{ij} n$,  and the lost sales cost is bounded by $\max_{i,j}l_{i,j}n U$ because the amount of demand leaving every location is bounded by $U$ according to Assumption~\ref{assumption3}. This bound is very loose, but it suffices for the purpose of proving boundedness. %We show a stricter bound on the repositioning cost when proving Lemma~\ref{lemma:bounded_difference_in_value}.
On the other hand, by the optimality of   $v^*_\rho(\xv)$, we have
$v^*_\rho (\xv) \leq v^\sigma_\rho(\xv)$
and plugging this back into \eqref{eq:sigma_rho}, we have 
$
v^*_\rho (\xv) \leq \max_{i,j}c_{ij}n + n U\max_{i,j}l_{i,j} +m_\rho.
$
Therefore we have shown that
$
r_\rho(\xv) = v^*_\rho (\xv) - m_\rho < +\infty.
$\halmos

\subsection{Proof of Theorem~\ref{thm:asymptotic_optimal}}
\proof{Proof of Theorem~\ref{thm:asymptotic_optimal}.}
We consider the best base-stock repositioning policy $\bslevel^*$, which is defined by
\[
\bslevel^* \in \arg\,\min_{\bslevel \in \Delta_{n-1}} \limsup_{T\rightarrow \infty} \frac{1}{T} \sum_{t=1}^T \Expt^{\pi_{\bslevel}}[C_t].
\]

Observing that under the base-stock repositioning policy, the costs across time periods are all independent and identically distributed except the first period. Therefore, we can equivalently characterize the optimal base-stock level $\bslevel^*$ as follows, 
\begin{equation}
\begin{split}
    \bslevel^* \in \arg\,\min_{\bslevel} &\quad \Expt \left[  M(\bslevel - \xv^\bslevel_0) + L(\bslevel, \dv, \Pv)\right],\\
    \text{s.t.} &\quad \xv^\bslevel_0 = (\bslevel - \dv_0)^+ + \Pv_0^\T \min \{\bslevel, \dv_0\},
\end{split}
\label{eq:base_stock_oneperiod}
\end{equation}
where $(\dv_0,\Pv_0)$ and $(\dv,\Pv)$ independently follow distribution $\muv$.

% the main intuition is that the lost sales cost dominates 

Let $\pi^*$ denote a stationary optimal policy, then
\begin{equation}
\begin{split}
\frac{1}{T}\sum_{t=2}^{T+1} \Expt^{\pi^*}[ C_t ] =& \frac{1}{T} \sum_{t=2}^{T+1} \Expt^{\pi^*}[ M(\pi^*(\xv_t)-\xv_t) + L(\pi^*(\xv_t), \dv_t,\Pv_t) ]\\
\geq & \frac{1}{T} \sum_{t=2}^{T+1} \min_{\bslevel} \Expt [  M(\bslevel - \xv_t)  + L(\bslevel, \dv_t, \Pv_t)], 
\end{split}
\label{eq:T_bslevel_optimal}
\end{equation}
where $\{\xv_t\}_{t\geq 2}$ is the sequence of inventory levels generated under the policy $\pi^*$.

We define {$\Gamma := \sum_{i,j}l_{ij}/\sum_{i,j}c_{ij} $}, then for all $\yv,\zv, \zv' \in \Delta_{n-1}$, it holds that
$
\Expt[L(\yv, \dv_t, \Pv_t)] \geq \alpha_0 \sum_{i,j} l_{ij} = \alpha_0 \Gamma  \sum_{i,j} c_{ij} \geq \alpha_0 \Gamma  \Expt[M(\zv' - \zv)] \geq 0,
$
where the last inequality follows directly from the definition of repositioning cost in \eqref{eq:repo_cost} and the fact that the decision variables $\xi_{ij}$ are bounded in $[0,1]$. Combing with \eqref{eq:T_bslevel_optimal}, we have
\begin{equation}
\begin{split}
\frac{1}{T}\sum_{t=2}^{T+1} \Expt^{\pi^*}[ C_t ] &\geq \frac{1}{T} \sum_{t=2}^{T+1} \min_{\bslevel}\Expt\left[  M(\bslevel - \xv^{\bslevel}_0 ) + (1- \alpha_0^{-1} \Gamma^{-1}) L(\bslevel, \dv, \Pv)\right]\\
& \geq (1- \alpha_0^{-1} \Gamma^{-1}) \frac{1}{T} \sum_{t=2}^{T+1} \min_{\bslevel} \Expt \left[  M(\bslevel - \xv^{\bslevel}_0 ) +  L(\bslevel, \dv, \Pv)\right]
\label{eq:t_alpha_gamma}
\end{split}
\end{equation}
where we used the fact that $ \Expt M(\bslevel - \xv_t) \geq 0$ and $\Expt M(\bslevel - \xv^{\bslevel}_0 ) \leq  \alpha_0^{-1} \Gamma^{-1} \Expt[L(\bslevel, \dv, \Pv)]$.

Furthermore, due to the optimality of $\bslevel^*$ that is characterized in \eqref{eq:base_stock_oneperiod},
\[
(1- \alpha_0^{-1} \Gamma^{-1}) \frac{1}{T} \sum_{t=2}^{T+1} \min_{\bslevel}\Expt\left[  M(\bslevel - \xv^{\bslevel}_0 ) +  L(\bslevel, \dv, \Pv)\right] =  (1- \alpha_0^{-1} \Gamma^{-1}) \frac{1}{T} \sum_{t=2}^{T+1} \Expt^{\pi_{\bslevel^*}}[C_t],
\]
and it follows from \eqref{eq:t_alpha_gamma} that
$\DS
\frac{1}{T}\sum_{t=2}^{T+1} \Expt^{\pi^*}[ C_t ]\geq (1- \alpha_0^{-1} \Gamma^{-1}) \frac{1}{T} \sum_{t=2}^{T+1} \Expt^{\pi_{\bslevel^*}}[C_t].
$
Letting $T\rightarrow \infty$, we have
$\DS
1\leq \limsup_{T\rightarrow\infty}  \frac{ \sum_{t=1}^T \Expt^{\policy_{\bslevel^*}}[C_t|\xv_1] }{T\lambda^*   }  \leq \frac{1}{1-\alpha_0^{-1}\Gamma^{-1}},
$
where the left inequality is clear due to the optimality of $\lambda^*$. When  $l_{ij}/c_{ij} \rightarrow \infty$ and thus $\Gamma \rightarrow \infty$, the right hand side ${1}/{(1-\alpha_0^{-1}\Gamma^{-1})}$ goes to $1$. Therefore, we have shown the asymptotic optimality of the best base-stock repositioning policy. \halmos

\subsection{Proof of Theorem~\ref{theorem:asymptotic_optimal_largen}}

\proof{Proof of Theorem~\ref{theorem:asymptotic_optimal_largen}.}

We first examine the probability distribution of each location, $d_{t,i}$. Because demand is assumed to be independent and identically distributed across locations in this theorem, we have
\[
\Expt[d_{t,i}] = \frac{1}{n} \Expt[D_t] = \frac{1}{n}, \mathrm{Var}(d_{t,i}) = \frac{1}{n} \mathrm{Var}(D_t) = \frac{\sigma^2}{n}.
\]

We use $\delta:= \frac{\sigma}{\sqrt{n}}$ to denote the variance of $d_{t,i}$. Based on the assumption, we have that
\begin{equation}
\Prob \left(d_{t,i}\geq \frac{1}{n} +\delta  \right) \geq p_0>0.
\label{eq:tail_bd}
\end{equation}

For any $\yv\in\Delta_{n-1}$, the expected lost sales cost  has the following bound,
\begin{align}
\Expt[L(\yv, \dv_t,\Pv_t)]& = \sum_{i=1}^n \sum_{j=1}^n l_{ij} \cdot P_{t,ij}\Expt[(d_{t,i} - y_i)^+] \\
&  \geq \sum_{i=1}^n \sum_{j=1}^n l_0 P_{t,ij} \cdot \left(\frac{1}{n} +\delta - y_i\right) \cdot \Prob \left( d_{t,i} \geq \frac{1}{n}+\delta\right) \label{eq:l0} \\
& \geq l_0 \sum_{i=1}^n\sum_{j=1}^n P_{t,ij}\left(\frac{1}{n} +\delta - y_i\right) p_0\label{eq:use_tailbd}\\
& = l_0 \sum_{i=1}^n \left(\frac{1}{n} +\delta - y_i\right) p_0 \label{eq:P_ij_sum}\\
& = l_0 n \delta p_0 . \label{eq:sum_y}
\end{align}
In \eqref{eq:l0}, $l_0: = \min_{i,j}l_{ij}$ is the smallest unit lost sales cost; in \eqref{eq:use_tailbd} we invoke the inequality in \eqref{eq:tail_bd}; in \eqref{eq:P_ij_sum} we use the fact that the sum of probability $\sum_{j=1}^n P_{t,ij}$ is $1$; in \eqref{eq:sum_y} we use the fact that $\yv \in\Delta_{n-1}$ and $\sum_{i} y_i = 1$.

For any $\yv, \zv\in \Delta_{n-1}$, the repositioning cost 
\begin{equation}
M(\yv- \zv) \leq  M(\yv- \mathbf{1}_1) + M(\mathbf{1}_1-\zv) \leq 2 c_{\mathrm{M}},
\label{eq:repo_cost_ub}
\end{equation}
where we use the sub-additivity of the repositioning cost function, $\mathbf{1}_1$ denotes the inventory level that sets all inventory of size $1$ at location $1$, and $c_{\mathrm{M}}:= \max_{i,j} c_{ij}$ is the largest unit repositioning cost.

Similarly to the proof of Theorem~\ref{thm:asymptotic_optimal}, we use the following observation on the best base-stock repositioning policy $\bslevel^*$. Under the base-stock repositioning policy, the costs across time periods are all independent and identically distributed except the first period. Therefore, we can equivalently characterize the optimal base-stock level $\bslevel^*$ as follows, 
\begin{equation}
\begin{split}
    \bslevel^* \in \arg\,\min_{\bslevel} &\quad \Expt \left[  M(\bslevel - \xv^\bslevel_0) + L(\bslevel, \dv, \Pv)\right],\\
    \text{s.t.} &\quad \xv^\bslevel_0 = (\bslevel - \dv_0)^+ + \Pv_0^\T \min \{\bslevel, \dv_0\},
\end{split}
\label{eq:base_stock_oneperiod_largen}
\end{equation}
where $(\dv_0,\Pv_0)$ and $(\dv,\Pv)$ independently follow distribution $\muv$.

Therefore, for any stationary optimal policy $\pi^*$, we have
\begin{align}
 \sum_{t=1}^T \Expt^{\pi^*} \left[ L(\yv_t, \dv_t,\boldsymbol{P_t})+M(\yv_t-\xv_t) \right] & \geq  \sum_{t=1}^T \min_{\bslevel} \Expt \left[ L(\bslevel, \dv_t,\boldsymbol{P_t})+M(\bslevel-\xv_t) \right] \label{eq:cost_rewrite}\\
 & \geq \sum_{t=1}^T \min_{\bslevel} \Expt \left[ (1 - \frac{2c_{\mathrm{M}}}{l_0n\delta p_0} )L(\bslevel, \dv_t,\boldsymbol{P_t})+ M(\bslevel-\xv_0^\bslevel) \right] \label{eq:repo_lessthan_lost}
\end{align}
where  in \eqref{eq:cost_rewrite}  $\{\xv_t\}_{t\geq 2}$ is the sequence of inventory levels generated under the policy $\pi^*$, and in \eqref{eq:repo_lessthan_lost} $ \xv^{\bslevel}_0 = (\bslevel - \dv_0)^+ + \Pv_0^\T \min \{\bslevel, \dv_0\}$  is defined as in \eqref{eq:base_stock_oneperiod_largen}. In \eqref{eq:repo_lessthan_lost}, we also use the two inequalities in \eqref{eq:sum_y} and \eqref{eq:repo_cost_ub} to obtain that 
\[
\frac{2c_{\mathrm{M}}}{l_0n\delta p_0}\Expt[L(\bslevel,\dv_t,\Pv_t)] \geq M(\bslevel - \xv_0^\bslevel)
\]
as well as the fact that $M(\bslevel - \xv_t)\geq 0$.

Furthermore, for any $\bslevel$,
\[
 M(\bslevel - \xv^{\bslevel}_0 ) + (1 -  \frac{2c_{\mathrm{M}}}{l_0n\delta p_0} ) L(\bslevel, \dv, \Pv) \geq (1 - 18 \frac{c_{\mathrm{M}}}{l_0n\delta} )  \Expt^{\pi_{\bslevel}}[C_t],
\]
and since the limit holds for arbitrary $T$,
\begin{equation}
\limsup_{T\rightarrow \infty} \frac{1}{T}\sum_{t=1}^T \Expt^{\pi^*}[C_t|\xv_1] \geq (1 -  \frac{2c_{\mathrm{M}}}{l_0n\delta p_0} ) \limsup_{T\rightarrow\infty} \frac{1}{T} \sum_{t=1}^T \Expt^{\policy_{\bslevel^*}}[C_t|\xv_1] .
\label{eq:costs_compare_largen}
\end{equation}
When  the number of locations $n$ goes to infinity, $ \frac{2c_{\mathrm{M}}}{l_0n\delta p_0}  = \frac{2c_{\mathrm{M}}}{l_0\sqrt{n} \sigma p_0} $ approaches to $0$ and thus the right-hand side of \eqref{eq:costs_compare_largen} approaches $\sum_{t=1}^T \Expt^{\policy_{\bslevel^*}}[C_t|\xv_1]$. Therefore, we have shown the asymptotic optimality of the best base-stock repositioning policy. \halmos

\section{Proofs for Best Stock Policy Computation}
\label{subsec:proof_reformulation}

\subsection{Proof of Proposition \ref{prop:offline_MIP}}

\proof{Proof of Proposition \ref{prop:offline_MIP}.}
For ease of exposition, we first assume $d_{1,i}\leq\dots\leq d_{t,i}$ for all $i=1,\dots,n$. We will later address the sorting of $d_{t,i}$ by incorporating permutation matrices in the reformulation. 

We claim that the offline problem \eqref{opt:offline} can be represented by the following mixed integer linear programming problem.
\begin{align}
    \min\  & \sum\limits_{s=1}^{t}\sum_{i=1}^n \sum_{j=1}^n  c_{ij} \xi_{s,ij}  
    -\sum\limits_{s=1}^{t} \sum\limits_{i=1}^{n}\sum\limits_{j=1}^{n}l_{ij} P_{s,ij}m_{s,i}
    +\sum\limits_{s=1}^{t} \sum\limits_{i=1}^{n}\sum\limits_{j=1}^{n}l_{ij} P_{s,ij}d_{s,i}\label{opt:ofl_obj}\\
    \text{subject to}\ &\sum_{i=1}^n \xi_{s,ij} - \sum_{k=1}^n \xi_{s,jk} = m_{s,j}-\sum\limits_{i=1}^{n} P_{s,ij}m_{s,i}, \text{ for all } j = 1,\dots, n \text{ and $s=1,\dots,t$,} \label{opt:ofl_xi1}\\
    & \xi_{s,ij} \geq 0, \forall i = 1,\dots, n, \text{ for all } j = 1,\dots, n\text{ and $s=1,\dots,t$,}\label{opt:ofl_xi2}\\
    &\sum\limits_{i=1}^{n} S_i=1,\label{opt:ofl_s1}\\
    &\bslevel=\{S_i\}_{i=1}^{n}\in[0,1]^{n} ,\label{opt:ofl_s2}\\        
    &\sum \limits_{s=1}^{t} z_{s+1,i}\cdot d_{s,i}\leq  S_i
    \leq \sum\limits_{s=1}^{t} z_{s,i}\cdot d_{s,i}+z_{t+1,i}, \text{ for all $i=1,\dots,n$,}\label{opt:ofl_s3}\\
    &-2(1-z_{s',i})\leq m_{s,i}-S_i\leq 2(1-z_{s',i}), \text{for all $1\leq s'\leq s\leq t$ and $i=1,..,n$}\label{opt:ofl_m1}\\
    &-2(1-z_{s',i})\leq m_{s,i}-d_{s,i}\leq 2(1-z_{s',i}), \text{for all $1\leq s < s'\leq t+1$ and $i=1,..,n$}\label{opt:ofl_m2}\\
    &\sum\limits_{s=1}^{t+1} z_{s,i} = 1, \text{ for all $i=1,\dots,n$,}\label{opt:ofl_z1}\\
    &\zv_{s}=\{z_{s,i}\}_{i=1}^{n}\in\{0,1\}^n, \text{ for all $s=1,\dots,t+1$}\label{opt:ofl_z2},
\end{align}
where decision variables are $\xi_{s,ij}$, $m_{s,i}$, $S_i$, and  $z_{s,i}$ for $s=1,\cdots,t$ and $i,j=1,\cdots,n$. To see this, assume $m_{s,i}=\min(S_i,d_{s,i})$, which will be shown later. We then have the first term in the objective \eqref{opt:ofl_obj} and constraints \eqref{opt:ofl_xi1} and \eqref{opt:ofl_xi2} represent the network flow cost $M(\cdot)$ at time $s=1,\dots,t$; the second and third terms in the objective correspond to the lost sale cost. Consequently, if $m_{s,i}=\min(S_i,d_{s,i})$ holds, we have the objective function of \eqref{opt:offline} is the as as \eqref{opt:ofl_obj}. 

Now, we check that the constraints of \eqref{opt:offline} are the same as \eqref{opt:ofl_xi1} to \eqref{opt:ofl_z2}. Specifically, constraints \eqref{opt:ofl_s1} and \eqref{opt:ofl_s2} correspond to the constraint $\bslevel\in\Delta_{n-1}$. 

Next, we show that constraints \eqref{opt:ofl_s3} to \eqref{opt:ofl_z2} are equivalent to $m_{s,i}=\min(S_i,d_{s,i})$ for any $\bslevel,d_{s,i}\in[0,1]^{n}$ and all $s=1,...,t$, $i=1,...,n$. Without loss of generality, we only show the statement for $s=1,i=1$, and others can be shown by a similar analysis. Particularly, we first show that constraints \eqref{opt:ofl_s3} to \eqref{opt:ofl_z2} imply $m_{1,1}=\min(S_1,d_{1,1})$. From \eqref{opt:ofl_z1} and \eqref{opt:ofl_z2}, we have exactly one element in $\{z_{s,1}\}_{s=1}^{t+1}$ is $1$. From \eqref{opt:ofl_s3}, we have $S_i\in[d_{s-1,1},d_{s,1}]$ if $z_{s,1}=1$ for all $s=1,...,t+1$. Thus, on the one hand, if $z_{1,1}=1$, we have $\min(S_1,d_{1,1})=S_1$, in which case, constraints \eqref{opt:ofl_m1} and \eqref{opt:ofl_m2} imply $m_{1,1}=S_1$; on the other hand, if $z_{s,1}=1$ for some $s>1$, similarly, we have  $\min(S_1,d_{1,1})=d_{1,1}$ and $m_{1,1}=d_{1,1}$. That is, $m_{1,1}=\min(S_1,d_{1,1})$. Then, we show that constraints \eqref{opt:ofl_s3} to \eqref{opt:ofl_z2} are still feasible given $m_{1,1}=\min(S_1,d_{1,1})$. To show this, we only need to verify constraints \eqref{opt:ofl_m1} to \eqref{opt:ofl_z2} hold for $s=1,i=1$. In particular, if $z_{1,i}=1$ and $z_{s,i}=0$ for $s>1$, we have $0\leq S_i\leq d_{1,1}$ and $\min(S_1,d_{1,1})=S_i$, which imply $m_{s,i}=S_i$. In this case, \eqref{opt:ofl_z1} and \eqref{opt:ofl_z2} are satisfied; \eqref{opt:ofl_m1} is $0\leq0\leq0$ for $s'=s=1$ and $i=1$; \eqref{opt:ofl_m2} is $-2\leq m_{1,1}-d_{1,1}\leq2$, which is also satisfied since $m_{1,1},d_{1,1}\in[0,1]$. Thus, combining the above two aspects, we have constraints \eqref{opt:ofl_s3} to \eqref{opt:ofl_z2} can characterize the min function $m_{s,i}=\min(S_i,d_{s,i})$ for any $\bslevel,d_{s,i}\in[0,1]^{n}$ and all $s=1,...,t$, $i=1,...,n$, and we finish the proof. Finally, putting all together, this mixed integer linear programming problem has $nt^2+n^2t+2nt+3n+1$ constraints with $n^2t+2nt+2n$ decision variables.

We now address the case where $d_{1,i}, d_{2,i},\dots,d_{t,i}$ are not necessarily listed in a non-decreasing order for $i = 1,\dots, n$. For each $i$, we introduce a permutation matrix $\per_i$ of size $t \times t$ such that the elements in $\per_i d_{:,i}$ are in non-decreasing order, where $d_{:,i} = (d_{1,i}, d_{2,i},\dots,d_{t,i})^\T$ is a column vector.  It is a well-established fact that the inverse of a permutation matrix is its transpose, i.e., $\per_i^{-1} = \per_i^\T$. The construction is thus completed by leveraging the permutation. \halmos

\subsection{Proof of Proposition \ref{prop:offline_LP}}
\label{subsec:lp_reformulate}
\begin{proposition}[LP Formulation]
    \label{prop:offline_LP}
    Suppose Assumption \ref{assump:lsc>rc} holds for $s = 1,\dots,t$. The offline problem \eqref{opt:offline} can be formulated as the following linear programming problem.
    \begin{align}
        \label{opt:offline_LP}
        \min_{S_i,\xi_{s,ij}, w_{s,i}}\  & \sum\limits_{s=1}^{t}\sum_{i=1}^n \sum_{j=1}^n  c_{ij} \xi_{s,ij}  
        -\sum\limits_{s=1}^{t} \sum\limits_{i=1}^{n}\sum\limits_{j=1}^{n}l_{ij} P_{s,ij}w_{s,i}
        %+\sum\limits_{s=1}^{t} \sum\limits_{i=1}^{n}\sum\limits_{j=1}^{n}l_{ij} P_{s,ij}d_{s,i}
        \\
        \mathrm{subject\  to}\ &\sum_{i=1}^n \xi_{s,ij} - \sum_{k=1}^n \xi_{s,jk} = w_{s,j}-\sum\limits_{i=1}^{n} P_{s,ij}w_{s,i}, \mathrm{\  for\  all \ } j = 1,\dots, n \mathrm{\ and \ }s=1,\dots,t,  \nonumber \\
        & \xi_{s,ij} \geq 0,\  \forall i = 1,\dots, n, \mathrm{ for\  all \ } i,j = 1,\dots, n\mathrm{\  and\ } s=1,\dots,t, \nonumber\\
        &\sum\limits_{i=1}^{n} S_i=1,\ \bslevel=\{S_i\}_{i=1}^{n}\in[0,1]^{n} ,\nonumber\\    
        &w_{s,i}\leq d_{s,i},\ w_{s,i}\leq S_{i}, \ w_{s,i}\geq 0,\ \mathrm{\  for\  all\ } s=1,.\dots,t,i=1,\dots,n. \nonumber
    \end{align}
\end{proposition}
\begin{remark}
We emphasize that Proposition~\ref{prop:offline_LP} does \emph{not} imply that the cost function $\widetilde{C}_t(\xv_t,\yv_t,\dv_t,\Pv_t)$ is convex in $\yv_t$ under Assumption \ref{assump:lsc>rc}. The non-convexity persists under Assumption~\ref{assump:lsc>rc}, which necessitates additional algorithmic design in the online setting and we address this in detail in Section~\ref{section:disentangling}.
\end{remark}

The linear programming formulation \eqref{opt:offline_LP} appears to be a direct translation of the original offline problem \eqref{opt:offline}, but there is a key difference in the characterization of the censored demand $w_{s,i}$. Specifically, the equality $w_{s,i}  = \min\{d_{s,i}, S_i\}$ is replaced with inequality constraints $w_{s,i}\leq d_{s,i},\ w_{s,i}\leq S_{i}, \ w_{s,i}\geq 0$. Note that the original definition $w_{s,i} = \min\{d_{s,i}, S_i\}$ is not linear, and thus cannot be directly included as a constraint in a linear programming problem. The validity of the linear programming reformulation shows that even if the service provider has the flexibility to choose the fulfilled demand $w_{s,i}$, when the cost structure satisfies Assumption~\ref{assump:lsc>rc}, it is always optimal for the service provider to satisfy as much demand as possible, i.e., $w_{s,i} = \min\{d_{s,i}, S_i\}$. 
%The proof of Proposition~\ref{prop:offline_LP} is in Appendix~\ref{subsec:proof_reformulation}.

\proof{Proof of Proposition \ref{prop:offline_LP}.}
        By observing that any feasible repositioning plan is feasible to \eqref{opt:offline_LP}, we only need to show that one optimal solution of \eqref{opt:offline_LP} satisfies
        $w_{s,i}=\min\left\{d_{s,i},S_i \right\}$ for all $s,i$, which can represent a repositioning plan, under the condition $\DS \sum_{i=1}^{n} l_{ji} P_{s,ji} \geq \sum\limits_{i=1}^{n}P_{s,ij}c_{ji}$ for all $j=1,\dots,n$ and $s=1,\dots,t$. If not, suppose $\{S_i', \ \xi_{s,ij}',\ w_{s,i}':i,j=1,\dots,n,s=1,\dots,t\}$ is an optimal solution of \eqref{opt:offline} that satisfies
        $$  
            w_{s',i'}'< \min(d_{s',i'},S_{i'}),
        $$
        for some $s',i'$, and denote $\epsilon=\min(d_{s',i'},S_{i'})-w_{s',i'}$. Then, let 
        \begin{align}
            \label{eq:new_sol}
            \tilde{w}_{s,i} = 
            \begin{cases}
                {w}_{s,i}' + \epsilon, &\text{if } s=s',i=i',\\
                {w}_{s,i}', &\text{otherwise,}
            \end{cases}
            \   
            \tilde{\xi}_{s,ij} = 
            \begin{cases}
                {\xi}_{s,ij}' + P_{s',ji}\cdot \epsilon &\text{if } s=s',j=i',\\
                {\xi}_{s,ij}' & \text{otherwise.}
            \end{cases}
        \end{align}
        Based on this construction, we can verify that $\{S_i', \ \tilde{\xi}_{s,ij}',\ \tilde{w}_{s,i}':i,j=1,\dots,n,s=1,\dots,t\}$ is also an optimal solution of \eqref{opt:offline}. Specifically, we have
        \begin{align}
            \label{eq:feas1_newsol}
            \sum_{i=1}^n \tilde{\xi}_{s',ii'} - \sum_{k=1}^n \tilde{\xi}_{s',i'k} 
            &= 
            \sum_{i=1}^n \xi_{s',ii'} - \sum_{k=1}^n \xi_{s',i'k}+ \sum_{i=1}^{n}P_{s',ii'}\cdot\epsilon\nonumber\\
            &=
            w_{s',i'}-\sum\limits_{i=1}^{n} P_{s',ii'}w_{s,i}+\sum_{i=1}^{n}P_{s',ii'}\cdot\epsilon\\
            &=
            \tilde{w}_{s',i'}-\sum\limits_{i=1}^{n} P_{s',ii'}\tilde{w}_{s,i}\nonumber,
        \end{align}
        where the first and the last lines in \eqref{eq:feas1_newsol} come from the construction \eqref{eq:new_sol}, and the second line in \eqref{eq:feas1_newsol} comes from the first constraint of \eqref{opt:offline_LP} and the feasibility of the solution $\{S_i', \ \xi_{s,ij}',\ w_{s,i}':i,j=1,\dots,n,s=1,\dots,t\}$. Similarly, we have
        \begin{align}
            \label{eq:feas2_newsol}
            \sum_{i=1}^n \tilde{\xi}_{s',ij} - \sum_{k=1}^n \tilde{\xi}_{s',jk}
            = 
            \sum_{i=1}^n \xi_{s',ij} - \sum_{k=1}^n \xi_{s',jk}- P_{s',i'j}\cdot\epsilon
            =
            w_{s',j}-\sum\limits_{i=1}^{n} P_{s',ij}w_{s,i}- P_{s',i'j}\cdot\epsilon
            =
            \tilde{w}_{s',j}-\sum\limits_{i=1}^{n} P_{s',ii'}\tilde{w}_{s,i}.
        \end{align}     
        Now, combining \eqref{eq:feas1_newsol} and \eqref{eq:feas2_newsol}, we can verify that the new solution $\{S_i', \ \xi_{s,ij}',\ w_{s,i}':i,j=1,\dots,n,s=1,\dots,t\}$ is also feasible to \eqref{opt:offline_LP}. Next, we show that this new solution is also optimal by verifying the objective achieved by the new solution is no larger than the optimal objective. In particular,
        \begin{align*}
            &\sum\limits_{s=1}^{t}\sum_{i=1}^n \sum_{j=1}^n  c_{ij} \tilde{\xi}_{s,ij}  
            -\sum\limits_{s=1}^{t} \sum\limits_{i=1}^{n}\sum\limits_{j=1}^{n}l_{ij} P_{s,ij}\tilde{w}_{s,i}\\
            =&
            \sum\limits_{s=1}^{t}\sum_{i=1}^n \sum_{j=1}^n  c_{ij} {\xi}_{s,ij}  
            -\sum\limits_{s=1}^{t} \sum\limits_{i=1}^{n}\sum\limits_{j=1}^{n}l_{ij} P_{s,ij} {w}_{s,i} +\sum\limits_{i=1}^{n}c_{ii'}P_{s',i'i}\cdot\epsilon-\sum\limits_{j=1}^{n}l_{i'j}P_{s',i'j}\cdot\epsilon\\
            \leq&
            \sum\limits_{s=1}^{t}\sum_{i=1}^n \sum_{j=1}^n  c_{ij} {\xi}_{s,ij}  
            -\sum\limits_{s=1}^{t} \sum\limits_{i=1}^{n}\sum\limits_{j=1}^{n}l_{ij} P_{s,ij} {w}_{s,i},
        \end{align*}
        where the inequality comes from the construction \eqref{eq:new_sol} and the second line comes from the condition $\sum_{i=1}^{n} l_{ji} P_{t,ji} \geq \sum\limits_{i=1}^{n}P_{t,ji}c_{ij}$ for all $j=1,\dots,n$ and $t=1,\dots,T$. Thus, through this construction, we can transfer any optimal solution of \eqref{opt:offline_LP} to an optimal solution such that
        $w_{s,i}=\min\left\{d_{s,i},S_i \right\}$ is satisfied for all $s,i$, and, we finish the proof. 
        \halmos

%% file: files/appendix_new_regret.tex
% change the equation numbering in the appendix
\renewcommand{\thesection}{\Alph{section}}
\numberwithin{equation}{section}
%--------------

\section{Generalization Bound}
\label{appendix:generalization_bound}
In this section, we prove the generalization bound that holds for all base-stock repositioning levels uniformly.

 \subsection{Technical Lemmas}
 \label{appendix:subsection_techlem}

\begin{lemma}[Rademacher Complexity]
    \label{lem:rademacher_complexity}
    Let $\mathcal{F}$ be a class of functions $f: \mathcal{X} \rightarrow [a,b]$, and $\{X_t\}_{t=1}^{T}$ be i.i.d. random variables taking values in $\mathcal{X}$. Then the following inequality holds for any $s>0$
    $$
        \mathbb{P}\left(
        \sup_{f\in\mathcal{F}}\left|
        \frac{1}{T}\sum\limits_{t=1}^{T}f(X_t)
        -
        \mathbb{E}[f(X_1)]
        \right|
        \geq
        \mathbb{E}\left[
        \sup_{f\in\mathcal{F}}\left|
        \frac{1}{T}\sum\limits_{t=1}^{T}\sigma_{t}f(X_t)
        \right|
        \right]+s
        \right)
        \leq
        \exp\left(-\frac{2Ts^2}{(b-a)^2}\right),
    $$
    where $\{\sigma_t\}_{t=1}^{T}$ denotes a set of i.i.d. random signs satisfying $\mathbb{P}(\sigma_{t}=1)=\mathbb{P}(\sigma_t=-1)=\frac{1}{2}$.     
\end{lemma}

\proof{Proof of Lemma \ref{lem:rademacher_complexity}.}
    This is a standard result regarding Rademacher Complexity, and we refer to Theorem 4.10 in \cite{wainwright2019high} for the proof.
\halmos
%%%%%%%%%%%%%%%%%%%%%%%%%%%%%%%%%%%%%%%%%%%%%%%%%%%%

\begin{lemma}[Generalized Massart’s Finite Class Bound]   
    \label{lem:massart}
Let $\mathcal{G}$ be a family of functions that are defined on $\mathcal{X}$ and take values in $\{0,+1\}$. Then the following holds:
\begin{align*}
    \mathbb{E}\left[
        \sup_{g\in\mathcal{G}}\left|
        \frac{1}{m}\sum\limits_{i=1}^{m}\sigma_{i}g(X_i)
        \right|
        \right]\le \sqrt{\frac{2\log \Pi_{\mathcal{G}}(m)}{m}},
\end{align*}
where $\{x_1,\dots,x_m\}$ are $n$ points in $\mathcal{X}$, $\{\sigma_i\}_{i=1}^{m}$ is a set of independent uniform distributions on $\{-1,+1\}$, the growth function $\Pi_{\mathcal{G}}(m): \mathbb{N} \rightarrow \mathbb{N}$ for a hypothesis set $\mathcal{G}$ is the maximum number of distinct ways in which $m$ points in $\mathcal{C}$ can be classified using hypotheses in $\mathcal{G}$, i.e.,
\begin{align*}
    \forall m \in \mathbb{N}, \ \Pi_{\mathcal{G}}(m) = \max_{ \{x_1, \dots, x_m\}\subseteq \mathcal{X} } \left|\left\{ \left( g(x_1),\dots, g(x_m)\right): g\in \mathcal{G} \right\} \right|. 
\end{align*}
\end{lemma}
\proof{Proof of Lemma \ref{lem:massart}.}
    This is an upper bound on the Rademacher Complexity for a class of functions that only take finite values, and we refer to Corollary 3.8 in \cite{mohri2018foundations} for the proof.
\halmos

\subsection{Proof of Lemma \ref{lem:h_lipschitz}} 
\label{appendix:lipschitz}

\proof{Proof of Lemma \ref{lem:h_lipschitz}.}
In the new notation, the Lipschitz property is equivalent to 
\begin{align*}
        \left\vert
            h(\yv, \Pv)-h(\yv', \Pv')
        \right\vert
        \leq
        n^2\cdot (2\max_{i,j}c_{ij}+\max_{i,j}l_{ij}) \cdot(\|\yv-\yv'\|_2+\|\Pv-\Pv'\|_{F}),
    \end{align*}
  In particular, for any $\yv=(y_1,\dots,y_n)^{\top},\yv'=(y_1',\dots,y_n')^{\top}\in[0,1]^{n}$, and probability transition matrices $\Pv=\{P_{ij}\}_{i,j=1}^{n},\Pv'=\{P'_{ij}\}_{i,j=1}^{n}\in[0,1]^{n\times n}$,
    \begin{align}
    \label{ieq:h_diff}
        \left\vert
        h(\yv,\Pv)-h(\yv',\Pv')
        \right\vert
        &=
        \left\vert
        M\left((\boldsymbol{I}-\Pv^{\top})\yv\right)
        +
        \sum\limits_{i=1}^{n} \sum\limits_{j=1}^{n} l_{ij}\cdot P_{ij}y_i
        -
        M\left((\boldsymbol{I}-(\Pv')^{\top})\yv'\right)
        -
        \sum\limits_{i=1}^{n} \sum\limits_{j=1}^{n} l_{ij}\cdot P'_{ij}y_i'
        \right\vert\nonumber\\
        &\leq
        \left\vert
        M\left((\boldsymbol{I}-\Pv^{\top})\yv\right)
        -
        M\left((\boldsymbol{I}-(\Pv')^{\top})\yv'\right)
        \right\vert
        +
        \left\vert
        \sum\limits_{i=1}^{n}\sum\limits_{j=1}^{n}l_{ij}\cdot \left(P_{ij}y_i-P'_{ij}y_i'\right)
        \right\vert\\
        &\leq
        2 \max_{i,j} c_{ij} \cdot \|
        (\boldsymbol{I}-\Pv^{\top})\yv
        -
        (\boldsymbol{I}-(\Pv')^{\top})\yv'
        \|_1
        +
        \left\vert
       \sum\limits_{i=1}^{n} \sum\limits_{j=1}^{n} l_{ij}\cdot \left(P_{ij}y_i-P'_{ij}y_i'\right)
        \right\vert,\nonumber
    \end{align}
    where the first line comes from the definition of $h$, i.e., \eqref{def:h}, the second line comes from the triangle inequality of the absolute value function, and the last line is due to the properties of the repositioning cost. That is, $|M(\xv_1) - M(\xv_2)| \leq M(\xv_1-\xv_2)$ and $M(\xv) \leq 2 \max_{ij}{c_{ij}} \Vert \xv\Vert_1$. We next bound the right-hand side in \eqref{ieq:h_diff}. For the first term in the right-hand side of \eqref{ieq:h_diff}, we have
    \begin{align}
        \label{ieq:h_diff_t1}
        \|
        (\boldsymbol{I}-\Pv^{\top})\yv
        -
        (\boldsymbol{I}-(\Pv')^{\top})\yv'
        \|_1
        &\leq
        \sqrt{n}\|
        (\boldsymbol{I}-\Pv^{\top})\yv
        -
        (\boldsymbol{I}-(\Pv')^{\top})\yv'
        \|_2\nonumber\\
        &\leq
        \sqrt{n}\|
        (\boldsymbol{I}-\Pv^{\top})(\yv-\yv')
        \|_2
        +
        \sqrt{n}\|
        (\Pv-\Pv')^{\top}\yv'
        \|_2\\
        &\leq 
        n^{3/2}(\|y-y'\|_2
        +\|\Pv-\Pv'\|_{F}),\nonumber
    \end{align}
    where $\|\Xv\|_{F}=\sqrt{\sum\limits_{i,j=1}^{n}X_{ij}^2} \textstyle $ denotes the Frobenius norm for any $\Xv\in\mathbb{R}^{n\times n}$. Here, the first inequality is obtained by the relation between 1-norm and 2-norm $\|\yv\|_1\leq\sqrt{n}\|\yv\|_2$, the second inequality is obtained by the triangle inequality, and the last line is obtained by matrix-vector inequalities and the boundedness of $\yv$ and $\Pv$. 
    
    For the second term in the right-hand side of \eqref{ieq:h_diff}, 
    \begin{align}
        \label{ieq:h_diff_t2}
        \left\vert
        \sum\limits_{i=1}^{n} \sum\limits_{j=1}^{n}l_{ij}\cdot \left(P_{ij}y_i-P'_{ij}y_i'\right)
        \right\vert
        &\leq
        n\cdot\max_{i,j} l_{ij}\cdot\|
            \Pv^\top\text{diag}(\yv)-(\Pv')^{\top}\text{diag}(\yv)'
        \|_{F}\nonumber\\
        &\leq
        n\cdot\max_{i,j} l_{ij}\cdot\left(
            \|
            \Pv^{\top}(\text{diag}(\yv)-\text{diag}(\yv)')
        \|_{F}
        +
        \|
            (\Pv-\Pv')^{\top}\text{diag}(\yv')
        \|_{F}
        \right)\\
        &\leq
        n^2\cdot\max_{i,j} l_{ij}\cdot\left(
        \|    
            \yv-\yv'
        \|_{2}
        +
        \|\Pv-\Pv'\|_{F}
        \right)\nonumber,
    \end{align}
    where $\text{diag}(\yv)$ denotes the square diagonal matrix with the elements of vector $\yv$ on the main diagonal. For the above inequalities, the first inequality comes from Cauchy's inequality, the second inequality comes from the triangle inequality, and the last line comes from the property of the Frobenius norm and the boundedness of $\yv$ and $\Pv$. Then, plugging inequalities \eqref{ieq:h_diff_t1} and \eqref{ieq:h_diff_t2} into \eqref{ieq:h_diff}, we have 
    \begin{align*}
        \left\vert
            h(\yv, \Pv)-h(\yv', \Pv')
        \right\vert
        \leq
        n^2\cdot (2\max_{i,j}c_{ij}+\max_{i,j}l_{ij}) \cdot(\|\yv-\yv'\|_2+\|\Pv-\Pv'\|_{F}),
    \end{align*}
    for any $\yv,\yv'=\in[0,1]^{n}$, and probability transition matrices $\Pv,\Pv'\in[0,1]^{n\times n}$.
    That is, $h$ is a Lipschitz-continuous function with Lipschitz constant $2n^2\cdot(\max_{i,j}c_{ij}+\max_{i,j}l_{ij})$.
    \halmos

\subsection{Proof of Proposition \ref{prop:offline_concentration}}
\label{appendix:proof_of_lipschitz_uniform}
    Then, by leveraging the Lipschitz property in Lemma \ref{lem:h_lipschitz} and technical lemmas in Appendix~\ref{appendix:subsection_techlem}, we can show the generalization bound for any base-stock repositioning level $\bslevel\in\Delta_{n-1}$ as below.
    
% \todo{concentration, how we can use sample to get some uniform convergence but we need better sampling, before we use discretization to point by point learn the objective}

    \proof{Proof of Proposition \ref{prop:offline_concentration}.}
        The main tool we use to derive the generalization bound is Rademacher complexity. However, computing and bounding Rademacher complexity of our problem setting as it involves vector-valued functions. To tackle this difficulty, In the following, we will leverage the technical results in Lemma \ref{lem:rademacher_complexity}, Lemma \ref{lem:contraction}, and Lemma \ref{lem:massart}.
        
        We first apply Lemma \ref{lem:rademacher_complexity} to obtain the form of the generalization bound. Specifically, consider the function class $\mathcal{F}=\{\fv_{\bslevel}:\bslevel\in\Delta_{n-1}\}$, where $\fv_{\bslevel}$ is defined by \eqref{def:h}. Then, we have
        \begin{align}
            \label{ieq:gen}
            &\sup_{\bslevel\in\Delta_{n-1}} \left\vert
                \frac{1}{t}\sum\limits_{s=1}^{t}C_s(\xv_{s+1}^{\bslevel},\bslevel,\dv_s,\Pv_s)
                -
                \mathbb{E}[{C}_1(\xv_1^{\bslevel},\bslevel,\dv_1,\Pv_1)]
            \right\vert\\
            \leq&
            \sup_{\fv_{\bslevel}\in\mathcal{F}}
            \left|
                \frac{1}{t}  \sum\limits_{s=1}^{t}h(\fv_{\bslevel}(\dv_s,\Pv_s))
                -
                \mathbb{E}[h(\fv_{\bslevel}(\dv_1,\Pv_1))]
            \right|
            +
            \left\vert
                \frac{1}{t}\sum\limits_{s=1}^{t}\sum\limits_{i,j=1}^{n}l_{ij}P_{s,ij}d_{s,i}-\mathbb{E}\left[\sum\limits_{i,j=1}^{n}l_{ij}P_{s,ij}d_{s,i}\right]\right\vert\nonumber
        \end{align}
        by the triangle inequality. Regarding the first term in the right-hand-side of \eqref{ieq:gen}, by  Lemma \ref{lem:rademacher_complexity},
        \begin{align}
      \label{ieq:gen1}
            &\sup_{\fv_{\bslevel}\in\mathcal{F}}
            \left|
                \frac{1}{t}  \sum\limits_{s=1}^{t}h(\fv_{\bslevel}(\dv_s,\Pv_s))
                -
                \mathbb{E}[h(\fv_{\bslevel}(\dv_1,\Pv_1))]
            \right|
            \\
            \leq&
            \mathbb{E}\left[
                \sup_{\fv_{\bslevel}\in\mathcal{F}}
                \left\vert
                    \frac{1}{t} \sum\limits_{s=1}^{t} \sigma_s h(\fv_{\bslevel}(\dv_s,\Pv_s))
                \right\vert
            \right]
            +
            2\left(\max_{i,j}c_{ij}+\max_{i,j}l_{ij}\right)\frac{ \sqrt{\log T}}{\sqrt{t}},\nonumber
        \end{align}
        holds with probability no less than $1-\frac{1}{T^2}$, where $\{\sigma_s\}_{s=1}^{t}$ is a set of independent uniform random variables on $\{-1,1\}$. 
        Here, we note that since the second term is negligible in the final concentration bound, the proved result here also holds for the modified costs $\widetilde{C}$.
        For the second term in \eqref{ieq:gen}, by Hoeffding's inequality, we have
        \begin{align}
            \label{ieq:gen2}
            \left\vert
                \frac{1}{t}\sum\limits_{s=1}^{t}\sum\limits_{i,j=1}^{n}l_{ij}P_{s,ij}d_{s,i}-\mathbb{E}\left[\sum\limits_{i,j=1}^{n}l_{ij}P_{s,ij}d_{s,i}\right]\right\vert
                \leq
                n\max_{i,j}l_{ij} \cdot \frac{\sqrt{\log T}}{\sqrt{t}}
        \end{align}
        holds with probability no less than $1-\frac{2}{T^2}$. Then, plugging \eqref{ieq:gen1} and \eqref{ieq:gen2} into \eqref{ieq:gen}, we have
        \begin{align}
            \label{ieq:offline_concentration_rada}
            &\sup_{\bslevel\in\Delta_{n-1}} \left\vert
                \frac{1}{t}\sum\limits_{s=1}^{t}C_s(\xv_{s}^{\bslevel},\bslevel,\dv_s,\Pv_s)
                -
                \mathbb{E}[{C}_1(\xv_1^{\bslevel},\bslevel,\dv_1,\Pv_1)]
            \right\vert\\
            \leq
            &
            \mathbb{E}\left[
                \sup_{\fv_{\bslevel}\in\mathcal{F}}
                \left\vert
                    \frac{1}{t} \sum\limits_{s=1}^{t} \sigma_s h(\fv_{\bslevel}(\dv_s,\Pv_s))
                \right\vert
            \right]
            +
            2n\left(\max_{i,j}c_{ij}+\max_{i,j}l_{ij}\right)\frac{ \sqrt{\log T}}{\sqrt{t}}\nonumber
        \end{align}
        holds with probability no less than $1-\frac{3}{T^2}$.

        Next, we bound the first term on the right-hand side of \eqref{ieq:offline_concentration_rada} by the contraction lemma (Lemma \ref{lem:contraction}). Recall the definition of $\fv_{\bslevel}$ that      \begin{align*}
            \fv_{\bslevel}(\dv,\Pv) = (\min
            \{\dv,\bslevel\},\Pv)
        \end{align*} 
        for any  $\bslevel,\dv=\{d_i\}_{i=1}^{n}\in\Delta_{n-1}$ and transition probability matrix $\Pv=\{P_{ij}\}_{i,j=1}^{n}\in[0,1]^{n\times n}$. Denote 
        \begin{align*}
            f_{\bslevel,k}(\dv,\Pv)
            =
            \begin{cases}
                d_k, &\text{ if $k=1,\dots,n$}\\
                P_{ij}, &\text{ if $k=n+1,\dots,n(n+1)$ and $ni+j=k-n$}
            \end{cases}
        \end{align*}
        as the $k$-th entry of $\fv_{\bslevel}$.
        
        Then, based on the Lipschitzness of $h$ shown in Lemma \ref{lem:h_lipschitz}, we can apply Lemma \ref{lem:contraction} and have 
        \begin{align}
            \label{ieq:h_contraction}
            \mathbb{E}\left[
                \sup_{\fv_{\bslevel}\in\mathcal{F}}
                \left\vert
                    \frac{1}{t} \sum\limits_{s=1}^{t} \sigma_s h(\fv_{\bslevel}(\dv_s,\Pv_s))
                \right\vert
            \right]
            \leq
            2\sqrt{2}n^2\left(\max_{i,j}c_{ij} + \max_{i,j}l_{ij}\right)
            \cdot
            \mathbb{E}\left[
                \sup_{\fv_{\bslevel}\in\mathcal{F}
                }
                    \frac{1}{t}
                    \sum\limits_{s=1}^{t} \sum\limits_{k=1}^{n(n+1)} \sigma_{s,k} f_{\bslevel,k}(\dv_s,\Pv_s)
            \right],
        \end{align}
        where $\sigma_{s,k}$'s are independent uniform random variables on $\{-1,1\}$ for $k=1,\dots,n(n+1)$ and $s=1,\dots,t$. To see this,
        \begin{align*}
            \mathbb{E}\left[
                \sup_{\fv_{\bslevel}\in\mathcal{F}}
                \left\vert
                    \frac{1}{t} \sum\limits_{s=1}^{t} \sigma_s h(\fv_{\bslevel}(\dv_s,\Pv_s))
                \right\vert
            \right]
            &\leq
            \mathbb{E}\left[
                \left\vert
                \sup_{\fv_{\bslevel}\in\mathcal{F}}
                    \frac{1}{t} \sum\limits_{s=1}^{t} \sigma_s h(\fv_{\bslevel}(\dv_s,\Pv_s))
                \right\vert
                +
                \left\vert
                \sup_{\fv_{\bslevel}\in\mathcal{F}}
                    \frac{1}{t} \sum\limits_{s=1}^{t} -\sigma_s h(\fv_{\bslevel}(\dv_s,\Pv_s))
                \right\vert
            \right]\nonumber\\
            &
            =
            2\mathbb{E}\left[
                \left\vert
                \sup_{\fv_{\bslevel} \in\mathcal{F}}
                    \frac{1}{t} \sum\limits_{s=1}^{t} \sigma_s h(\fv_{\bslevel}(\dv_s,\Pv_s))
                \right\vert
            \right]\\
            &=
            2\mathbb{E}\left[
                \sup_{\fv_{\bslevel} \in\mathcal{F}}
                    \frac{1}{t} \sum\limits_{s=1}^{t} \sigma_s h(\fv_{\bslevel}(\dv_s,\Pv_s))
            \right]            \\
            &\leq 
            2\sqrt{2}n^2\left(\max_{i,j}c_{ij} + \max_{i,j}l_{ij}\right)
            \cdot
            \mathbb{E}\left[
                \sup_{\fv_{\bslevel} \in\mathcal{F}
                }
                    \frac{1}{t}
                    \sum\limits_{s=1}^{t} \sum\limits_{k=1}^{n(n+1)} \sigma_{s,k} f_{\bslevel,k}(\dv_s,\Pv_s)
            \right].          
        \end{align*}
        Here, the first line comes from the property of the supremum function, the second line comes from the fact that $\{\sigma_s\}_{s=1}^{t}$ shares the same distribution with $\{-\sigma_s\}_{s=1}^{t}$, and the last line comes from Lemma \ref{lem:contraction}. We remark that the third line can hold without loss of generality by enlarging the function class $\mathcal{F}$ with an additional mapping $\fv_0(\dv,\Pv)=(\boldsymbol{0},\Pv)$, where $\boldsymbol{0}\in\mathbb{R}^n$ denotes the all zero vector. After this modification, $\sup_{\fv_{\bslevel}\in\mathcal{F}}
                    \frac{1}{t} \sum\limits_{s=1}^{t} \sigma_s h(\fv_{\bslevel}(\dv_s,\Pv_s))$
        will always be non-negative for any realized samples $\{(\dv_s,\Pv_s)\}_{s=1}^{t}$ so that the absolute value function can be dropped from the second line to the third line.
        
        In the following, we give an upper bound of $\mathbb{E}\left[
                \sup_{\fv_{\bslevel} \in\mathcal{F}
                }
                    \frac{1}{t}
                    \sum\limits_{s=1}^{t} \sum\limits_{k=1}^{n(n+1)} \sigma_{s,k} f_{\bslevel,k}(\dv_s,\Pv_s)
            \right]$. 
        Particularly, we have
        \begin{align}
            \label{ieq:F_rademacher}
            \mathbb{E}\left[
                \sup_{\fv_{\bslevel} \in\mathcal{F}
                }
                    \frac{1}{t}
                    \sum\limits_{s=1}^{t} \sum\limits_{k=1}^{n(n+1)} \sigma_{s,k} f_{\bslevel,k}(\dv_s,\Pv_s)
            \right]
            &\leq
            \sum\limits_{k=1}^{n(n+1)} 
            \mathbb{E}\left[
                \sup_{\fv_{\bslevel} \in\mathcal{F}
                }
                    \frac{1}{t}
                    \sum\limits_{s=1}^{t} \sigma_{s,k} f_{\bslevel,k}(\dv_s,\Pv_s)
            \right]\nonumber\\
            &=
            \sum\limits_{k=1}^{n} 
            \mathbb{E}\left[
                \sup_{\fv_{\bslevel}\in\mathcal{F}
                }
                    \frac{1}{t}
                    \sum\limits_{s=1}^{t} \sigma_{s,k} f_{\bslevel,k}(\dv_s,\Pv_s)
            \right]\\
            &=
            \sum\limits_{k=1}^{n} 
            \mathbb{E}\left[
                \sup_{{\bslevel}\in\Delta_{n-1}
                }
                    \frac{1}{t}
                    \sum\limits_{s=1}^{t} \sigma_{s,k} \min\{S_k,d_{s,k}\}
            \right],\nonumber
        \end{align}
        where $S_k,\ d_{s,k}$ denote the $k$-th entry of $\bslevel,\ \dv_s$, respectively, for all $k=1,\dots,n$ and $s=1,\dots,t$.
        In the above equalities and inequality, the first one comes from the property of the supremum function, the second line comes from the fact that for any $k=n+1,\dots,n(n+1)$,  $\{f_{\bslevel,k}(\dv_s,\Pv_s)\}_{\bslevel\in\Delta_{n-1}}$ is a singleton so that  $\mathbb{E}\left[
                \sup_{\fv_{\bslevel} \in\mathcal{F}
                }
                    \frac{1}{t}
                    \sum\limits_{s=1}^{t} \sigma_{s,k} f_{\bslevel,k}(\dv_s,\Pv_s)
            \right]=0$, and the last line comes from the definition of $\fv_{\bslevel}$. 
            
            In addition, notice that there are at most $t$ different elements in $\{\mathbbm{1}_{\{S_k>d_{1,k}\}},\dots,\mathbbm{1}_{\{S_k>d_{t,k}\}}\}$
            for any $k=1,\dots,n$ and fixed samples $\{(\dv_s,\Pv_s)\}_{s=1}^{t}$. Thus, by Lemma \ref{lem:massart}, we have
            \begin{align}
                &\mathbb{E}\left[
                \sup_{{\bslevel}\in\Delta_{n-1}
                }
                    \frac{1}{t}
                    \sum\limits_{s=1}^{t} \sigma_{s,k} \min(S_k,d_{s,k})
            \right]\nonumber\\
            \leq&
            \mathbb{E}\left[
                \sup_{{\bslevel}\in\Delta_{n-1}
                }
                    \frac{1}{t}
                    \sum\limits_{s=1}^{t} \sigma_{s,k} d_{s,k}\mathbbm{1}_{\{S_k\leq d_{s,k}\}}
            \right]+
            \mathbb{E}\left[
                \sup_{{\bslevel}\in\Delta_{n-1}
                }
                    \frac{S_k}{t}
                    \sum\limits_{s=1}^{t} \sigma_{s,k} \mathbbm{1}_{\{S_k> d_{s,k}\}}
            \right]\label{ieq:F_rademacher2}\\
            \leq&
                        \mathbb{E}\left[
                \sup_{{\bslevel}\in\Delta_{n-1}
                }
                    \frac{1}{t}
                    \sum\limits_{s=1}^{t} \sigma_{s,k} d_{s,k}\mathbbm{1}_{\{S_k\leq d_{s,k}\}}
            \right]+
            \mathbb{E}\left[
                \sup_{{\bslevel}\in\Delta_{n-1}
                }
                    \frac{1}{t}
                    \sum\limits_{s=1}^{t} \sigma_{s,k} \mathbbm{1}_{\{S_k> d_{s,k}\}}
            \right]\nonumber\\
            \leq&
            \frac{2\sqrt{2\log T}}{\sqrt{t}},\nonumber
            \end{align} for any $k=1,\dots,n$. Here, the first inequality comes from the triangle inequality, the second inequality comes from the non-negativity of the second term, and the last line comes from Lemma \ref{lem:massart}.

            Finally, combining inequalities \eqref{ieq:offline_concentration_rada}, \eqref{ieq:h_contraction}, \eqref{ieq:F_rademacher} and \eqref{ieq:F_rademacher2}, we can draw the generalization bound below holds with probability no less than $1-\frac{3}{T^2}$
        \begin{align*}
            \sup_{\bslevel\in\Delta_{n-1}} \left\vert
                \frac{1}{t}\sum\limits_{s=1}^{t}C_s(\xv_{s}^{\bslevel},\bslevel,\dv_s,\Pv_s)
                -
                \mathbb{E}[{C}_1(\xv_1^{\bslevel},\bslevel,\dv_1,\Pv_1)]
            \right\vert
            \leq
            10n^3 \left( \max_{i,j}c_{ij} + \max_{i,j}l_{ij} \right)\cdot \frac{\sqrt{\log T}}{\sqrt{t}}.
        \end{align*}
    \halmos

\section{Analysis of \pcr{SOAR} Algorithm}
\label{sec:appd_gd}

\subsection{Proof of Lemma \ref{lem:disentangle}}
\proof{Proof of Lemma \ref{lem:disentangle}.}
 To prove the lemma, we need to show \eqref{eq:disentangle}, i.e., 
    \begin{align*}
        \left|\sum\limits_{t=1}^{T}\widetilde{C}_t(\xv_t,\yv_t,\dv_t,\Pv_t)
        -
        \sum\limits_{t=1}^{T}\widetilde{C}_t(\xv_{t+1},\yv_t,\dv_t,\Pv_t)\right|
        \leq
        2\cdot\left(\max_{i,j=1,\dots,n} c_{ij}\right)\cdot\sum\limits_{t=1}^{T}\|\yv_{t}-\yv_{t-1}\|_1.
    \end{align*}
By definition, we have
$
        \widetilde{C}_t(\xv_{t}, \yv_t, \dv_t, \Pv_t) = M(\yv_t - \xv_{t}) - \sum_{i=1}^n\sum_{j=1}^nl_{ij} P_{t,ij}\min\{d_{t,i}, y_{t,i}\},
$
and
$
        \widetilde{C}_t(\xv_{t+1}, \yv_t, \dv_t, \Pv_t) = M(\yv_t - \xv_{t+1}) - \sum_{i=1}^n\sum_{j=1}^nl_{ij} P_{t,ij}\min\{d_{t,i}, y_{t,i}\}. $
In particular,
$
\yv_t - \xv_{t+1} = (\mathbf{I}-\Pv_t) \min\{y_t,\dv_t\},
$
so it is clear that the relabeled modified cost $\widetilde{C}_t(\xv_{t+1}(\yv_t),\yv_t,\dv_t,\Pv_t)$ depends only on the repositioning policy and realized demands and transition matrix at time $t$, for all $t=1,\dots,T$. To obtain the bound, we have
\begin{align}
        \left|\sum_{t=1}^T \widetilde{C}_t(\xv_{t+1}, \yv_t, \dv_t, \Pv_t) - \sum_{t=1}^T \widetilde{C}_t(\xv_t,\yv_t,\dv_t,\Pv_t) \right|
        %\nonumber\\
        % = & \left| \sum_{t=1}^T M(\yv_t - \xv_{t+1}) - \sum_{t=1}^T M(\yv_t-\xv_t)  \right|\nonumber\\
        % = & \left| \sum_{t=2}^{T+1} M(\yv_{t-1} - \xv_{t}) - \sum_{t=1}^T M(\yv_t-\xv_t)  \right| \nonumber \\
        \leq  & \sum_{t=2}^T M(\yv_t - \yv_{t-1})  + M(\yv_1 - \xv_1)  \label{eq:m_triangle}\\
        \leq & 2\cdot\left(\max_{i,j=1,\dots,n} c_{ij}\right)\cdot\sum\limits_{t=2}^{T}\|\yv_{t}-\yv_{t-1}\|_1
        \label{eq:m_bounded},
\end{align}
where the first two equations follow from the cost definition, \eqref{eq:m_triangle} follows from the triangle inequality of the repositioning functions $M$, and \eqref{eq:m_bounded} follows from the fact that $\DS M(\zv) \leq 2\cdot\left(\max_{i,j=1,\dots,n} c_{ij}\right)\Vert \zv\Vert_1$ and the notation $\yv_0:= \xv_1$.
\halmos

\subsection{Proof of Lemma \ref{lem:gd}}
\label{appendix:pf_lem_gd}
\proof{Proof of Lemma \ref{lem:gd}.}
    First, we show the convexity by definition. That is, for any $\bslevel_1,\bslevel_2\in\mathbb{R}_{+}^{n}$, without loss of generality, it is sufficient to show that
    \begin{align}
        \label{ieq:cvx}
        \alpha\widetilde{C}_{t}(\xv_{t+1}(\bslevel_1),\bslevel_1,\dv_t,\Pv_t)
        +
        (1-\alpha)\widetilde{C}_{t}(\xv_{t+1}(\bslevel_2),\bslevel_2,\dv_t,\Pv_t)
        \geq
        \widetilde{C}_{t}(\xv_{t+1}(\bslevel_3),\bslevel_{{\alpha}},\dv_t,\Pv_t),
    \end{align}
    for all $\alpha\in(0,1)$, where $\bslevel_{\alpha}=\alpha\bslevel_1+(1-\alpha)\bslevel_2$. For simplicity, we assume the optimal solutions of LPs \eqref{opt:one_step} corresponding to $\widetilde{C}_{t}(\xv_{t+1}(\bslevel_k),\bslevel_k,\dv_t,\Pv_t)$ are attainable without loss of generality, and we denote them as $(\boldsymbol{\xi_k}^*= \{\xi^*_{k,ij}\}_{i,j=1}^{n}\}, \boldsymbol{w_k}^*=\{w^*_{k,i}\}_{i=1}^{n})$  for $k=1,2,\alpha$. Then, if $(\alpha\boldsymbol{\xi}_1^*+(1-\alpha)\boldsymbol{\xi}_2^*,\alpha\boldsymbol w_1^*+(1-\alpha)\boldsymbol{w}_2^*)$ is feasible to the LP \eqref{opt:one_step} corresponding to $\widetilde{C}_{t}(\xv_{t+1}(\bslevel_{\alpha}),\bslevel_{{\alpha}},\dv_t,\Pv_t)$, we have \eqref{ieq:cvx} holds since 
    \begin{align*}
        &\alpha\widetilde{C}_{t}(\xv_{t+1}(\bslevel_1),\bslevel_1,\dv_t,\Pv_t)
        +
        (1-\alpha)\widetilde{C}_{t}(\xv_{t+1}(\bslevel_2),\bslevel_2,\dv_t,\Pv_t)\\
        =&
        \sum\limits_{i=1}^{n}\sum\limits_{j=1}^{n} c_{ij}(\alpha\xi_{1,ij}+(1-\alpha)\xi_{2,ij}) -\sum\limits_{i=1}^{n}\sum\limits_{j=1}^{n} l_{ij}P_{t,ij}(\alpha w_{1,i}+(1-\alpha)w_{2,i})\\
        \geq
        &
        \sum\limits_{i=1}^{n} \sum\limits_{j=1}^{n} c_{ij}\xi_{\alpha,ij} -\sum\limits_{i=1}^{n}\sum\limits_{j=1}^{n} l_{ij}P_{t,ij}w_{\alpha,i}
        \\=&
        \widetilde{C}_{t}(\xv_{t+1}(\bslevel_\alpha),\bslevel_\alpha,\dv_t,\Pv_t),       
    \end{align*}
    where the second and last lines come from the definitions for $\boldsymbol{\xi}_k,\boldsymbol{w}_k$ for $k=1,2,\alpha$, and the third line comes from the optimality of $(\boldsymbol{\xi}_{\alpha},\boldsymbol{w}_{\alpha})$. Thus, to finish the proof for convexity, we only need to verify the feasibility of $(\alpha\boldsymbol{\xi}_1^*+(1-\alpha)\boldsymbol{\xi}_2^*,\alpha\boldsymbol w_1^*+(1-\alpha)\boldsymbol{w}_2^*)$. Here, we only verify \eqref{ieq:S_gd}, and other constraints can be checked similarly. To see this, we have 
    \begin{align*}
        \alpha w_{1,i}+(1-\alpha)w_{2,i}
        &\leq
        \alpha\min(\bslevel_1,\dv_t)+(1-\alpha)\min(\bslevel_2,\dv_t)\\
        &\leq
        \min(\alpha\bslevel_1+(1-\alpha)\bslevel_2,\dv_t),
    \end{align*}
    where the first line comes from the definition of $w_{1,i},w_{2,i}$, and the second line comes from the concavity of the min function $\min(\cdot,\dv_t)$.

    Next, we show $\gv_t$ is a subgradient of $\widetilde{C}_t(\xv_{t+1}(\bslevel),\bslevel,\dv_t,\Pv_t)$. The main proof is enlightened by Section 4 of \cite{luenberger1984linear}.

As discussed in the main text, we consider the following LP \eqref{opt:eqilLP_appendix}.
\begin{align}
    \text{LP(}\yv_t) = \min_{\xi_{t,ij},w_{t,i}}  & \sum_{i=1}^n \sum_{j=1}^n  c_{ij} \xi_{t,ij}  
            - \sum\limits_{i=1}^{n}\sum\limits_{j=1}^{n}l_{ij} P_{t,ij}w_{t,i} \label{opt:eqilLP_appendix}\\
            \mathrm{subject\  to}\ &\sum_{i=1}^n \xi_{t,ij} - \sum_{k=1}^n \xi_{t,jk} = w_{t,j}-\sum\limits_{i=1}^{n} P_{t,ij}w_{t,i}, \mathrm{\  for\  all \ } j = 1,\dots, n,\nonumber \\
            & w_{t,i}\geq 0,\ \xi_{t,ij} \geq 0,\  \mathrm{ for\  all \ } i,j = 1,\dots, n,\nonumber \\
            &w_{t,i}\leq y_{t,i}, \ \mathrm{\  for\  all\ } i=1,\dots,n,\label{const:yt_appendix}\\
            & w_{t,i}\leq d_{t,i}, \ \mathrm{\  for\  all\ } i=1,\dots,n. \label{const:dt_appendix}
\end{align}
LP \eqref{opt:eqilLP_appendix} shares the same optimal objective value as LP \eqref{opt:one_step} since constraint \eqref{ieq:S_gd} is equivalent to the combination of \eqref{const:yt_appendix} and \eqref{const:dt_appendix}. Here, in order to differentiate, we additionally denote \eqref{opt:one_step} as OLP (Original LP). We only need to show that $\gv_t$ in Algorithm \ref{alg_nloc:ogr} is the gradient of LP \eqref{opt:eqilLP_appendix} with respect to $\yv_t$ for all $t$. To see this, consider the following dual LP of LP$(\yv_t)$:
\begin{align}
    \label{opt:dual_appendix}
    \text{D-LP}(\yv_t) = \max_{\mu_{t,i},\eta_{t,i},\pi_{t,i}} &   \boldsymbol{\mu}_t^\T \yv_t + \boldsymbol{\eta}_t^\T \dv_t\\
\text{subject to}\  & \pi_{t,j} - \pi_{t,i} \leq c_{ij}, \ \mathrm{\  for\  all\ } i, j = 1,\dots, n,  \nonumber \\
& - \pi_{t,i} + \sum_{j=1}^n P_{t,ij} \pi_{t,j} + \mu_{t,i} + \eta_{t,i} \leq -\sum_{j=1}^n l_{ij}P_{t,ij},   \ \mathrm{\  for\  all\ } i = 1,\dots,n, \nonumber \\
& \mu_{t,i}, \eta_{t,i} \leq 0, \ \mathrm{\  for\  all\ } i=1,\dots,n. \nonumber
\end{align}
where $\boldsymbol{\mu}_t$ and $\boldsymbol{\eta}_t$ are the dual variables, or Lagrangian multipliers, corresponding to constraints \eqref{const:yt_appendix} and \eqref{const:dt_appendix}, respectively. Denote $\boldsymbol{\mu}_t$ and $\boldsymbol{\eta}_t$ as any optimal solutions of D-LP$(\yv_t)$. Then, for any $\yv_t'\in[0,1]^{n}$, 
\begin{equation}
\begin{split}
    \text{D-LP}(\yv_t') - \text{D-LP}(\yv_t)
    &\geq
    \boldsymbol{\mu}_t^{\top}\yv_t'+\boldsymbol{\eta}_t^{\top}\dv_t - \text{D-LP}(\yv_t)\\
    &=
    \boldsymbol{\mu}_t^{\top}\yv_t'+\boldsymbol{\eta}_t^{\top}\dv_t - (\boldsymbol{\mu}_t^{\top}\yv_t+\boldsymbol{\eta}_t^{\top}\dv_t)
    \\
    &=
    \boldsymbol{\mu}_t^{\top}(\yv_t'-\yv_t),
\end{split}
\label{ieq:subg_appendix}
\end{equation}
where the first inequality comes from the feasibility of $\boldsymbol{\mu}_t$ and $\boldsymbol{\eta}_t^{\top}$ to D-LP$(\yv_t')$ and the maximality of the objective value of this dual problem, the second line comes from the strong duality of LP$(\yv_t)$, and the last equality is by direct calculation. 

Furthermore, \eqref{ieq:subg_appendix} implies that any dual optimal solution $\boldsymbol{\mu}_t$ is one subgradient of \eqref{opt:eqilLP_appendix} with respect to $\yv_t$. To show $\gv_t$ is a subgradient, we need to verify that $\gv_t$ is a dual optimal solution to \eqref{opt:eqilLP_appendix}. We note that $ g_{t,i} = \lambda_{t,i} \cdot \mathbbm{1}{\left\{(\dv^c_{t})_i=y_{t,i}\right\}}$, where $\lambda_{t,i}$ is optimal solution to 
\begin{align}
    \label{opt:dual_eqilLP}
    \text{D-OLP}(\yv_t) = \max_{\lambda_{t,i},\pi_{t,i}} &   \boldsymbol{\lambda}_t^\T\dv^c_t\\
\text{subject to}\  & \pi_{t,j} - \pi_{t,i} \leq c_{ij}, \ \mathrm{\  for\  all\ } i, j = 1,\dots, n,  \nonumber \\
& - \pi_{t,i} + \sum_{j=1}^n P_{t,ij} \pi_{t,j} + \lambda_{t,i} \leq -\sum_{j=1}^n l_{ij}P_{t,ij},   \ \mathrm{\  for\  all\ } i = 1,\dots,n, \nonumber \\
& \lambda_{t,i} \leq 0, \ \mathrm{\  for\  all\ } i=1,\dots,n. \nonumber
\end{align}
We now define $h_{t,i} = \lambda_{t,i} \cdot \mathbbm{1}{\left\{(\dv^c_{t})_i\neq y_{t,i}\right\}}$. Therefore, $\lambda_{t,i} = g_{t,i}+h_{t,i}$, and 
\[
\begin{split}
    \boldsymbol{\lambda}_t^\T\dv^c_t = \sum_{i:(\dv^c_{t})_i= y_{t,i}} g_{t,i} y_{t,i} + \sum_{i:(\dv^c_{t})_i\neq y_{t,i}} h_{t,i} d_{t,i} = \gv_t^\T\yv_t + \hv_t^\T \dv_t.
\end{split}
\]
Finally, since \eqref{opt:dual_eqilLP} and \eqref{opt:eqilLP_appendix} share the same optimal objective function value, we have that $\gv_t$ is a dual optimal solution to \eqref{opt:eqilLP_appendix}.

\halmos

\subsection{Proof of Theorem \ref{thm:gdr}}

\begin{lemma} 
    \label{lem:ogd}
    For any sequence of functions $\{f_1,f_2,\dots\}$ defined on a convex set $\mathcal{K}$ and any initialization $\xv_1\in\mathcal{K}$, recursively define
    \begin{align*}
        \xv_t = \Pi_{\mathcal{K}}\left(\xv_{t-1}-\frac{\eta}{\sqrt{t}}\nabla f_{t-1}(\xv_{t-1})\right),
    \end{align*}
    where $\Pi_{\mathcal{K}}(\cdot)$ is the projection function on $\mathcal{K}$. This algorithm is known as the projected online gradient descent algorithm.
    Suppose $f_t$'s are convex and $\mathcal{K}$ is closed, bounded and convex. Let $D$ be an upper bound for the diameter of $\mathcal{K}$, which satisfies
    \begin{align*}
        \|\xv-\yv\|_2\leq D, \text{ for all } \xv,\yv\in\mathcal{K},
    \end{align*}
    and $G$ be an upper bound on the norm of the subgradients of $f_t$'s, i.e., $\|\nabla f_t(\xv)\|_2\leq G$ for all $\xv\in\mathcal{K}$ and $t\geq 1$. Then, with $\eta = D/G$, the online gradient descent guarantees the following for all $T\geq1 $:
    \begin{align*}
        \sum\limits_{t=1}^{T}f_t(\xv_t) - \min_{\xv^*\in\mathcal{K}}f_{t}(\xv^*)\leq 3DG\sqrt{T}.
    \end{align*}
\end{lemma}
\proof{Proof of Lemma \ref{lem:ogd}.}
    The Projected Online Gradient Descent algorithm is a well-established online convex optimization algorithm. This is a standard theoretical performance guarantee for the online gradient descent algorithm, we refer to Theorem 3.1 in \cite{hazan2022introduction} for the proof.
\halmos

\proof{Proof of Theorem \ref{thm:gdr}.}

By the Lipschitz property in Lemma \ref{lem:h_lipschitz}, we know that the subgradient norms can be bounded by 
\[
\Vert \gv\Vert_2 \leq n^2(\max_{i,j} c_{ij}+\max_{i,j} l_{ij}) .
\]
On the other hand, for any two points $\xv,\yv \in \Delta_{n-1}$, $\Vert \xv - \yv\Vert_2 \leq \Vert \xv \Vert_2 + \Vert \yv\Vert_2 \leq 2$. By Lemma \ref{lem:gd}, we have the convexity of $\widetilde{C}_{t}(\xv_t(\bslevel),\bslevel,\dv_t,\Pv_t)$, and thus we invoke the convergence rate of online gradient descent in Lemma \ref{lem:ogd} to obtain that
    \begin{align}
        \label{ieq:ogd}
        \sum\limits_{t=1}^{T}\widetilde{C}_{t}(\xv_{t+1}(\bslevel_{t}),\bslevel_{t},\dv_t,\Pv_t)
        \leq \min_{\bslevel\in\Delta_{n-1}} \sum\limits_{t=1}^{T}\widetilde{C}_{t}(\xv_t(\bslevel),\bslevel,\dv_t,\Pv_t)+ 6n^2\left(\max_{i,j}c_{ij}+\max_{i,j}l_{ij}\right)\cdot \sqrt{T}     
    \end{align}
    holds for all $\dv_t\in[0,1]^{n}$ and transition probability matrix $\Pv_t$ for all $t=1,\dots T$. 

 In addition, by the approximation error in Lemma \ref{lem:disentangle}, one can show 
    \begin{align}
        \label{ieq:lipschitz_x}
        \left|\widetilde{C}_{t}(\xv_t(\bslevel_{t-1}),\bslevel_{t},\dv_t,\Pv_t)
        -
        \widetilde{C}_{t}(\xv_{t+1}(\bslevel_{t}),\bslevel_{t},\dv_t,\Pv_t)
        \right|
        \leq
        \left(\max_{ij} c_{ij}\right)\|\bslevel_{t-1}-\bslevel_{t}\|_1
    \end{align}
    for all $t=1,\dots,T$.
    We first notice that $\bslevel_{t+1} = \Pi_{\Delta_{n-1}}(\bslevel_{t} - \frac{1}{\sqrt{t}}\gv_{t})$, and thus by triangle inequality, we have
    \[
    \begin{split}
    \Vert \bslevel_{t+1} - \bslevel_{t}\Vert_1 & \leq \Vert \bslevel_{t+1} - \bslevel_{t} + \frac{1}{\sqrt{t}}\gv_{t} \Vert_1 + \Vert \frac{1}{\sqrt{t}}\gv_{t} \Vert_1\\
    & \leq \sqrt{n} \Vert  \bslevel_{t+1} - \bslevel_{t} + \frac{1}{\sqrt{t}}\gv_{t} \Vert_2 + \sqrt{n} \Vert \frac{1}{\sqrt{t}}\gv_{t} \Vert_2 \\
    & \leq \sqrt{n} \Vert  \bslevel_{t} - \bslevel_{t} + \frac{1}{\sqrt{t}}\gv_{t} \Vert_2 + \sqrt{n} \Vert \frac{1}{\sqrt{t}}\gv_{t} \Vert_2\\
    & = 2 \frac{\sqrt{n}}{\sqrt{t}} \Vert \gv_t\Vert_2,
    \end{split}
    \]
    where the first line is by the triangle inequality, the second line follows from the fact that $\Vert \zv\Vert_1 \leq \sqrt{n} \Vert \zv\Vert_2$ for any $n$-dimensional vector $\zv$, the third line follows from the projection definition and the minimality of distance, and the last line is by direct calculation. Since $\sum_{t=1}^T \frac{1}{\sqrt{t}} \leq \sum_{t=1}^T \frac{2}{\sqrt{t} + \sqrt{t-1}} =2 \sum_{t=1}^T (\sqrt{t}- \sqrt{t-1}) = 2\sqrt{T}$, it follows that 
    \begin{equation}
    \label{eq:bound_sum_S_diff}
        \sum\limits_{t=1}^{T}\|\bslevel_{t-1}-\bslevel_{t}\|_1 \leq 4n^{5/2}\sqrt{T} \left(\max_{i,j}c_{ij}+\max_{i,j}l_{ij}\right).
    \end{equation}
     Next, combining \eqref{ieq:ogd}, \eqref{ieq:lipschitz_x} and \eqref{eq:bound_sum_S_diff}, we can show \eqref{ieq:regret_noniid} by
    \begin{align*}
        \sum\limits_{t=1}^{T}\widetilde{C}_{t}(\xv_t(\bslevel_{t-1}),\bslevel_{t},\dv_t,\Pv_t)
        &\leq
        \sum\limits_{t=1}^{T}\widetilde{C}_{t}(\xv_{t+1}(\bslevel_{t}),\bslevel_{t},\dv_t,\Pv_t)+\left(\max_{ij}c_{ij}\right)\sum\limits_{t=1}^{T}\|\bslevel_{t-1}-\bslevel_{t}\|_1\\
        &\leq
        \sum\limits_{t=1}^{T}\widetilde{C}_{t}(\xv_{t+1}(\bslevel_{t}),\bslevel_{t},\dv_t,\Pv_t)+ 4n^{5/2}\sqrt{T} \left(\max_{i,j}c_{ij}+\max_{i,j}l_{ij}\right)^2\\
        &\leq
        \min_{\bslevel\in\Delta_{n-1}} \sum\limits_{t=1}^{T}\widetilde{C}_{t}(\xv_t(\bslevel),\bslevel,\dv_t,\Pv_t)+(6n^2+4n^{5/2})\sqrt{T} \left(\max_{i,j}c_{ij}+\max_{i,j}l_{ij}\right)^2,
    \end{align*}
   where the first line is obtained by \eqref{ieq:lipschitz_x}, the second line follows from \eqref{eq:bound_sum_S_diff}, and the last line is obtained by \eqref{ieq:ogd}. Next, we prove that if the demand and transition probability pairs $\left\{(\dv_t,\Pv_t)\right\}_{t=1}^{T}$ are i.i.d.,  \eqref{ieq:regret_iid}  holds. To see this, we have
    \begin{align*}
        \mathbb{E}\left[ \min_{\bslevel\in\Delta_{n-1}} \sum\limits_{t=1}^{T}\widetilde{C}_{t}(\xv_t(\bslevel),\bslevel,\dv_t,\Pv_t)\right]
        \leq
        \min_{\bslevel\in\Delta_{n-1}} T\mathbb{E}\left[ \widetilde{C}_{1}(\xv_1(\bslevel),\bslevel,\dv_1,\Pv_1)\right]
    \end{align*}
     by Jensen's inequality, and then \eqref{ieq:regret_iid} is obtained by taking expectation in both sides of \eqref{ieq:regret_noniid}.
\halmos

%% file: files/appendix_omitted_proofs.tex
%%%%%%%%%%%%%%%%%%%%%%%%%%%%%%%%%%%%%%%%%%%%%%%%%%%%%%%%%%%%

\section{Supplements for Section~\ref{section:lb_and_dimension} and Section~\ref{section:further_and_extension}}

\subsection{Proof of Proposition \ref{lem:infeas}}
\label{subsec:proof_lem_infeas}
\proof{Proof of Proposition \ref{lem:infeas}.}    
We define a set of probability distributions $\mathcal{P}_c$ for $c\in (0.5,1)$ as follows, 
\begin{align*}
   \mathcal{P}_c = \{ & (X,Y) \mid \mathbb{P}(X=1,Y=1) = \mathbb{P}(X=c,Y=c) = p,  \\ 
   &\ \mathbb{P}(X=1,Y=c) = \mathbb{P}(X=c,Y=1) = 0.5 - p\text{, for some } p \in (0,0.5)\}
\end{align*}
From the construct, we can see that $\mathcal{P}_c$ is a set of distributions indexed by the probability $p \in (0,0.5)$. Then, for any $x_0,y_0\geq0$ satisfying $x_0+y_0=1$, we can calculate the probability density distribution of $\left(\min(X,x_0),\min(Y,y_0)\right)$ as follows,  
\begin{align*}
    &\left(\min(X,x_0),\min(Y,y_0)\right)
    \\
    =&
    \begin{cases}
        (x_0,y_0) &\text{ with probability 1 if $x_0,y_0\leq c$},\\
        (c,y_0) \text{ or } (x_0,y_0)  &\text{ with probability 0.5 and 0.5, respectively,  if $c< x_0\leq 1$},\\
        (x_0,c) \text{ or } (x_0,y_0)  &\text{ with probability 0.5 and 0.5, respectively,  if $c< y_0<1$}.
    \end{cases}
\end{align*}
Therefore we have shown that  $(X,Y)$ in $\mathcal{P}_c$ have different distributions, but their censored versions share the same distribution.
\halmos

\subsection{Proof of Theorem~\ref{theorem:sqrt_t_lb}}
\label{subsection:proof_lb}

\proof{Proof of Theorem~\ref{theorem:sqrt_t_lb}.}
To see this, we consider an extreme case where the repositioning costs are $0$, and in this case, the best base-stock policy is optimal based on Theorem~\ref{thm:asymptotic_optimal}. We note that in this special setting, Assumption~\ref{assump:lsc>rc} automatically holds. Additionally, we assume that demand is large, i.e., $D_i=1$ for all $i\in \mathcal{N}$. Suppose a repositioning level $\bslevel = (S_1,S_2,\dots,S_n)$ is applied at time $t$, then the expected cost at time $t$ is
\[
\sum_{i=1}^n \sum_{j=1}^n l_{ij} P_{t,ij} \Expt[D_{t,i} - S_{i}] = \sum_{i=1}^n \left(\sum_{j=1}^n l_{ij} P_{t,ij}\right) \Expt[D_{t,i}] - \sum_{i=1}^n \left(\sum_{j=1}^n l_{ij} P_{t,ij}\right) S_i.
\]

We denote $\mu_i = \left(\sum_{j=1}^n l_{ij} P_{t,ij}\right) - C$ for $i = 1,\dots,n$ and some $C>0$. Then the lost sales cost can be rewritten as 
\[
\sum_{i=1}^n \left(\sum_{j=1}^n l_{ij} P_{t,ij}\right) \Expt[D_{t,i}]  -C - \sum_{i=1}^n \mu_i S_i,
\]
where the first two terms are independent of the policy/arm at time $t$, and the third term can be exactly understood as a stochastic linear optimization. Based on \citet{dani2008stochastic}, there exists an instance such that the regret lower bound is at least $\bigo(n\sqrt{T})$ and thus we conclude our proof.
\halmos

\subsection{Proof of Theorem \ref{thm:regret_1/2}}
\begin{algorithm}[!htb]
\caption{\pcr{DL-Uncensored:} Dynamic Learning Algorithm with Uncensored Demand Data}
\label{alg_nloc:dl_uncensored_doubling}
\begin{algorithmic}[1]
\State \textbf{Input:} Number of iterations $T$, initial repositioning policy $\bslevel_1$, initial epoch number $e=1$;
\While{$t < T$}
    \For{$t = 2^{e-1}, \dots, \min\{2^{e} - 1, T\}$}
        \State Apply base-stock repositioning policy $\widetilde{\bslevel}_e$ at period $t$ and record $\bslevel_t = \widetilde{\bslevel}_e$;
        \State Collect \emph{uncensored} data $(\dv_t, \Pv_t)$ from period $t$;
    \EndFor
    \State Solve offline problem \eqref{opt:offline} with  data $\left\{(\dv_s,\Pv_s) \right\}_{s=1}^{2^{e}-1}$ and denote the solution by $\widetilde{\bslevel}_{e+1}$;
    \State Update $e \gets e+1$;
\EndWhile
\State \textbf{Output:} $\left\{\bslevel_t\right\}_{t=1}^{T}$.
\end{algorithmic}
\end{algorithm}
\proof{Proof of Theorem \ref{thm:regret_1/2}.}
By Proposition \ref{prop:offline_concentration}, the total regret at time period $t = 2^{e-1},\dots, 2^{e}-1$ is bounded by 
\[
\begin{split}
&\sum_{t=2^{e-1}}^{ 2^{e}-1}  15n^3\left(\max_{i,j} c_{i,j} + \max_{i,j} l_{i,j}\right)\cdot \sqrt{\frac{\log T}{t}} \\
\leq & 15 n^3\left(\max_{i,j} c_{i,j} + \max_{i,j} l_{i,j}\right)\cdot \sqrt{\log T} 2^{e-1} \frac{1}{\sqrt{2^e}} = 15 n^3\left(\max_{i,j} c_{i,j} + \max_{i,j} l_{i,j}\right)\sqrt{\log T}\cdot  2^{e/2-1}.
\end{split}
\]
Summing up, we know that the total regret is bounded by 
\[
\sum_{e = 1}^{\lceil \log_2T\rceil} 15 n^3\left(\max_{i,j} c_{i,j} + \max_{i,j} l_{i,j}\right)\sqrt{\log T}\cdot  2^{e/2-1} = O(n^3\sqrt{T \log T}). 
\]
We note that at the beginning of each epoch, one might need to rematch the initial inventory levels, but since there are at most $\lceil \log_2T\rceil$ epochs, the incurred regret $O(\log T)$ has been dominated.
\halmos

\subsection{Proof of Theorem \ref{thm:regret_2/3}}
\begin{algorithm}[!htb]
\caption{\pcr{OTL:} One-Time Learning Algorithm }
\label{alg_nloc:otl}
\begin{algorithmic}[1]
\State \textbf{Input:} Number of iterations $T$, initial repositioning policy $\bslevel_1$;
\For{$s =1,..., T_0$, $i = 1,\dots, n$ } \label{alg:data_collect_st}
\State At time $t = n(s-1)+i$: Reposition all inventory to location $i$;  Collect demand $d_{n(s-1)+i, i}$ and transition probability element 
$P_{n(s-1)+i, ij}$ for all $j$;
\EndFor  \label{alg:data_collect_end}
\State For $s= 1,\dots, T_0$, construct $\hat{\dv}_s = (d_{n(s-1)+1, 1}, \dots, d_{ns+n, n})$ and also construct $\hat{\Pv}_s$ by $(\hat{\Pv}_s)_{ij} = P_{n(s-1)+i, ij}$ for $i,j\in \mathcal{N}$;
\State Solve offline problem \eqref{opt:offline} with $T_0$ constructed data pairs $\left\{(\hat{\dv}_s, \hat{\Pv}_s)\right\}_{s=1}^{T^{2/3}}$ to obtain $\hat{\bslevel}$;
\For{time $t = n T_0, n T_0+1, \dots, T$}
\State Apply base-stock repositioning policy $\bslevel_t = \hat{\bslevel}$;
\EndFor
\State \textbf{Output:} $\left\{\bslevel_t\right\}_{t=1}^{T}$.
\end{algorithmic}
\end{algorithm}
\proof{Proof of Theorem \ref{thm:regret_2/3}.}
We can prove this theorem straightforwardly by applying the generalization bound in Proposition \ref{prop:offline_concentration}. Specifically, by collecting $nT_0$ uncensored samples for different locations, we construct $t=T_0$ uncensored joint demand data based on Assumption \ref{assmp:indepedence}, and then draw a policy $\hat{\bslevel}$ through solving the offline problem. Let $T_0 = \eta T^{2/3}$, then by Proposition \ref{prop:offline_concentration}, we have
\begin{align}
    \label{ieq:one_step}
    \mathbb{E}[\widetilde{C}_1^{\hat{\bslevel}}]
    &\leq
    \frac{1}{t} \sum\limits_{s=1}^{t}\widetilde{C}_s^{\hat{\bslevel}}+10n^3\left(\max_{i,j} c_{i,j} + \max_{i,j} l_{i,j}\right)\cdot \sqrt{\frac{\log T}{t}}\nonumber\\
    &\leq
    \frac{1}{t} \sum\limits_{s=1}^{t}\widetilde{C}_s^{\bslevel^*}+10n^3\left(\max_{i,j} c_{i,j} + \max_{i,j} l_{i,j}\right)\cdot \sqrt{\frac{\log T}{t}}\\
    &\leq
    \mathbb{E}[\widetilde{C}_1^{{\bslevel}^*}] +15n^3\left(\max_{i,j} c_{i,j} + \max_{i,j} l_{i,j}\right)\cdot \sqrt{\frac{\log T}{t}}\nonumber\\
    &=
    \mathbb{E}[\widetilde{C}_1^{{\bslevel}^*}] +15n^{3}\left(\max_{i,j} c_{i,j} + \max_{i,j} l_{i,j}\right)\cdot \frac{\sqrt{\log T}}{\eta^{1/2} T^{1/3}}\nonumber,
\end{align}
where the first and third lines come from Proposition \ref{prop:offline_concentration}, the second line comes from the optimality of $\hat{\bslevel}$ in the empirical offline problem, and the last line comes from plugging in the value of $t$. Thus, the total regret can be obtained by
\begin{align*}
 \text{Regret} &\leq (2+n)\cdot \left(\max_{i,j} c_{i,j} + \max_{i,j} l_{i,j}\right)\cdot n \eta T^{2/3}+ (T-t)\cdot15n^{3}\left(\max_{i,j} c_{i,j} + \max_{i,j} l_{i,j}\right)\cdot \frac{\sqrt{\log T}}{\eta^{1/2} T^{1/3}}\\
    &=
    O\left((\eta+n\eta^{-1/2}) n^{2} T^{2/3}\sqrt{\log T}\right),
\end{align*}
where the first part comes from the exploration in collecting samples which can be bounded using Lemma~\ref{lem:h_lipschitz}, and the second part is the accumulative regret in the remaining $T-n\eta T^{2/3}$ periods. Combined the two regrets together, we obtain the desired regret bound.
\halmos

%% file: files/appendix_numerical.tex
% change the equation numbering in the appendix
\renewcommand{\thesection}{\Alph{section}}
\numberwithin{equation}{section}
%--------------

\section{Details of Numerical Experiments}
%\label{appendix:numerical_Details}

We provide a comprehensive description of our numerical experiments setup supplementing Section~\ref{section:numerical}.

%------------Will move to appendix-------------------%

%\subsection{Details of Numerical Experiments}
\label{appendix:synthetic_exp}

We then explain in detail how the synthetic data used in our numerical experiments is generated. 

For each sample of transition probability matrix $P$, we first generate a matrix $Q$ as follows: the elements in the first and second column of $Q$ are generated randomly from an exponential distribution with mean $10$, and all the elements in the other columns are generated randomly from $\mathrm{Unif}(0,1)$. We then adjust all diagonal elements into $10$ times their original value respectively. Our synthetic idea is calibrated based on real-world scenarios: there is heterogeneity in terms of locations and in this synthetic data we choose locations $1$ and $2$ as popular destination locations; additionally, most trips are more likely to end at the same location as the origin, and therefore we increase the values of all the diagonal elements. Lastly, we then normalize the sum of each row of $Q$ into $1$ so that we obtain $P$ as a probability matrix.

We consider the following demand scenarios.
\begin{enumerate}[label=(\roman*)]
    \item Network independence: we generate the demand for different locations independently, and for location $i$, demand $d_i$ is generated from uniform distribution $\mathrm{Unif}\left(0.3\times{i}/{n}, 0.6\times{(i+1)}/{n}\right)$.
    \item Network dependence: we first sample vector $\mathbf{v}$ from a multivariate normal distribution with mean $2/n \times \mathbf{1}_n$ and covariance matrix $10 \times A^\T A$, where $\mathbf{1}_n$ denotes an $n$-dimensional all-one vector and $A$ is a random matrix with each element sampled from $\mathrm{Unif}(0,1)$. For $i\in\mathcal{N}$, then obtain the demand $d_i$ by truncating $v_i$ it into the interval $[l(i), u(i)]$, where $l(i) = 0.2 + 0.2i/n$ and $u(i) = 0.4 + 0.8i/n$. 
\end{enumerate}

We consider the following cost scenarios.
\begin{enumerate}[label=(\roman*)]
    \item High lost sales cost: For $i,j\in\mathcal{N}$, the unit lost sales cost is randomly generated from $\mathrm{Unif}(1,2)$ and the unit repositioning cost is randomly generated from $\mathrm{Unif}(0.5,1)$. We call this scenario high lost sales cost since it is sufficient to make the Assumption~\ref{assump:lsc>rc} hold. We comment that the difference between the two costs here is not strong and they are still at a very similar scale. This is the default cost setting for most of our experiments. 
    \item High repositioning cost (Table~\ref{table:milp_comparison}): For $i,j\in\mathcal{N}$, the unit lost sales cost is randomly generated from $\mathrm{Unif}(1,2)$ and the unit repositioning cost is randomly generated from $\mathrm{Unif}(5,10)$. With repositioning cost increased $10$ times, Assumption~\ref{assump:lsc>rc} fails to hold, and we aim to test the performance of our MILP formulation. We test under 125 time periods and still adopt an exploration period of length $60$, and we consider network independence setup.
\end{enumerate}

For the one-time learning algorithm, the length of exploration is set as $20n$. 
For each setting, we repeat the experiments for $20$ times and report both the average performances, and the total number of periods is set as $500$ if not specified otherwise. The 95\% confidence intervals for both regret and relative regret are computed in the linear scale as mean $\pm 1.96 \times \mathrm{SE}$ where $\mathrm{SE}$
is the standard error across $K=20$ experiments. The regret is subsequently displayed on a log scale, while relative regret is shown as a percentage on a linear scale.

%% file: files/appendix_extension.tex
\section{Analysis of Extended Model}
% \subsection{\pcr{SOAR-Extended} Algorithm}
\subsection{Theoretical Results and Proofs}
\label{appendix:ogr_extended_theory}

\begin{assumption}[Cost Condition in Multi-subperiod Setting]
\label{assump:lsc>rc_extended}
For any period $t$ and subperiod $h$, 
 \begin{equation}
     \sum_{i=1}^{n} l_{ji} P_{th,ji} \geq \sum\limits_{i=1}^{n}P_{th,ji}c_{ij}, \text{ for all $j=1,\dots,n$}.
\end{equation}
\end{assumption}
Assumption~\ref{assump:lsc>rc_extended} generalizes Assumption~\ref{assump:lsc>rc}, with the latter being a special case where $H=1$. While Assumption~\ref{assump:lsc>rc_extended} imposes a stronger condition by requiring the inequality to hold for each subperiod rather than only in aggregate, its practical validity is supported by real-world scenarios, particularly when lost sales costs are linked to market growth. 

%For empirical validation, we direct readers to the detailed cost calibration using real data in \citet[Appendix I.3]{akturk2022managing}.

\begin{proposition}
    \label{prop:offline_LP_extended}
    Under Assumption \ref{assump:lsc>rc_extended} and oracle of uncensored demands, the best base-stock repositioning policy of the $H$-subperiod extended model can be computed by the following linear programming problem.
    \begin{equation}
     \label{opt:offline_LP_extended}
    \begin{aligned}
        \min_{S_i,\xi_{s,ij}, w_{sk,i}, x_{sk,i}, \gamma_{sk,i}}\  & \sum\limits_{s=1}^{t}\sum_{i=1}^n \sum_{j=1}^nc_{ij} \xi_{s,ij}  
        - \sum_{k=1}^H\sum\limits_{s=1}^{t} \sum\limits_{i=1}^{n}\sum\limits_{j=1}^{n}l_{ij} P_{sk,ij}w_{sk,i}
        \\
        \mathrm{subject\  to}\ &\sum_{i=1}^n \xi_{s,ij} - \sum_{k=1}^n \xi_{s,jk} = \sum_{k=1}^H w_{sk,j}-\sum_{k=1}^H\sum\limits_{i=1}^{n} P_{sk,ij}w_{sk,i}, \forall j = 1,\dots, n , s=1,\dots,t,   \\
          & \xv_{s(k+1)} = \xv_{sk} - \wv_{sk} + \Pv_{sk}^\T \left[\wv_{sk} + \gammav_{sk}  \right],  \forall s = 1,\dots, t,  k = 1,\dots, H, \\
    &\gammav_{s(k+1)}  = \left[ \wv_{sk} + \gammav_{sk} \right] \circ \left[ (\Iv-\Pv_{sk}) \ev\right], \forall s= 1,\dots, t, k = 1,\dots, H,  \\
    & \gamma_{t1, i} = 0, x_{t1,i} = S_i, \forall i = 1,\dots, n, \\
        & \xi_{s,ij} \geq 0,\  \forall i = 1,\dots, n, \forall i,j = 1,\dots, n\mathrm{\  and\ } s=1,\dots,t, \\
        &\sum\limits_{i=1}^{n} S_i=1,\ \bslevel=\{S_i\}_{i=1}^{n}\in[0,1]^{n} ,\\    
        &w_{sk,i}\leq d_{sk,i},\ w_{sk,i}\leq x_{sk, i}, \ w_{sk,i}\geq 0,\forall s=1,\dots,t,i=1,\dots,n, k = 1,\dots, H. 
    \end{aligned}
    \end{equation}
\end{proposition}

\begin{lemma}
    \label{lem:disentangle_extended}
    Let $\{\yv_t\}_{t=1}^{T}\subseteq\Delta_{n-1}$ be any sequence of repositioning policies. Then, the relabeled modified cost $\widetilde{C}(\xv_{t+1}(\yv_t),\yv_t,\{(\dv_{tk}, \Pv_{tk})\}_{k=1}^H)$ depends only on the repositioning policy and realized demands and transition matrices at time $t$, for all $t=1,\dots,T$. Here, $\xv_{t+1}$ follows the dynamics described in \eqref{eq:inv_gen_x} and \eqref{eq:inv_gen_gamma} for all $t=1,\dots,T$.
    
    Furthermore, the gap between the cumulative modified cost and the cumulative relabeled modified cost can be bounded by the following inequality where $\yv_0:=\xv_1$,
    \begin{equation}
        \left|\sum_{t=1}^T \widetilde{C}(\xv_t, \yv_t, \{(\dv_{tk}, \Pv_{tk})\}_{k=1}^H) - \sum_{t=1}^T \widetilde{C}(\xv_{t+1}, \yv_t, \{(\dv_{tk}, \Pv_{tk})\}_{k=1}^H)\right| \leq 2\cdot\left(\max_{i,j=1,\dots,n} c_{ij}\right)\cdot\sum\limits_{t=2}^{T}\|\yv_{t}-\yv_{t-1}\|_1.
    \end{equation}
\end{lemma}

We abbreviate the proofs of  Proposition~\ref{prop:offline_LP_extended} and Lemma~\ref{lem:disentangle_extended} due to space limits.

\begin{algorithm}[!htb]
\caption{\pcr{SOAR-Extended}: Surrogate Optimization and Adaptive Repositioning Algorithm for Extended Model}
\label{alg_nloc:ogr_extended}
\begin{algorithmic}[1]
\State \textbf{Input:} Number of iterations $T$, number of subperiods $H$, initial repositioning policy $\yv_1$;
\For{$t =1,...,$}
    %\Statex \hspace{1em} \emph{\color{BlueViolet} \% Observe censored data}
    \State Set the target inventory as $\xv_{t1} = \yv_t$ and observe realized censored demand $\dv_{tk}^{c}=\min(\xv_{tk},\dv_{tk})$ for $k \in [H], t\in[T]$; \label{alg_extended_step:set}
    %\Statex \hspace{1em} \emph{\color{BlueViolet} \% Solve linear programming involving surrogate costs}
    \State Denote $\boldsymbol{\lambda}_{tk}$ be the optimal dual solution corresponding to (Constraint-$k$); %aeq means appendix eq
    \begin{align}
      %\widetilde{C}(\xv_{t+1}, \yv_t, \{(\dv_{tk}, \Pv_{tk})\}_{k=1}^H) = 
      \min &\sum_{i=1}^n \sum_{j=1}^n c_{ij} \xi_{t, ij} - \sum_{k=1}^H \sum_{i=1}^n \sum_{j=1}^n l_{ij} P_{th, ij} w_{th,i}, \nonumber\\
   \text{ subject to} & \sum_{i=1}^n \xi_{t, ij} - \sum_{i' = 1}^n \xi_{t, ji'} = \sum_{k=1}^H \left[ w_{tk,j} -\sum_{i=1}^n P_{tk, ij} ( w_{tk,i} + \gamma_{tk,i})  \right], \forall j \in [n], \nonumber\\
   & \gamma_{t(k+1),i} = (w_{tk,i} + \gamma_{tk,i})\left(1- \sum_{j=1}^n P_{tk,j}\right), \forall k \in [H], i \in [n], \label{eq:gamma_update_alg} \\
    & \gamma_{t1,i} = 0, \forall i \in [n], \nonumber \\
   &  w_{tk, i} \geq 0,\, \xi_{t, ij} \geq 0, \forall i,j\in [n],\nonumber\\\
   & w_{t1, i} \leq d_{t1,i}^c, \forall i\in [n], \label{aeq:constraint_1} \tag{Constraint-$1$} \\
   & w_{t2, i} \leq d_{t2,i}^c, \forall i\in [n],\tag{Constraint-$2$} \\
   & \dots \nonumber\\
   & w_{tH, i} \leq d_{tH,i}^c, \forall i\in [n]. \tag{Constraint-$H$} \label{aeq:constraint_H}
\end{align}
    %\Statex \hspace{1em} \emph{\color{BlueViolet} \% Construct subgradient from dual solution}
    \State Let
    $\gv_{tk} =\lambdav_{tk} \circ \mathbbm{1}\left\{\dv^c_{tk}=\xv_{th}\right\}$ where $\lambda_{tk,i}$, $i\in[n]$ is the dual solution corresponding to (Constraint-$k$) for $k = 1,\dots, H$;
    \State Compute $\muv_{tk}$, $k = H, H-1,\dots, 1$ recursively through \eqref{aeq:recursion}; \label{alg_extended_step:compute_gradient}
     %\Statex \hspace{1em} \emph{\color{BlueViolet} \% Projected gradient descent update with step size $1/\sqrt{t}$}
    \State Update the repositioning policy $\yv_{t+1}=\Pi_{\Delta_{n-1}}\left(\yv_{t}-\frac{1}{H\sqrt{t}}\sum_{k=1}^H\muv_{tk}\right)$; \label{alg_extended_step:update_policy}
\EndFor
\State \textbf{Output:} $\left\{\yv_t\right\}_{t=1}^{T}$.
\end{algorithmic}
\end{algorithm}

\proof{Proof of Theorem~\ref{thm:gdr_extended}.}
% We take multiple steps to prove the regret result for the extended model.

% \noindent\emph{Validity of Subgradient.}
Similar to  Lemma~\ref{lem:gd}, we need to first show the convexity of the surrogate costs with respect to $\yv_t$ and prove the validity of the gradient to the surrogate costs. First, the convexity property follows from the linearity structure, and the fact that the $\dv_{tk}^c$ defined through a concave $\min$ function. Consider the following LP \eqref{eq:ogd_LP_extended} and denote its optimal value as a function of $\yv_t = \xv_{t1}$ as $\text{LP}(\yv_t) = \widetilde{C}(\xv_{t+1}(\yv_t), \yv_t, \{(\dv_{tk}, \Pv_{tk})\}_{k=1}^H)$. Compared to the original LP subproblem \eqref{eq:ogd_olp_extended} defined in Algorithm~\ref{alg_nloc:ogr_extended}, the inventory dynamics across subperiods are included, the constraints $w_{tk,i} \leq (\dv^c_{tk})_i$ (noting inequality instead of equality here thanks to Assumption~\ref{assump:lsc>rc_extended}) are separated into  $ w_{tk,i} \leq d_{tk,i}$ and $ w_{tk,i} \le x_{tk,i}$ for $i = 1,\dots, n$. To identify the role of $\yv_{t}$, we invoke \eqref{eq:inv_gen_x} to express $x_{tk,i}$ using $y_{t,i}$ and decision variables to rewrite $ w_{tk,i} \le x_{tk,i}$ into 
\begin{equation}
    w_{tk,i} \leq y_{t,i} - \sum_{h=1}^{k-1} w_{th,i} + \sum_{h=1}^{k-1}\sum_{j=1}^n P_{th, ji}(w_{th,j} + \gamma_{th,j}) \label{eq:wx_first_form}.
\end{equation}
To further removing $\gamma_{th,j}$ from \eqref{eq:wx_first_form}, we invoke \eqref{eq:inv_gen_gamma} to obtain
\[
\gammav_{th} = \sum_{j=1}^{h-1} \left(\wv_{tj} \circ \prod_{l=j}^{h-1} \left[ (\Iv - \Pv_{tl})\ev\right] \right),
\]
and plug it into \eqref{eq:wx_first_form} to obtain \eqref{eq:wx_second_form}. The converted form in \eqref{eq:wx_second_form} is essential as we construct subgradient with respect to $\yv_t$.
\begin{align}
     %\text{LP}(\xv_{t1}) = 
     \min_{\xi_{t,ij}, w_{tk,i}, \gamma_{tk,i}} \quad 
    & \sum_{i=1}^n \sum_{j=1}^n c_{ij}\,\xi_{t,ij}
      -
      \sum_{k=1}^H \sum_{i=1}^n \sum_{j=1}^n l_{ij}\,P_{tk,ij}\,w_{tk,i},
      \label{eq:ogd_LP_extended}\\
    \text{subject to}\quad
    & \sum_{i=1}^n \xi_{t,ij} - \sum_{k=1}^n \xi_{t,jk} 
      = 
      \sum_{k=1}^H \Bigl[w_{tk,j} - \sum_{i=1}^n P_{tk,ij}\,(w_{tk,i} + \gamma_{tk,i})\Bigr],
      \quad \forall\, j,
      \nonumber\\
    % & x_{t(k+1), i} = x_{tk, i} - w_{tk,i} + \sum_{j=1}^n P_{tk, ji}(w_{tk,j} + \gamma_{tk,j}), \forall k,i, \\
    % &x_{t1,i} = y_{t,i}, \forall i, \\
    & \gamma_{t1,i} = 0, \forall i,\nonumber\\
    & \gamma_{t(k+1),i} =
      (w_{tk,i} + \gamma_{tk,i})
      \left(1 - \sum_{j=1}^n P_{tk,ij}\right), 
      \quad \forall k ,i, \nonumber
      \\
    & w_{tk,i} \leq d_{tk,i}, \forall k, i \nonumber \\
    & w_{tk,i} \leq y_{t,i} - \sum_{h=1}^{k-1} w_{th,i} + \sum_{h=1}^{k-1}\sum_{j=1}^n P_{th, ji}\left(w_{th,j} + \sum_{o = 1}^{h-1} w_{to, j}  \prod_{l = o}^{h-1} (1 - \sum_{s=1}^nP_{tl,js}) \right), \forall k, i  \label{eq:wx_second_form}. 
    % & w_{tk,i} \ge 0,\quad 
    %   \xi_{t,ij}\ge 0,\quad 
    %   \gamma_{tk,i}\ge 0,\quad
    %   \forall\, k,i,j,
    %   \\
    % & w_{t1,i} \le x_{t1,i},  \quad w_{t1,i} \leq d_{t1,i}, \forall i, \\
    % & w_{t2,i} \le x_{t2,i}, \quad  w_{t2,i} \le  d_{t2,i}, \forall i, \\
    % & \cdots \\
    % & w_{tH,i} \le x_{tH,i}, \quad  w_{tH,i} \le  d_{tH,i}, \forall i.
\end{align}
Let $\muv_{tk}$ be the vector of dual variables associated with the constraints $ w_{t1,i} \leq y_{t,i} - \sum_{h=1}^{k-1} w_{th,i} + \sum_{h=1}^{k-1}\sum_{j=1}^n P_{th, ji}(w_{th,j} + \gamma_{th,j})$. As in \eqref{opt:dual_eqilLP}, by strong duality and optimality of $\muv_{tk}$, it holds that $\text{D-LP}(\yv') - \text{D-LP}(\yv) \geq \sum_{k=1}^H\muv_{tk} ^\T (\yv' - \yv)$, where we notice that the coefficient in front of $\yv_t$ is $1$ for the constraints in \eqref{eq:wx_second_form}. Recall that we define  $\lambdav_{tk}$ as the dual corresponding to the constraints $\wv_{tk} \leq \dv^c_{tk}$ in the original LP \eqref{eq:ogd_olp_extended}. Furthermore, we define $\gv_{tk} = \lambdav_{tk} \circ  \mathbbm{1}\{\dv_{tk}^c = \xv_{tk}\}$  and $\hv_{tk} :=\lambdav_{tk} \circ  \mathbbm{1}\{\dv_{tk}^c \neq \xv_{tk}\}$. Similarly to the single-subperiod case, $\hv_{tk}$ can serve as the dual corresponding to $\wv_{tk} \leq \dv_{tk}$. To recover $\muv_{tk}$ from $\gv_{tk}$, we derive the following recursive relationship between $\gv_{tk}$ and $\muv_{tk}$, which is obtained by aligning the constraints with respect to $\wv_{tk}$ in the dual problem of two LPs. Specifically, for $k = 1,\dots, H$,
\begin{equation}
    \gv_{tk} = \muv_{tk} + (\Iv - \Pv_{tk}) \sum_{l=k+1}^H \muv_{tl} - \sum_{l= k+2}^H \left\{ \sum_{s=k+1}^{l-1}\Pv_{ts} \muv_{tl} \circ \prod_{u=k}^{s-1} \left[(\Iv - \Pv_{tu})\ev  \right] \right\}.
    \label{aeq:recursion}
\end{equation}
Here, $\circ$ denotes Hadamard product, and with slight abuse of notation, $ \prod_{u=k}^{s-1} \left[(\Iv - \Pv_{tu})\ev \right] $ denotes the successive Hadamard product of vectors. Through \eqref{aeq:recursion}, we can solve it recursively for $k = H, H-1, \dots$ to obtain $\muv_{tk}$. We can then verify that the dual optimality condition is satisfied by $\muv_{tk}$ along with $\hv_{tk}$, and dual solutions corresponding to other constraints that are unchanged.

For any $\xv,\yv \in \Delta_{n-1}$, $\Vert \xv - \yv\Vert_2 \leq \Vert \xv \Vert_2 + \Vert \yv\Vert_2 \leq 2$. Invoking Lemma \ref{lem:ogd} to obtain that
\begin{align}
        & \sum\limits_{t=1}^{T}\widetilde{C}_{t}(\xv_t(\bslevel_{t}),\bslevel_{t},\{(\dv_{tk}, \Pv_{tk})\}_{k=1}^H)
        - \min_{\bslevel\in\Delta_{n-1}} \sum\limits_{t=1}^{T}\widetilde{C}_{t}(\xv_t(\bslevel),\bslevel,\{(\dv_{tk}, \Pv_{tk})\}_{k=1}^H)
        \label{eq:pgd_extended_lhs}
        \leq   6  \left\Vert \sum_{h=1}^H \muv_{th} \right\Vert_2\cdot \sqrt{T}.     
\end{align}
 
In Lemma~\ref{lem:disentangle_extended}, we have shown 
\[
\sum_{t=1}^T \left|\widetilde{C}(\xv_{t+1}, \yv_t, \{(\dv_{tk}, \Pv_{tk})\}_{k=1}^H) - \widetilde{C}(\xv_t, \yv_t, \{(\dv_{tk}, \Pv_{tk})\}_{k=1}^H)   \right| 
\leq  \sum_{t=1}^T 2\cdot\left(\max_{i,j=1,\dots,n} c_{ij}\right)\cdot\|\yv_{t}-\yv_{t-1}\|_1.
\]
Because of the update with step size $\frac{1}{\sqrt{t}H}$, 
$
\sum_{t=1}^T 2 \frac{1}{\sqrt{t}H} \sqrt{n}\Vert \gv_t \Vert_2 \leq 2\sqrt{nT} H^{-1}  \left\Vert \sum_{h=1}^H \muv_{th} \right\Vert_2.
$
Similar to the Lipschitz property in Lemma \ref{lem:h_lipschitz}, we can bound the subgradient norm by $\Vert \muv_{th}\Vert_2 \leq  n^2(\max_{i,j} c_{ij}+\max_{i,j} l_{ij})$, and by triangle inequality,
$
\left\Vert \sum_{h=1}^H\muv_{th} \right\Vert_2 \leq  H n^2(\max_{i,j} c_{ij}+\max_{i,j} l_{ij}).
$
Putting all together, the regret can be bounded by
$
n^2T^{1/2}(\max_{i,j} c_{ij}+\max_{i,j} l_{ij}) \cdot (6 H + 2 n^{1/2}) \in O\left( n^{2.5} H\sqrt{T}\right).
$
We note that the bound is with regard to the number of review periods whereas the number of rental subperiods is actually $\widetilde{T} = HT$, and thus the bound also equivalent to $O\left( n^{2.5} \sqrt{H\widetilde{T} }\right)$.
This bound is obtained for any demand and origin-to-destination matrices sequence. To obtain a stochastic version of the bound as in Corollary~\ref{cor:gdr}, one can impose some standard assumption and it follows directly by taking expectations on both sides of the inequality.
\halmos

\subsection{Numerical Results of Extended Model}
\label{appendix:ogr_extended_numerical}

To test the numerical performances of the \pcr{SOAR-Extended} algorithm, we use the optimal solution calculated from the linear programming offline solution as the benchmark to compute the regrets. We note that the validility of the linear program is established in Proposition~\ref{prop:offline_LP_extended}. The one-time learning algorithm (Algorithm~\ref{alg_nloc:otl}) is no longer practical in the extended model for the following reasons: it relies on the network independence assumption and sufficient total inventory to obtain uncensored demand data. However, with multi-subperiod setting, such guarantees are less viable and thus without uncensored demand, such one-time learning is less applicable. The dynamic learning algorithm (Algorithm~\ref{alg_nloc:dl_uncensored_doubling}) would still need the oracle of uncensored data in the extended and thus not eligible for comparison either.

%The key idea is that one can use a very high inventory in the beginning of the period so that the demand will be dominated by such high inventory and therefore avoids being censored.

\begin{figure}[!htb]
    \caption{Regret performances of \pcr{SOAR-Extended} with different model parameters.}\label{fig:regret_hsubperiod_} %
    \begin{subfigure}[b]{0.33\textwidth}
        %\centering
        \includegraphics[width=\linewidth]{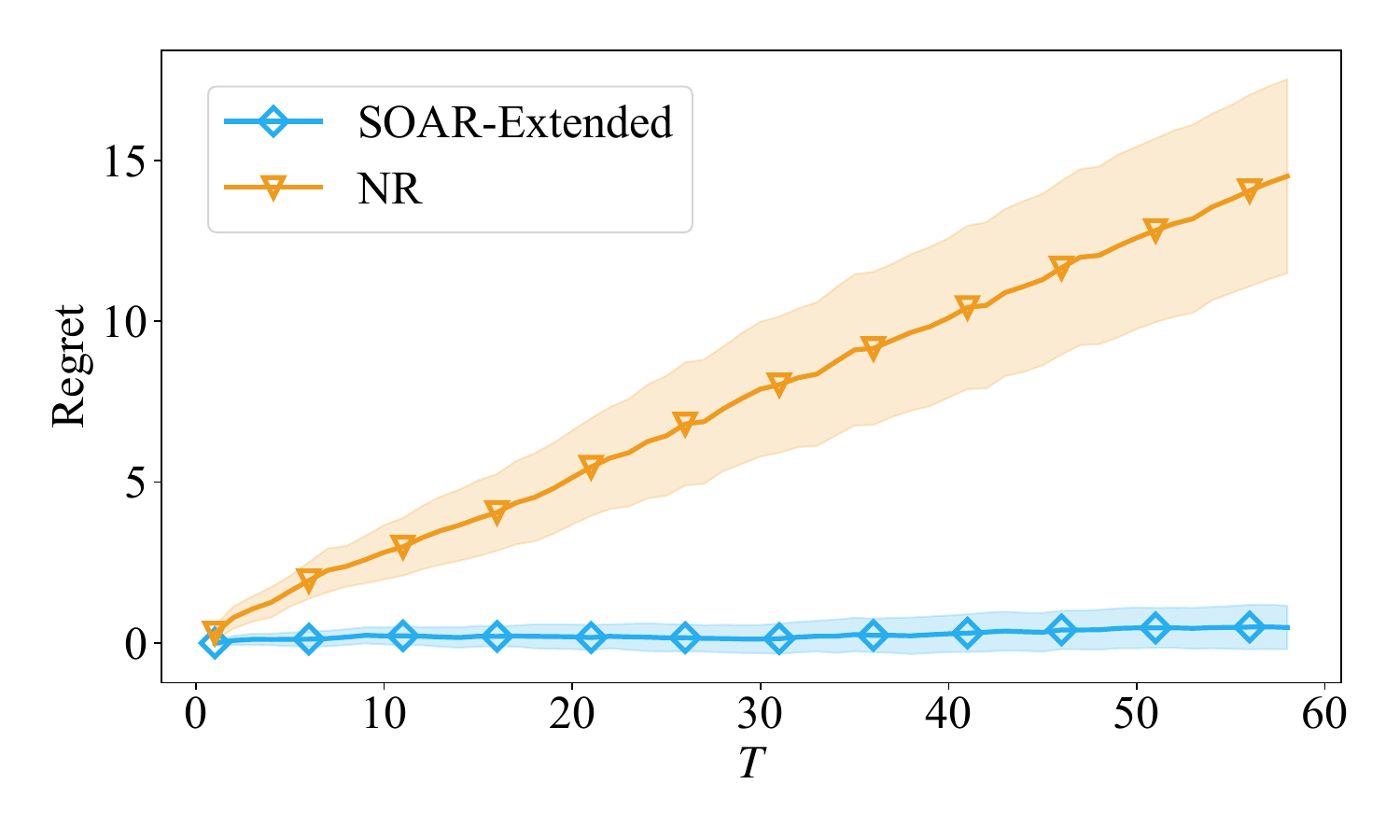}
        \caption{$n = 3, H=4$}
        \label{fig:regret_hsubperiod_subfig1}
    \end{subfigure}
    \begin{subfigure}[b]{0.33\textwidth}
        %\centering
        \includegraphics[width=\linewidth]{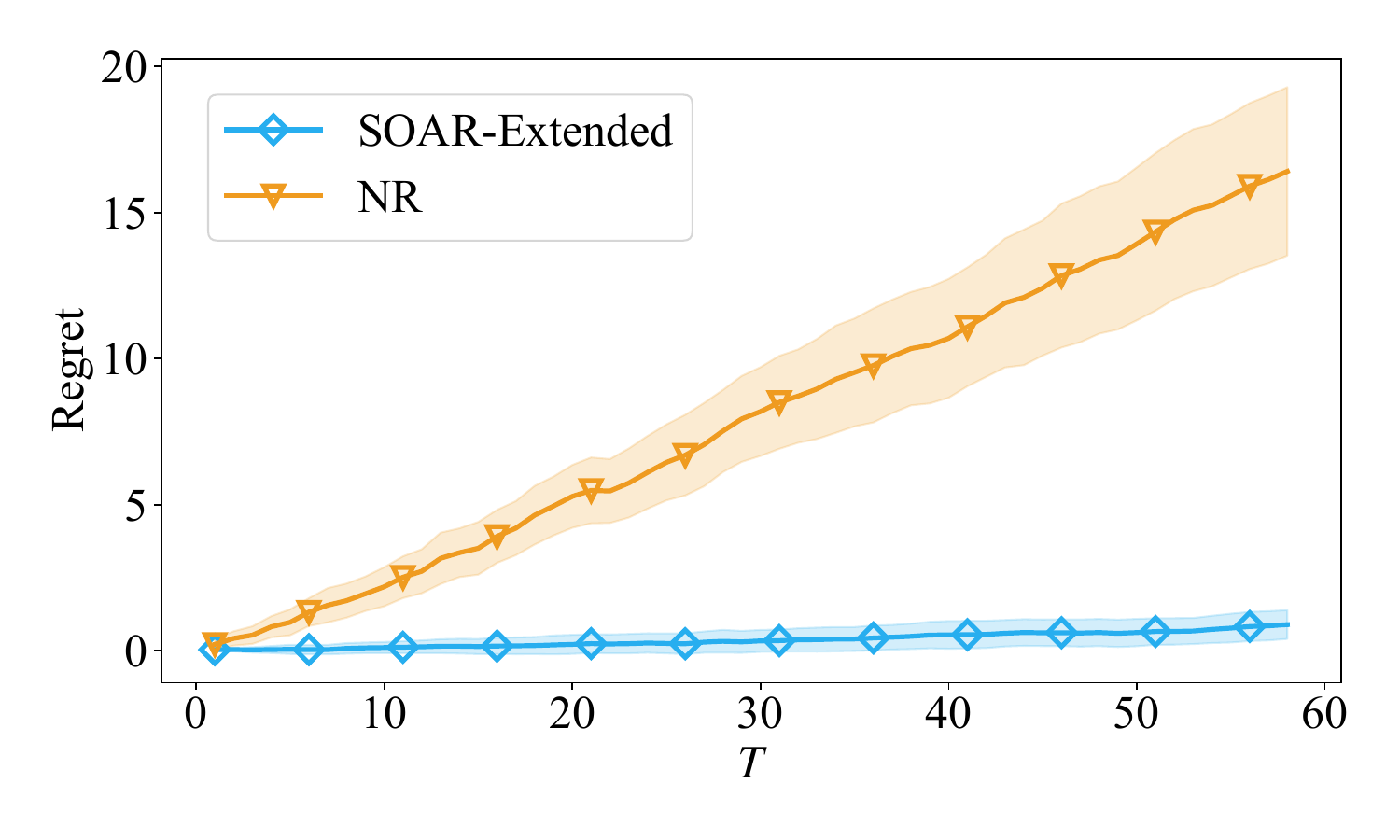}
        \caption{$n = 3, H=8$}
        \label{fig:regret_hsubperiod_subfig2}
    \end{subfigure}
    \begin{subfigure}[b]{0.33\textwidth}
        %\centering
        \includegraphics[width=\linewidth]{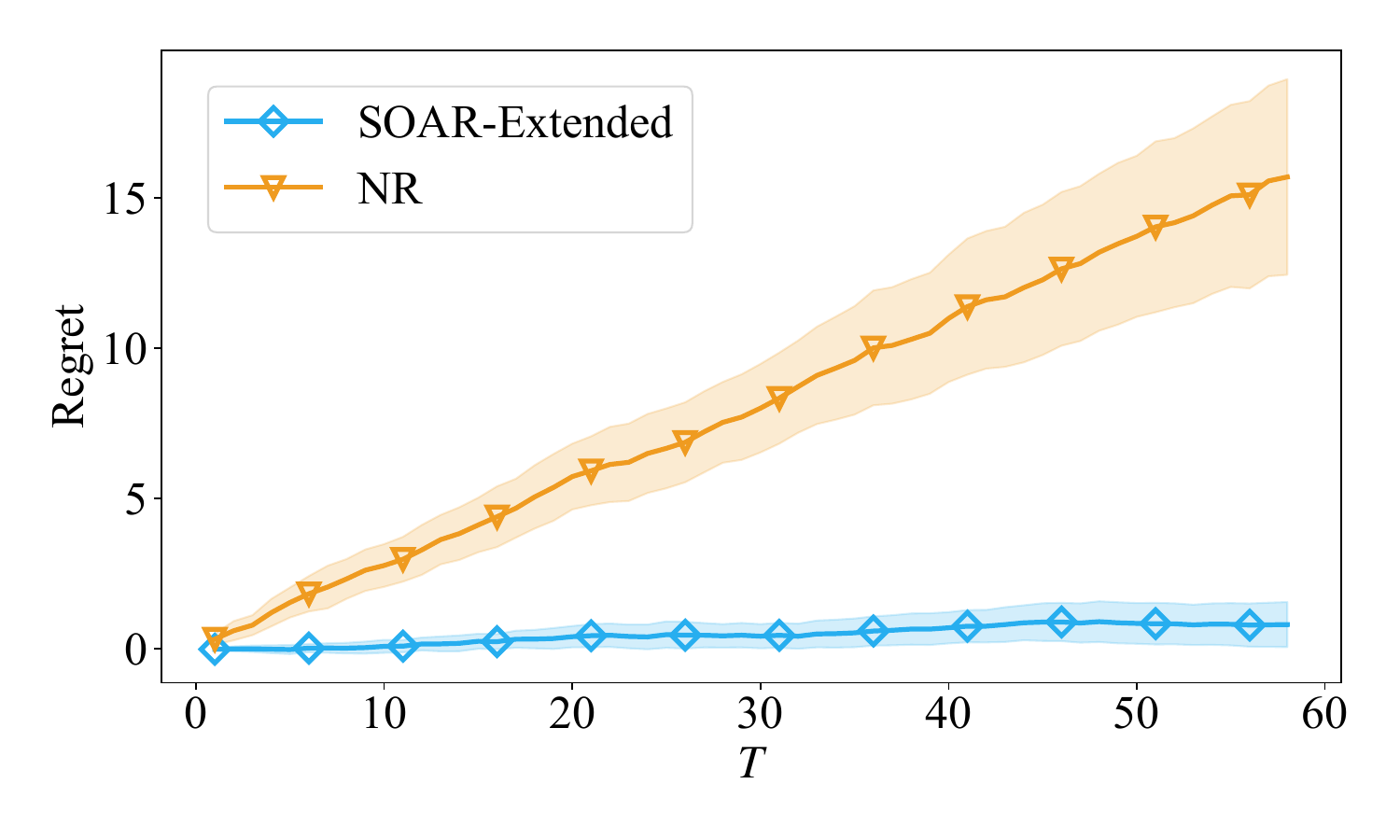}
        \caption{$n = 3, H=24$}
        \label{fig:regret_hsubperiod_subfig3}
    \end{subfigure}
    
    \begin{subfigure}[b]{0.33\textwidth}
        %\centering
        \includegraphics[width=\linewidth]{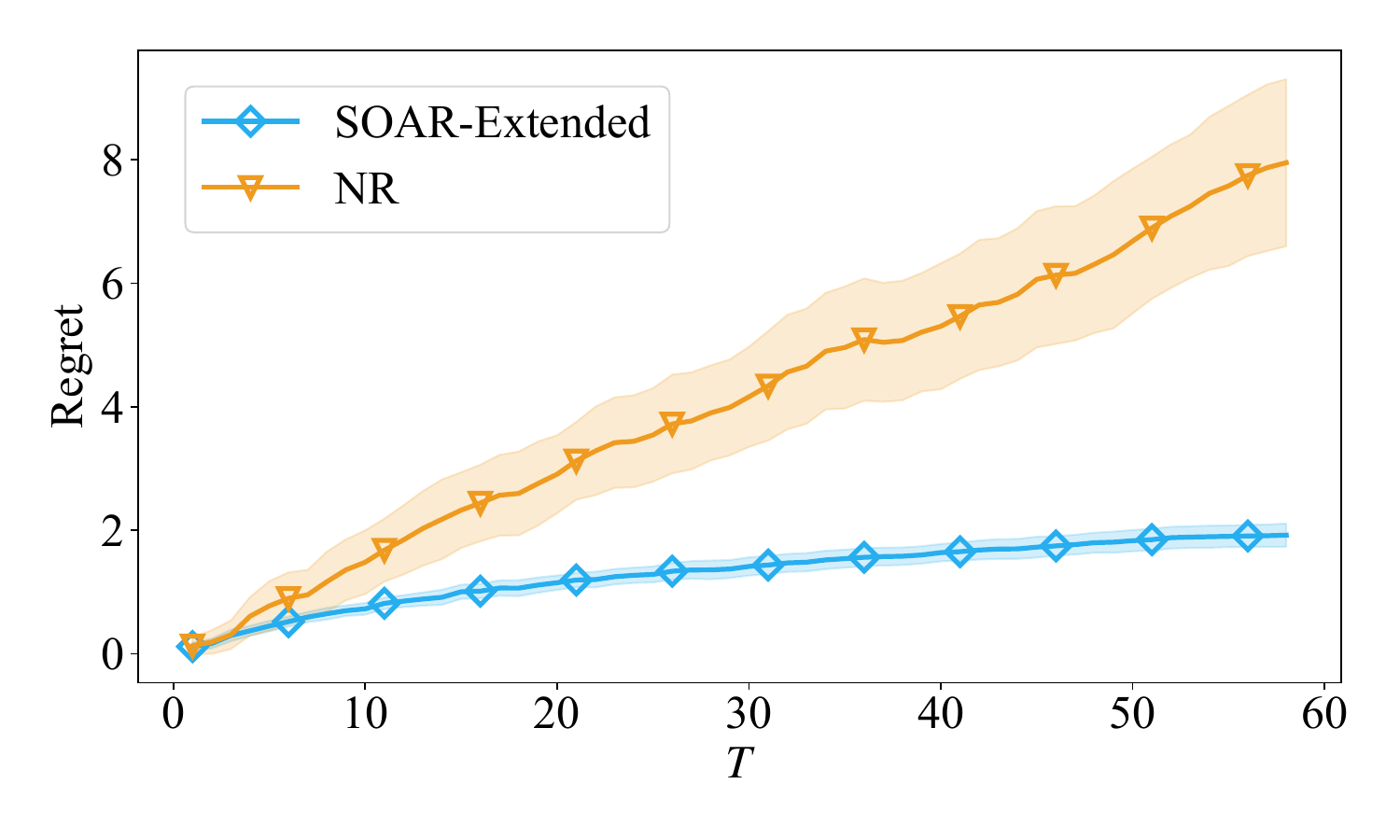}
        \caption{$n = 10, H=4$}
        \label{fig:regret_hsubperiod_subfig4}
    \end{subfigure}
    \begin{subfigure}[b]{0.33\textwidth}
        %\centering
        \includegraphics[width=\linewidth]{files/new-figures/Multi_subperiod_figures/Hsubperiods_n10_T60_H8_exp20_20251009_000402/cumulative_regret_plot_with_ci.pdf}
        \caption{$n = 10, H=8$}
        \label{fig:regret_hsubperiod_subfig5}
    \end{subfigure}
    \begin{subfigure}[b]{0.33\textwidth}
        %\centering
        \includegraphics[width=\linewidth]{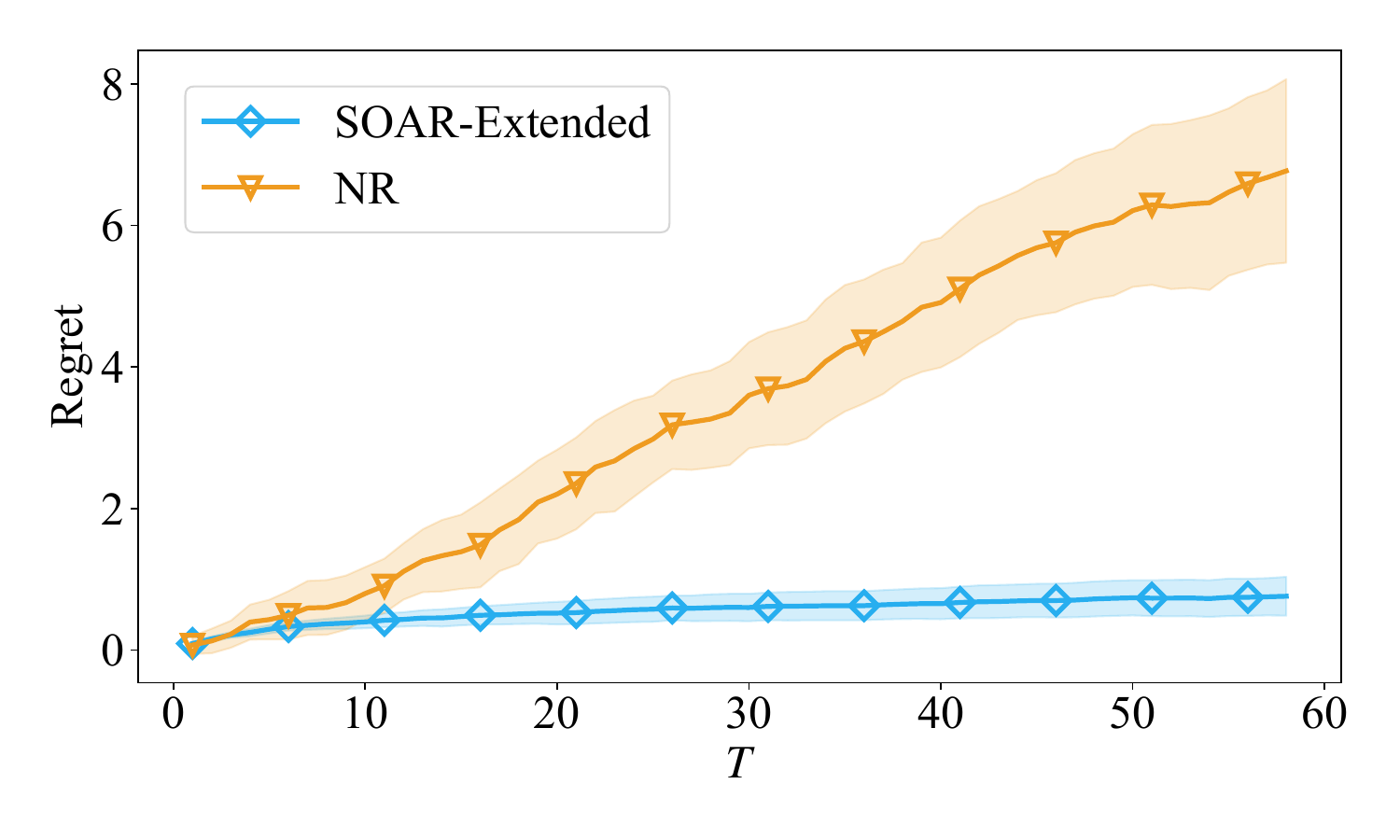}
        \caption{$n = 10, H=24$}
        \label{fig:regret_hsubperiod_subfig6}
    \end{subfigure}
    % {\footnotesize 
    % \emph{Notes.} Since the total inventory is normalized to $1$, and the mean demand originating from each location is scaled by $1/n$ in our data generating processes, an increase in $n$ does not necessarily lead to an increase in the cost or regret scale.}
\end{figure}

Therefore, we compare with the No Repositioning policy in the cumulative regret, with a focus on scenarios without network independence. We fix the length of time horizon as $T=60$ periods and vary the parameters $(n,H) = (3, 4), (3, 8), (3, 24), (10, 4), (10,8), (10, 24)$. For demand in each subperiod, we adopt the network dependence scenario as in Section~\ref{section:numerical}. Furthermore, to account for nonstationarity, we generate $H$ permutations of $[n]$, denoted by $\sigma_h$ for $h = 1,\dots, H$. For each $h$, we first sample a demand vector from the multivariate normal distribution with non-zero correlations, and then permute the demand vector by $\sigma_h$. This parameter choice captures demand nonstationarity, as exemplified by morning and evening rush hours, where locations with peak outbound demand can vary. 

For each origin-to-destination matrix $P$, we construct a matrix $Q$ as follows: elements in the first and second columns of $Q$ are drawn from an exponential distribution with mean $5$, while the remaining elements are drawn from $\mathrm{Unif}(0,1)$; furthermore, for each row, we generate a scale factor from $\mathrm{Unif}(0.80, 0.99)$ representing the total percentage of rental units originating from the locations being returned during this subperiod, and then normalize the row sum to this scale factor to account for the outstanding inventory. We conduct $20$ experimental runs and, and plot both the average and the $95\%$ confidence intervals of regrets computed from these repeated experiments. 

As observed from Figure~\ref{fig:regret_hsubperiod_}, the \pcr{SOAR-Extended} demonstrates superior performance in contrast with the linear regret of the No Repositioning policy. Interestingly, with $n=10$, the regret of \pcr{SOAR-Extended} is actually smaller when $H=8$ than when $H=4$. When the number of subperiods $H$ is increased to $24$, we observe that the regret gap is even lower.  This observation does not contradict our theoretical guarantee with positive dependence on $H$ as that was just an upper bound.  While the current regret dependence on $H$ is moderate, the effectiveness of our algorithm when $H$ is large is commendable, and a finer characterization of $H$'s role in the achievable performance bound is an interesting direction for further investigation. A key intuition behind this phenomenon is that infrequent repositioning naturally leads to less room for improvement between an optimal policy and an algorithmic one. Moreover, the narrow confidence bands of \pcr{SOAR-Extended} in Figure~\ref{fig:regret_hsubperiod_} indicate the robustness of the algorithm's performance.